\pdfoutput=1
\documentclass[11pt]{article}
\usepackage{times}
\usepackage{latexsym}
\usepackage[T1]{fontenc}
\usepackage[utf8]{inputenc}
\usepackage{microtype}
\usepackage{inconsolata}
\usepackage{graphicx}

\PassOptionsToPackage{a4paper,margin=2.3cm,heightrounded=true}{geometry} 
\RequirePackage{geometry}
\newlength\titlebox
 \twocolumn 
\RequirePackage{caption}
\DeclareCaptionFont{10pt}{\fontsize{10pt}{11pt}\selectfont}
\RequirePackage{natbib}
\PassOptionsToPackage{breaklinks}{hyperref}
\RequirePackage{hyperref}
\usepackage{xcolor}
\definecolor{linkcolor}{rgb}{0, 0, 0.5}
\hypersetup{colorlinks=true, citecolor=linkcolor, linkcolor=linkcolor, urlcolor=linkcolor}

\makeatletter
\renewcommand\section{\@startsection{section}{1}{\z@}%
                                  {-2.0ex \@plus -0.5ex \@minus -.2ex}%
                                  {1.5ex \@plus 0.3ex}%
                                  {\large\bfseries\raggedright}}
\makeatother

\makeatletter
\renewenvironment{abstract}{\begin{center}\large\textbf{Abstract}\end{center}\begin{list}{}{\setlength{\rightmargin}{0.55cm}\setlength{\leftmargin}{0.55cm}}\item[]\ignorespaces\@setsize\normalsize{12pt}\xpt\@xpt}{\unskip\end{list}}
\makeatother

\begin{document}

\title{Exploring Compositional Generalization (in COGS/ReCOGS\_pos) by Transformers using Restricted Access Sequence Processing (RASP)}

\author{William Bruns \\
  \texttt{adde.animulis@gmail.com}
}
\date{}

\maketitle

\begin{abstract}
Humans understand new combinations of words encountered if they are combinations of words recognized from different contexts, an ability called Compositional Generalization.
The COGS benchmark \citep{KimLinzen2020} reports 0\% accuracy for Transformer models on some structural generalizations.
We use \citep{Weiss2021}'s Restricted Access Sequence Processing (RASP), a Transformer-equivalent programming language, to demonstrate that a Transformer Encoder-Decoder can perform COGS and the semantically equivalent ReCOGS\_pos\citep{Wu2023} systematically and compositionally: Our RASP models attain near perfect scores on structural generalization splits on COGS (exact match) and ReCOGS\_pos (semantic exact match).
Our RASP models show the (Re)COGS tasks do not require a hierarchical or tree-structured solution (contrary to \citep{KimLinzen2020}, \citep{yao-koller-2022-structural}, \citep{murty2022characterizingintrinsiccompositionalitytransformers}, \citep{liu-etal-2021-learning-algebraic}): we use word-level tokens with an "embedding" layer that tags with possible part of speech\footnote{\begin{footnotesize}per \citep{tenney2019bertrediscoversclassicalnlp} by layer 0 the part-of-speech could be predicted for most words in Transformers trained on a masked language modeling objective, so we assume an equivalent embedding is learnable.\end{footnotesize}}, applying just once per encoder pass 19 attention-head compatible \textbf{flat pattern-matching} rules (easily identified with specific training examples), shown using grammar coverage \citep{fuzzingbook2023:GrammarCoverageFuzzer} to cover the non-recursive aspects of the input grammar, \textbf{plus masking} out prepositional phrases ('pp noun') and/or sentential complements (cp) when recognizing grammar patterns and extracting nouns related to the main verb in the sentence, and output the next logical form (LF) token (repeating until the LF is complete). The models do not apply recursive, tree-structured rules like `np\_det pp np -> np\_pp -> np`, but score near perfect semantic and string exact match on both COGS and ReCOGS pp recursion, cp recursion using the decoder loop.
\end{abstract}

\section{Introduction}

It was long argued that connectionist models (i.e. neural networks) were incapable of compositional generalization \citep{FodorPylyshyn1988}.\footnote{\begin{footnotesize}
More specific versions of this debate continue, for example re: syntax, one can read \citep{vanschijndel2019quantitydoesntbuyquality} vs \citep{goldberg2019assessingbertssyntacticabilities} or re: hierarchical generalization by Transformers, \citep{petty2021transformersgeneralizelinearly} vs \citep{murty2023grokkinghierarchicalstructurevanilla}.\end{footnotesize}} 
However, large language models based on the Transformer architecture \citep{vaswani2017attention} compose seemingly fluent and novel text and are excellent few or zero shot learners \citep{Brown2020}.

Some observations do contradict that Transformers learn systematic, compositional solutions to problems that generalize, for example structural generalization performance on the COGS task \citep{KimLinzen2020} and ReCOGS variant \citep{Wu2023} , benchmarks based on extracting semantics (logical form) from the syntax (grammatical form) of synthetic sentences in a simplified subset of English grammar, requiring models trained only on certain grammar examples to generalize to sentences with unseen grammar built up / recombined from parts present in the training examples.

We use \citep{Weiss2021}'s Restricted Access Sequence Processing (RASP) language that can be compiled to concrete Transformer weights to prove by construction that a Transformer Encoder-Decoder\footnote{\begin{footnotesize}We follow \citep{Zhou2024} and \citep{friedmanrepresentingrulebasedchatbotstransformers} who used RASP to analyze auto-regressive decoder-loop cases, not just Transformer encoders as in \citep{Weiss2021}.\end{footnotesize}} (hereafter EncDec) can perform COGS and ReCOGS\_pos\footnote{\begin{footnotesize}official variant, closer to COGS than non-positional ReCOGS, as COGS is also positional, and means we can also measure string exact match, not just semantic exact match\end{footnotesize}} over the vocabulary and grammar of that task in a systematic, compositional way as a rigorous starting point to investigating when Transformers might learn or not actually learn such compositional/systematic solutions. We find a flat, not hierarchical/tree-structured model which lacks any handling for the recursive rules in the grammar (for prepositional phrase recursion and sentential complement recursion) can perform the task accurately, but requires learning a masking rule\footnote{\begin{footnotesize}to mask out prepositional phrases when extracting noun-verb relationships (which are separable from pp relationship output in the (Re)COGS logical forms), see center-embedded pp examples in the (Re)COGS training set, see Table \ref{RASP-model-flat-patterns-after-masking-to-nv-relationships-table}.\end{footnotesize}} for avoiding "attraction" errors\footnote{\begin{footnotesize}These attraction errors are similar to those discussed elsewhere in NLP and psycholinguistics literature on hierarchical vs linear processing by language models and humans, see "Appendix: Attraction errors" (\ref{attraction_errors}).\end{footnotesize}} where inserted prepositional phrase nouns replace agent/theme/recipient nouns in the logical form by accident. \textbf{This is our main result and suggests that Transformers can perform the (Re)COGS task accurately even for novel prepositional phrase substitution positions or structural recursion depths unseen in training}, turning efforts to learnability of such a generalizing solution (flat pattern matching \textbf{with masking}), and also adds to the literature a caveat on interpreting success on (Re)COGS by noting a hierarchical or tree-structured representation is not necessarily required (contrary to \citep{KimLinzen2020} and assumption of \citep{murty2022characterizingintrinsiccompositionalitytransformers}). Finally, we find that these "attraction" errors predicted for flat pattern matching without masking (and not for recursive/hierarchical/tree-structured representations) are contributing to the high error rate of the \citep{Wu2023} baseline Transformer trained from scratch (EncDec only 2 layers deep).

\section{Prior Literature}

\citep{KimLinzen2020} introduce the COmpositional Generalization Challenge based on Semantic Interpretation (COGS) benchmark, arguing that Transformers have low accuracy on the generalization splits (35\% overall), especially structural generalization splits where near 0\% accuracy is reported, using a 2-layer EncDec Transformer (2 layers for Encoder, 2 layers for Decoder), and speculate that a hierarchical/tree-structured inductive bias is likely to be required to overcome the difficulty in structural generalization. \citep{murty2022characterizingintrinsiccompositionalitytransformers} study how Transformers learn to solve COGS and other compositional datasets and measure the tree-structuredness of the representations and argue it predicts which training checkpoints exhibit better compositional generalization. \citep{yao-koller-2022-structural} survey seq2seq models broadly (both LSTMs and Transformers) and find they all struggle on COGS structural generalizations and give two examples of explicitly tree-structured models (forcing tree-structured representations) as an alternative that do very well \citep{liu-etal-2021-learning-algebraic}\footnote{\citep{liu-etal-2021-learning-algebraic} uses two modules, one that "learns to model the latent syntactic algebra [using a Tree-LSTM to] produce the latent syntactic structure of each expression" and a second that "learns to assign semantic operations to syntactic operations" so that they can "transform a syntactic tree to the final composed semantic meaning", and in this way achieves 97.7\% on COGS. However this is neither a Transformer architecture nor a vanilla unstructured neural network, it uses an LSTM specifically architected based on the assumption of a recursive, tree-structured context-free grammar. Note LeAR also uses what appear to be heavily crafted semantic primitives in their Interpreter module, see their Table 8 for COGS, requiring manual task specific design.} and \citep{weissenhorn-etal-2022-compositional}. Finally, \citep{lake2023human} use a "meta-learning for compositionality" approach with a 3-layer EncDec Transformer architecture and achieve what they call "human-like systematic generalization", achieving high scores on everything in the COGS benchmark (>99\% on lexical generalizations) EXCEPT the structural generalization splits where they also still score 0\% accuracy.
However, one notices these networks are shallow compared with those used in successful large-pretrained Transformer models (e.g. 24-layer BERT where compositional parse trees seem to be encoded in its vector space representation \citep{hewitt-manning-2019-structural}), and it is claimed, by e.g. \citep{Csordas2022} that for compositional operations, like parsing, the depth of the network must be at least the maximum number of compositional operations, e.g. the height of the parse tree. Remarkably, \citep{petty2024impactdepthcompositionalgeneralization} finds that increasing the layer depth of the Transformer models (up to 32 layers) does not improve the near 0\% accuracy on COGS structural generalization splits like prepositional phrase modification of subject when the network has only seen it on the object during training and also input length/depth generalizations (like pp/cp recursion), perhaps surprising as for the simpler logical inferences problem in \citep{Clark2020} they observed successful logical inference depth generalization even by Transformer Encoder-only Transformers.

\begin{figure}
\includegraphics[scale=0.560]{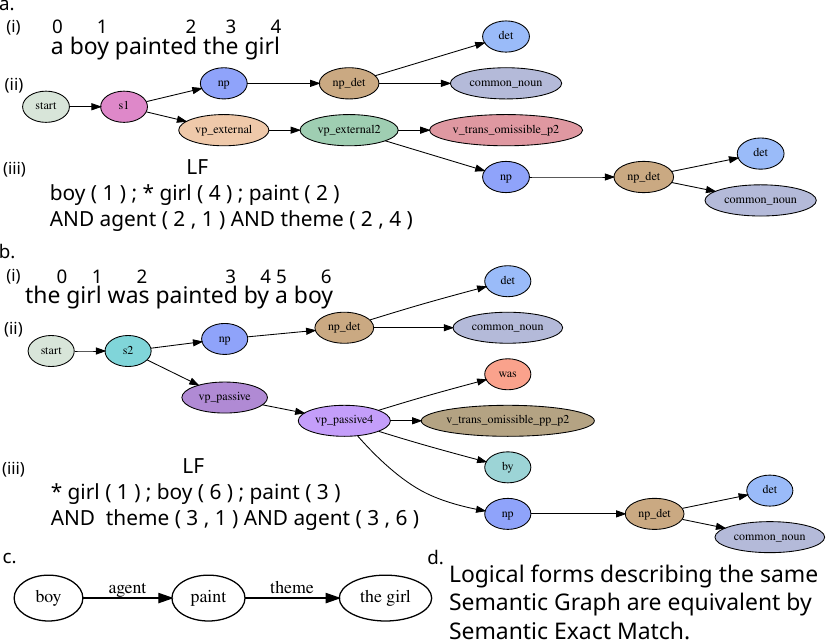}
\caption{\begin{small}Introducing parse trees, logical form, and semantic graphs. Two semantically identical but syntactically distinct (i) sentences (a) "a boy painted the girl" and (b) "the girl was painted by a boy" are shown with (ii) their distinct parse tree (parsed into the shared (Re)COGS input grammar), (iii) the string form of their semantics (ReCOGS logical form; differs in indices and ordering), and (c) the graph representation of their logical form (semantic graph\protect\footnotemark, not different at all between the two examples). Note the (iii) logical forms (LFs) differ by String Exact Match but not \citep{Wu2023}'s Semantic Exact Match (order and indices do not match but nouns, normalized verbs, and relationships between nouns and verbs are same). Note the "agent", "theme" order in the logical form string is not required to match for Semantic Exact Match. COGS and ReCOGS tasks require extracting the semantics (c) encoded in LFs (iii) of sentences (i). \textbf{RASP-for-(Re)COGS shows decoding the LF} (Figures \ref{rasp-for-recogs-decoder-loop-supplementary-figure-incl-encoder-and-decoder-and-grammar-vertical} (ReCOGS), \ref{rasp-for-cogs-encoder-decoder-with-grammar-patterns} (COGS)) \textbf{does NOT require representing the parse tree.}\end{small}}
\label{data-and-task-figure}
\end{figure}

Thankfully, \citep{Wu2023} are able to begin to get traction (low but nonzero accuracy) for the shallow EncDec Transformer models on structural generalizations in a modified but semantically equivalent form of the COGS task they call ReCOGS. They remove redundant symbols, and use Semantic Exact Match instead of Exact Match (Fig \ref{data-and-task-figure}).

\citep{Zhou2024} apply \citep{Weiss2021}'s RASP language to explain some inconsistent findings regarding generalization, using RASP to predict some cases of generalization that come easily to Transformers and some which do not.

\citep{Zhou2024} seem to reveal \citep{Weiss2021} has provided the framework we seek by demonstrating how to apply RASP to Transformer decoders with intermediate steps, and even use it to learn how to modify difficult-to-learn tasks like Parity\footnote{\begin{footnotesize}See \citep{Strobl2024} for context from formal language theory, computational complexity, circuit complexity theory, and experimental papers together, providing robust lower and upper bounds on what Transformers can do, including discussion of under what conditions Parity can be solved by Transformers and how whether it can be learned by randomly initialized Transformers under simple training schemes is a different question (general feed-forward neural networks can learn to solve Parity per \citep{10.7551/mitpress/4943.003.0128}). \citep{delétang2023neuralnetworkschomskyhierarchy} also. See Appendix (\ref{appendix_long_addition_and_parity})\end{footnotesize}} and long addition in seemingly incidental ways based on RASP analysis to make them readily learnable by Transformers in a compositional, length generalizing way!

Thus we apply techniques similar to \citep{Zhou2024} and \citep{Weiss2021} to (Re)COGS to (1) argue Transformers should be able to perform the (Re)COGS task accurately even for novel prepositional phrase substitution positions or structural recursion depths unseen in training, and that the problem is learning not capability and (2) try to understand the prepositional phrase modification related generalization errors \citep{Wu2023}'s baseline EncDec Transformers are making.
\footnotetext{\begin{footnotesize}As a convention, in converting ReCOGS logical forms to Semantic Graphs (used for our illustrations and to explain Semantic Exact Match, not in the code) we use the logical form (source, target) index order for directed semantic graph edges (from verb to related entity) EXCEPT for the agent relationship which is from the agent of a verb to the verb (opposite direction from logical form in that case), which gives our semantic graphs of ReCOGS sentences an unambiguous starting point (layout starts from agent) without affecting comparison of the graphs (generated by a consistent rule), see also Figure \ref{baseline_transformer_standard_training_cannot_do_v_dat_p2_generalization}. For more information on equivalence of (Re)COGS LF with semantic graphs see e.g. Section 3.2 of \citep{weissenhorn-etal-2022-compositional} who use a semantic graph based representation in their model (they use different conventions but show such mappings exist).\end{footnotesize}}
\begin{figure}
\includegraphics[scale=0.55]{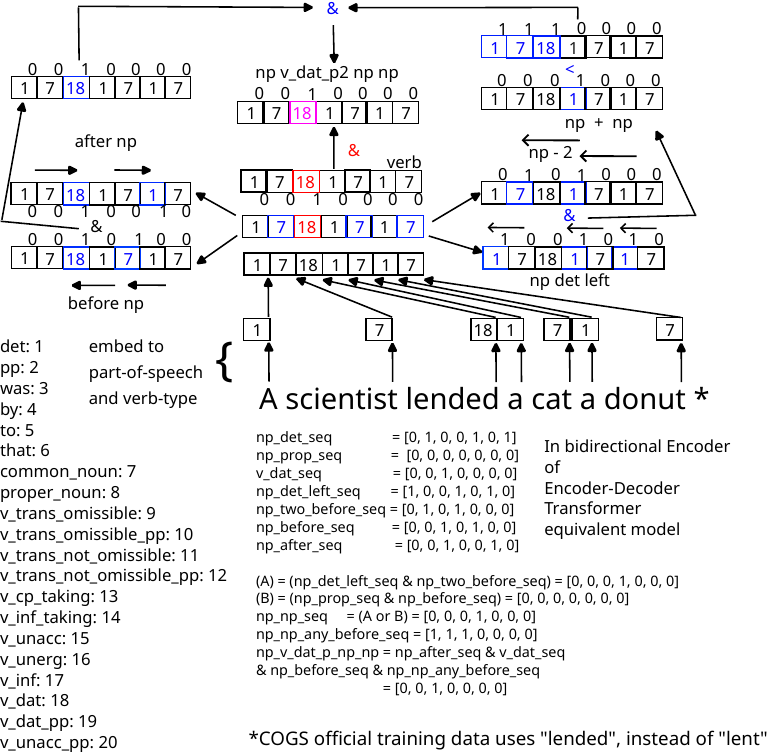}
\caption{\begin{small}RASP-for-(Re)COGS Encoder determines an example contains "np v\_dat\_p2 np np" via flat pattern matching (maps to agent, recipient, theme noun-verb relation ordering (\ref{RASP-model-flat-patterns-after-masking-to-nv-relationships-table})). See Figure \ref{rasp-model-flat-pattern-match-example-v_dat_p2_match_with_middle_pp} for middle-noun pp modification, Figure \ref{rasp-model-flat-pattern-match-example-v_dat_p2_non_matching} for the non-matching case, and Figure \ref{rasp-for-recogs-decoder-loop-supplementary-figure-incl-encoder-and-decoder-and-grammar-vertical} (ReCOGS), \ref{rasp-for-cogs-encoder-decoder-with-grammar-patterns} (COGS) for the full Encoder-Decoder models translating input to logical form. 1d embed. equiv. to higher dim orthogonal vector idx (A\ref{rasp-word-level-model-design}).\end{small}}
\label{rasp-model-flat-pattern-match-example}
\end{figure}

\section{Data}

Figure \ref{data-and-task-figure}.
COGS\footnote{\begin{footnotesize}\href{https://github.com/najoungkim/COGS}{https://github.com/najoungkim/COGS}\end{footnotesize}} \citep{KimLinzen2020} and ReCOGS\footnote{\begin{footnotesize}\href{https://github.com/frankaging/ReCOGS}{https://github.com/frankaging/ReCOGS}\end{footnotesize}} \citep{Wu2023} based on COGS \citep{KimLinzen2020} (input sentences/grammar are the same but logical form output are different) datasets were used, with special attention on the structural generalization splits (especially prepositional phrase Object-to-Subject generalization). The grammar and vocabulary description for COGS English input sentences provided in the utilities associated with the IBM CPG project \citep{klinger2024compositionalprogramgenerationfewshot}\footnote{\begin{footnotesize}https://github.com/IBM/cpg/blob/
c3626b4e03bfc681be2c2a5b23da0b48abe6f570
/src/model/cogs\_data.py\#L523
\end{footnotesize}}
were used in designing our RASP solution and analyzing the ways in which this task could be learned (we did not actually use their grammar though, and our RASP solution is flat and non-hierarchical unlike their description of the COGS probabilistic context-free grammar which is hierarchical and recursive).

\section{Model}
\label{model}
Figures \ref{rasp-model-flat-pattern-match-example} ((Re)COGS shared encoder), \ref{rasp-for-recogs-decoder-loop-supplementary-figure-incl-encoder-and-decoder-and-grammar-vertical} (ReCOGS EncDec) , \ref{rasp-for-cogs-encoder-decoder-with-grammar-patterns} (COGS EncDec).
Model code is available: \href{https://github.com/willy-b/RASP-for-COGS}{RASP-for-COGS} and \href{https://github.com/willy-b/learning-rasp}{RASP-for-ReCOGS}\footnote{\begin{footnotesize}\href{https://github.com/willy-b/RASP-for-COGS}{https://github.com/willy-b/RASP-for-COGS} (COGS) and \href{https://github.com/willy-b/learning-rasp}{https://github.com/willy-b/learning-rasp} (ReCOGS)\end{footnotesize}}. We used the RASP interpreter of \citep{Weiss2021}. For RASP model design and details see Appendix \ref{rasp-word-level-model-design}. We use word-level tokens for all RASP model results in this paper,\footnote{\begin{footnotesize}We believe any solution at the word-level can be converted to a character-level token solution (see Appendix \ref{rasp_character_level_model_notes}).
\end{footnotesize}
}
with an "embedding" layer\footnote{\begin{footnotesize}1d scalar embedding, the default in RASP. Equivalent to specifying an orthogonal vector by index in a higher dimensional space. E.g. embedding to common\_noun (7) is as setting 7th dimension to 1 in one-hot, or just using 7th orthogonal vector (given some numbering so we can assign one to each POS and verb type). Equality checks for the 1d scalars are equivalent to vector dot product on the indexed orthogonal vectors. Checking for EITHER common\_noun OR proper\_noun can use a dot product with the non-orthogonal vector given by the sum of (common\_noun + proper\_noun).
\end{footnotesize}} that tags with possible part of speech\footnote{\begin{footnotesize}Note we followed the \citep{klinger2024compositionalprogramgenerationfewshot} description of COGS and include in our RASP-for-(Re)COGS vocabulary (part-of-speech or verb-type embedding/mapping) all words occurring anywhere in the upstream (Re)COGS "train.tsv". For RASP-for-ReCOGS only, we also include two words in our vocab/embedding as common nouns accidentally left out of train.tsv vocabulary by the COGS author (COGS vocab is used by ReCOGS): "monastery" and "gardner" (only included in their train\_100.tsv and dev.tsv not also in train.tsv, but present in test/gen), a decision affecting just 22 or 0.1\% of generalization examples so would not affect any conclusions qualitatively. For RASP-for-COGS which was started after RASP-for-ReCOGS, we considered these words out-of-vocabulary to be more conservative. See also the discussion on COGS Github with the COGS author at \href{https://github.com/najoungkim/COGS/issues/2\#issuecomment-976216841}{https://github.com/najoungkim/COGS/issues/2
\#issuecomment-976216841} .\end{footnotesize}}, and apply just once per encoder pass 19 attention-head compatible flat pattern-matching rules (Table \ref{RASP-model-flat-patterns-after-masking-to-nv-relationships-table}, Figures \ref{rasp-model-flat-pattern-match-example}, \ref{rasp-model-flat-pattern-match-example-v_dat_p2_match_with_middle_pp}, \ref{rasp-model-flat-pattern-match-example-v_dat_p2_non_matching}; official training examples the rules can be derived from given in Table \ref{RASP-model-flat-patterns-after-masking-to-nv-relationships-table}), shown using grammar coverage \citep{fuzzingbook2023:GrammarCoverageFuzzer} to cover the non-recursive aspects of the input grammar, plus masking out all prepositional phrases and/or sentential complements (cp) when extracting nouns related to the main verb in the sentence (derived from center embedded pp in training) and separately unrolling local prepositional phrase noun modifiers (Figure \ref{rasp-for-recogs-decoder-loop-supplementary-figure_pp_depth_2}) / local sentential complement nouns and verb relationships (Figure \ref{rasp-for-recogs-decoder-loop-supplementary-figure_cp_depth_2}) one by one in the decoder loop. Each pattern handles "det common\_noun" and "proper\_noun" identically, a symmetry which is evident in the training data. The model does NOT apply recursive, tree-structured rules like `np\_det pp np -> np\_pp -> np`.

Consistent with \citep{Zhou2024} we use \citep{Weiss2021}'s RASP originally used for modeling Transformer encoders to model an encoder-decoder in a causal way by feeding the autoregressive output back into the program.\footnote{\begin{footnotesize}We only have aggregations with non-causal masks when that aggregation (or without loss of generality just before the aggregation product is used to avoid multiplying everywhere) is masked by an input mask restricting it to the sequence corresponding to the input.\end{footnotesize}}

\begin{table}
\centering
\begin{tabular}{p{0.225\linewidth} p{0.50\linewidth} p{0.15\linewidth}}
\hline
\begin{tiny}\textbf{Official COGS/ReCOGS training sentence}\end{tiny} & \begin{tiny}\textbf{POS/verb-type sequence matched}\end{tiny} & \begin{tiny}\textbf{Noun-verb relationship template}\end{tiny}\\
\hline
\hline
\begin{tiny}The donut 

was studied .\end{tiny} & \begin{tiny}((det common)|proper)

was v\_trans\_omissible\_pp\_p1\end{tiny} & \begin{tiny}theme\end{tiny} \\
\hline
\begin{tiny}The captain ate .\end{tiny} & \begin{tiny}((det common)|proper) v\_trans\_omissible\_p1\end{tiny} & \begin{tiny}agent\end{tiny} \\
\hline
\begin{tiny}The sailor 

dusted a boy .\end{tiny} & \begin{tiny}((det common)|proper)

v\_trans\_omissible\_p2 ((det common)|proper)\end{tiny} & \begin{tiny} agent,

theme \end{tiny} \\
\hline
\begin{tiny}A drink 

was eaten by 

a child .\end{tiny} & \begin{tiny}((det common)|proper)

was v\_trans\_omissible\_pp\_p2 by 

((det common)|proper)\end{tiny} & \begin{tiny}theme,

agent\end{tiny} \\
\hline
\begin{tiny}A girl 

liked the raisin .\end{tiny} & \begin{tiny}((det common)|proper)

v\_trans\_not\_omissible ((det common)|proper)\end{tiny} & \begin{tiny}agent,

theme\end{tiny} \\
\hline
\begin{tiny}The pen 

was helped .\end{tiny} & \begin{tiny}((det common)|proper)

was v\_trans\_not\_omissible\_pp\_p1\end{tiny} & \begin{tiny}theme\end{tiny} \\
\hline
\begin{tiny}A rose 

was helped by

a dog .\end{tiny} & \begin{tiny}((det common)|proper)

was v\_trans\_not\_omissible\_pp\_p2 by 

((det common)|proper)\end{tiny} & \begin{tiny}theme,

agent\end{tiny} \\
\hline
\begin{tiny}A cat 

disintegrated a girl .\end{tiny} & \begin{tiny}((det common)|proper)

v\_unacc\_p1 ((det common)|proper)\end{tiny} & \begin{tiny}agent,

theme\end{tiny} \\
\hline
\begin{tiny}A box was inflated .\end{tiny} & \begin{tiny}((det common)|proper) was v\_unacc\_pp\_p1\end{tiny} & \begin{tiny}theme\end{tiny} \\
\hline
\begin{tiny}The cake was frozen 

by the giraffe .\end{tiny} & \begin{tiny}((det common)|proper) was v\_unacc\_pp\_p2 

by ((det common)|proper)\end{tiny} & \begin{tiny}theme,

agent\end{tiny} \\
\hline
\begin{tiny}The girl needed to cook .\end{tiny} & \begin{tiny}((det common)|proper) v\_inf\_taking to v\_inf\end{tiny} & \begin{tiny}agent\end{tiny} \\
\hline
\begin{tiny}The sailor laughed .\end{tiny} & \begin{tiny}((det common)|proper) v\_unerg\end{tiny} & \begin{tiny}agent\end{tiny} \\
\hline
\begin{tiny}A cake rolled .\end{tiny} & \begin{tiny}((det common)|proper)

v\_unacc\_p2\end{tiny} & \begin{tiny}theme\end{tiny} \\
\hline
\begin{tiny}A girl 

was sold the cake .\end{tiny} & \begin{tiny}((det common)|proper)

was v\_dat\_pp\_p3 ((det common)|proper)\end{tiny} & \begin{tiny}recipient,

theme\end{tiny} \\
\hline
\begin{tiny}The girl was lended 

the balloon 

by Harper .\end{tiny} & \begin{tiny}((det common)|proper) was v\_dat\_pp\_p4 

((det common)|proper)

by ((det common)|proper)\end{tiny} & \begin{tiny}recipient,

theme,

agent\end{tiny} \\
\hline
\begin{tiny}The pen was offered

to the girl 

by Emma .\end{tiny} & \begin{tiny}((det common)|proper) was v\_dat\_pp\_p2 

to ((det common)|proper) 

by ((det common)|proper)\end{tiny} & \begin{tiny}theme,

recipient,

agent\end{tiny} \\
\hline
\begin{tiny}The melon was lended to a girl .\end{tiny} & \begin{tiny}((det common)|proper)

was v\_dat\_pp\_p1 to ((det common)|proper)\end{tiny} & \begin{tiny}theme,

recipient\end{tiny} \\
\hline
\begin{tiny}Liam 

forwarded the girl 

the donut .\end{tiny} & \begin{tiny}((det common)|proper)

v\_dat\_p2 ((det common)|proper) 

((det common)|proper)\end{tiny} & \begin{tiny}agent,

recipient,

theme\end{tiny} \\
\hline
\begin{tiny}Emma 

passed a cake to 

the girl .\end{tiny} & \begin{tiny}((det common)|proper)

v\_dat\_p1 ((det common)|proper)

to ((det common)|proper)\end{tiny} & \begin{tiny}agent,

theme,

recipient\end{tiny} \\
\hline
\hline
\begin{tiny}Center-embed. PP

training example \end{tiny}&\begin{tiny}RASP-for-(Re)COGS model


$\rightarrow$ mask "PP noun" rule\end{tiny} & \begin{tiny}(when finding NV rels.)\end{tiny} \\
\hline
\hline
\begin{tiny}Isabella 

forwarded a box 

on a tree 

to Emma .\end{tiny} & \begin{tiny}((det common)|proper) 

v\_dat\_p1 ((det common)|proper) 

\_(masked pp det common\_noun)\_ 

to ((det common)|proper)\end{tiny} & \begin{tiny}agent, 

theme,

(masked),

recipient\end{tiny} \\
\hline
\end{tabular}
\caption{19 flat part of speech / verb type patterns (matched by Encoder, Figures \ref{rasp-model-flat-pattern-match-example}, \ref{rasp-for-recogs-decoder-loop-supplementary-figure-incl-encoder-and-decoder-and-grammar-vertical}) derived from training examples are sufficient to cover the (Re)COGS input grammar after pp/cp masking (Figures \ref{rasp-for-recogs-decoder-loop-supplementary-figure_pp_depth_2}, \ref{rasp-for-recogs-decoder-loop-supplementary-figure_cp_depth_2}) to obtain noun-verb relationship ordering for the main verb.
}
\label{RASP-model-flat-patterns-after-masking-to-nv-relationships-table}
\end{table}

\begin{table}
\centering
\begin{tabular}{p{0.6\linewidth} p{0.3\linewidth}}
\hline
\begin{tiny}\textbf{ReCOGS\_pos Split}\end{tiny} & \begin{tiny}\textbf{Semantic Exact Match \% (95\% CI)}\end{tiny} \\
\hline
\begin{tiny}ReCOGS\_pos test set (held out, in-distribution)\end{tiny} & \begin{tiny} 100.00\% (99.88-100.00\%)\end{tiny} \\
\hline
\begin{tiny}Structural generalization splits (held out, out-of-distribution)\end{tiny} & \\
\hline
\begin{tiny}obj\_pp\_to\_subj\_pp\end{tiny} & \begin{tiny}  92.20\% (90.36- 93.79\%)\end{tiny} \\
\begin{tiny}pp\_recursion\end{tiny} & \begin{tiny} 100.00\% (99.63-100.00\%)\end{tiny} \\
\begin{tiny}cp\_recursion\end{tiny} & \begin{tiny} 100.00\% (99.63-100.00\%)\end{tiny} \\
\hline
\begin{tiny}all gen splits (21K examples, aggregate)\end{tiny} & \begin{tiny}99.63\% (99.54-99.71\%)\end{tiny} \\
\hline
\end{tabular}
\caption{\begin{small}\textbf{RASP-for-ReCOGS} Encoder-Decoder Transformer-compatible model performance on the ReCOGS\_pos test set (n=3000) and out-of-distribution structural generalization split performance (n=1000 per gen split). Model is deterministic, so Clopper-Pearson CIs are used. (Model was developed on training data or examples identical to training examples after embedding words to part of speech and verb type sequences, the first step in the model, see Tables \ref{RASP-model-flat-patterns-after-masking-to-nv-relationships-table} , \ref{specific-grammar-pattern-examples-mapped-to-part-of-speech-and-cogs-in-distribution-training-examples}). 100.00\% (CI 99.63-100.00\%) string exact match was also attained for recursion splits. See Table \ref{results-table-full} for the rest of the 21 generalization splits (only 3 structural shown).
\end{small}}
\label{results-table}
\end{table}

\begin{table}
\centering
\begin{tabular}{p{0.52\linewidth} p{0.20\linewidth} p{0.15\linewidth}}
\hline
\begin{tiny}\textbf{ReCOGS\_pos Split}\end{tiny} & \begin{tiny}\textbf{Semantic Exact Match \%}

+/- sample std\end{tiny} & \begin{tiny} \textbf{95\% CI}

(of 

sample mean)\end{tiny} \\
\hline
\begin{tiny}Structural generalization splits (held out, o.o.d.)\end{tiny} & \\
\hline
\begin{tiny}pp\_recursion (up to depth 12 vs 0-2 in training)\end{tiny} & \begin{tiny} 40.2\% +/- 9.3\% \end{tiny} & \begin{tiny}36.1-44.2\%\end{tiny} \\
\begin{tiny}cp\_recursion (up to depth 12 vs 0-2 in training)\end{tiny} & \begin{tiny} 52.4\% +/- 1.4\% \end{tiny} & \begin{tiny}51.8-53.0\%\end{tiny} \\
\begin{tiny}obj\_pp\_to\_subj\_pp (N=20)\end{tiny} & \begin{tiny} 19.7\% +/- 6.1\% \end{tiny} & \begin{tiny}17.0-22.4\%\end{tiny} \\
\begin{tiny} $\rightarrow$ + 1 layer (N=10)\end{tiny} & \begin{tiny} 16.2\% +/- 2.7\%\end{tiny} & \begin{tiny}14.6-17.9\%\end{tiny} \\
\begin{tiny} $\rightarrow$ + 2 layers (N=10)\end{tiny} & \begin{tiny} 19.3\% +/- 4.1\%\end{tiny} & \begin{tiny}16.8-21.8\%\end{tiny} \\
\hline
\end{tabular}
\caption{\begin{small}Out-of-distribution (o.o.d.) ReCOGS\_pos structural generalization split performance (n=1000 per gen split) for N=20 \textbf{\citep{Wu2023} baseline Encoder-Decoder Transformer models trained from scratch} on random perm. of the training set and random weight init. N=10 for +layer expts only. Results consistent with \citep{Wu2023}'s Fig 5.
\end{small}}
\label{wu2023-results-table}
\end{table}

\begin{table}
\centering
\begin{tabular}{p{0.6\linewidth} p{0.3\linewidth}}
\hline
\begin{tiny}\textbf{COGS Split}\end{tiny} & \begin{tiny}\textbf{String Exact Match \% (95\% CI)}\end{tiny} \\
\hline
\begin{tiny}COGS test set (held out, in-distribution)\end{tiny} & \begin{tiny} 99.97\% (99.81-99.99\%)\end{tiny} \\
\hline

\begin{tiny}Structural generalization splits (held out, out-of-distribution)\end{tiny} & \\
\hline
\begin{tiny}obj\_pp\_to\_subj\_pp\end{tiny} & \begin{tiny}  100.00\% (99.63-100.00\%)\end{tiny} \\
\begin{tiny}pp\_recursion\end{tiny} & \begin{tiny} 98.40\% (97.41-99.08\%)\end{tiny} \\
\begin{tiny}cp\_recursion\end{tiny} & \begin{tiny} 99.90\% (99.44-99.997\%)\end{tiny} \\
\hline
\begin{tiny}all gen splits (21K examples, aggregate)\end{tiny} & \begin{tiny}99.89\% (99.83-99.93\%)\end{tiny} \\
\hline
\end{tabular}
\caption{\begin{small}\textbf{RASP-for-COGS} Encoder-Decoder Transformer-compatible model performance on the COGS test set (n=3000) and out-of-distribution structural generalization split performance (n=1000 per gen split). Model is deterministic, so Clopper-Pearson CIs are used. See Table \ref{rasp-for-cogs-results-table-full} for the rest of the 21 generalization splits (only 3 structural shown).
\end{small}}
\label{rasp-for-cogs-results-table}
\end{table}

For training Transformers from scratch with randomly initialized weights, we use scripts derived from those provided by \citep{Wu2023}\footnote{
\begin{footnotesize}\href{https://github.com/frankaging/ReCOGS/blob/1b6eca8ff4dca5fd2fb284a7d470998af5083beb/run\_cogs.py}{https://github.com/frankaging/ReCOGS/blob/
1b6eca8ff4dca5fd2fb284a7d470998af5083beb/run\_cogs.py}

and \href{https://github.com/frankaging/ReCOGS/blob/1b6eca8ff4dca5fd2fb284a7d470998af5083beb/model/encoder\_decoder\_hf.py}{https://github.com/frankaging/ReCOGS/blob/
1b6eca8ff4dca5fd2fb284a7d470998af5083beb
/model/encoder\_decoder\_hf.py} 
\end{footnotesize}
}. See "Appendix: Model Detail" (\ref{model_detail}).
\section{Methods}
We use the RASP \citep{Weiss2021} interpreter\footnote{\begin{footnotesize}provided at \href{https://github.com/tech-srl/RASP/}{https://github.com/tech-srl/RASP/}
\end{footnotesize}
} to evaluate our RASP programs. Logical forms (LFs) generated by the models were scored by String Exact Match (COGS) and Semantic Exact Match\footnote{\begin{footnotesize}Used official scripts

\href{https://github.com/frankaging/ReCOGS/blob/1b6eca8ff4dca5fd2fb284a7d470998af5083beb/utils/train\_utils.py}{https://github.com/frankaging/ReCOGS/blob/
1b6eca8ff4dca5fd2fb284a7d470998af5083beb/utils/
train\_utils.py} 

and \href{https://github.com/frankaging/ReCOGS/blob/1b6eca8ff4dca5fd2fb284a7d470998af5083beb/utils/compgen.py}{https://github.com/frankaging/ReCOGS/blob/
1b6eca8ff4dca5fd2fb284a7d470998af5083beb/
utils/compgen.py}
\end{footnotesize}
} (ReCOGS) against ground truth.

We also measure grammar coverage \citep{fuzzingbook2023:GrammarCoverageFuzzer} (more detail in Appendix \ref{computing_grammar_coverage}) by input examples supported by our RASP model against the full grammar of COGS/ReCOGS input sentences provided in the utilities of the IBM CPG project \citep{klinger2024compositionalprogramgenerationfewshot}\footnote{\begin{footnotesize}https://github.com/IBM/cpg/blob/
c3626b4e03bfc681be2c2a5b23da0b48abe6f570
/src/model/cogs\_data.py\#L523\end{footnotesize}}. See "Appendix: Methods Detail" (\ref{methods_detail}). See also Appendix (\ref{results_notebook_links_by_section}) for notebooks with results and steps to reproduce.

\section{Results}
\textbf{Restricted Access Sequence Processing - grammar coverage using a flat pattern matching approach (not tree-based and not recursive) and autoregressive decoder loop}

\textbf{Figures} \ref{rasp-model-flat-pattern-match-example}, \ref{rasp-model-flat-pattern-match-example-v_dat_p2_match_with_middle_pp}, \ref{rasp-model-flat-pattern-match-example-v_dat_p2_non_matching}, \ref{rasp-for-recogs-decoder-loop-supplementary-figure-incl-encoder-and-decoder-and-grammar-vertical}, \ref{rasp-for-recogs-decoder-loop-supplementary-figure_pp_depth_2}, \ref{rasp-for-recogs-decoder-loop-supplementary-figure_cp_depth_2}, \ref{rasp-for-cogs-encoder-decoder-with-grammar-patterns}, \ref{rasp-for-cogs-decoder-loop-figure_pp_depth_2}, \ref{rasp-for-cogs-decoder-loop_cp_depth_2}, \textbf{Table} \ref{RASP-model-flat-patterns-after-masking-to-nv-relationships-table}. See also Appendix (\ref{grammar_coverage_analysis_for_model_design}) for more details. We selected 19 training-example equivalent sentences (identical to actual training examples after embedding to part-of-speech/verb-type, replacing with such training examples does not change the result) which cover 100\% of the COGS non-recursive input grammar\footnote{ReCOGS has same input grammar/vocabulary as COGS}(lexical differences ignored, under the context-free grammar, tree-based assumption which is violated for our non-tree non-recursive model for prepositional phrases, see below for how that is handled) per \citep{fuzzingbook2023:GrammarCoverageFuzzer}.
Attention-head compatible flat patterns/rules were derived from those examples (Tables \ref{RASP-model-flat-patterns-after-masking-to-nv-relationships-table} , \ref{specific-grammar-pattern-examples-mapped-to-part-of-speech-and-cogs-in-distribution-training-examples})\footnote{\begin{footnotesize}see "Appendix: Restricted Access Sequence Processing word-level token program/model design" (\ref{rasp-word-level-model-design})\end{footnotesize}}.
To handle prepositional phrases (recursive aspect of the input grammar) in a flat solution, we find it necessary on the training data\footnote{\begin{footnotesize}E.g. center embedded pp example "Isabella forwarded a box on a tree to Emma" where masking the center-embedded pp ("on a tree") transforms the sentence to “Isabella forwarded a box \_ to Emma” allows us to have noun-verb relationships extracted the same as other non-pp modified v\_dat\_p1 examples in Tables \ref{RASP-model-flat-patterns-after-masking-to-nv-relationships-table} , \ref{specific-grammar-pattern-examples-mapped-to-part-of-speech-and-cogs-in-distribution-training-examples} by simply mapping the 3 nouns (Isabella, box, Emma) to agent, theme, recipient in order (same for all v\_dat\_p1).\end{footnotesize}} to add a rule that ignores "det common\_noun" or "proper noun" preceded by a preposition when searching for noun indexes to report in relationships (agent, theme, recipient, etc) and as if we did that during pattern matching by using before/after matches instead of strict relative indexing (Figures \ref{rasp-for-recogs-decoder-loop-supplementary-figure_pp_depth_2} (ReCOGS) \ref{rasp-for-cogs-decoder-loop-figure_pp_depth_2} (COGS), which also avoids "attraction errors", Figure \ref{attraction_errors_figure}). Sentential complements (recursive) in training also required masking (Figures \ref{rasp-for-recogs-decoder-loop-supplementary-figure_cp_depth_2}, \ref{rasp-for-cogs-decoder-loop_cp_depth_2}). The resulting 19 matchers implemented with attention-head compatible operations in RASP are given as part of speech / verb type sequences in Table \ref{RASP-model-flat-patterns-after-masking-to-nv-relationships-table} .

\textbf{ Restricted Access Sequence Processing - test set and generalization set performance on (Re)COGS}

See ReCOGS results in Table \ref{results-table} and COGS results in Table \ref{rasp-for-cogs-results-table} .

\textbf{\citep{Wu2023} Encoder-Decoder Transformer from scratch baselines (ReCOGS\_pos)}

See Table \ref{wu2023-results-table}.

\textbf{\citep{Wu2023} Encoder-Decoder baseline 2-layer Transformer does not improve on the obj\_pp\_to\_subj\_pp split when adding 1 or 2 additional layers}
\label{wu-baseline-layer-variation-experiment-results}
\textbf{(even allowing parameter count to increase)}\footnote{\begin{footnotesize}
Since no improvement was observed, we did not run the costly experiments to increase the layers while controlling the parameter count (to distinguish improvements from increasing depth vs merely increasing the parameter count).
\end{footnotesize}
}

See Table \ref{wu2023-results-table}.

\textbf{Attraction Error Analysis for \citep{Wu2023} baseline Encoder-Decoder Transformer on obj\_pp\_to\_subj\_pp split}

(For additional methods detail see Appendix (\ref{appendix_error_analysis_for_baseline_transformer_methods}).) Of the obj\_pp\_to\_subj\_pp split single part errors in single verb sentences made by the \citep{Wu2023} baseline EncDec Transformer where the agent was to the left of the verb\footnote{\begin{footnotesize}Our hypothesis is in terms of nouns with a logical form relationship to a verb or other noun, where the relationship could be agent, theme, or recipient.
Since the obj\_pp\_to\_subj\_pp split is in terms of subject vs object prepositional modification (instead of agent, recipient, or theme), we use the subset of sentences within this split where the agent is to the left of the verb and pp-modified as it corresponds to the subject in that case.
\end{footnotesize}
}, 
across n=10 models, 765 out of 767 (99.74\%; 95\% CI 99.06 to 99.97\%) were in the agent part of the logical form (the predicted position for the error).

\textit{Critically across all n=10 \citep{Wu2023} models, for 96.73\% (740 of the 765 above; 95\% CI (Clopper-Pearson) 95.21 to 97.87\%) of the single point errors in logical forms for single verb sentences where the agent was on the left, modified by a prepositional phrase, and the error was in the agent part, the error in the logical form was that the agent index was accidentally assigned to the specific expected prepositional phrase noun (the one closest to the verb on the left side) instead of the original agent noun.} (Figure \ref{attraction_errors_figure})
This does not vary much from randomly initialized and trained model to model, with the model-level average at 97.07\% of such errors exactly as predicted (stderr=2.23\% (n=10)), with 7 of 10 models having 100\% of these errors exactly as predicted by our hypothesis\footnote{\begin{footnotesize}Fraction for each model as predicted: [0.970, 0.761, 1.0, 
1.0, 1.0, 1.0, 1.0, 1.0, 0.976, 1.0]).
\end{footnotesize}
}.
The attraction to the nearest noun hypothesis predicts that the offset in the agent index varies with prepositional phrase recursion depth (as at depth > 1, there are multiple attractor prepositional nouns to choose from).
\footnote{\begin{footnotesize}See "Appendix: Attraction errors" (\ref{attraction_errors}) for examples of different pp recursion depths.
\end{footnotesize}
}

We report that for all (n=22) single logical form part errors observed (n=10 models, random weight init, random train data order, eval on n=1000 examples, combine errors) where in the input the agent is left of the verb and has a depth=2 prepositional phrase modification in this split, in 100\% (95\% CI (Clopper-Pearson) 84.6 to 100\%; n=22) of those sentences the error in the agent right-index matched our prediction\footnote{\begin{footnotesize}50\% rate expected under null hypothesis of random pp noun substituting as agent rejected with p<0.05 by Clopper-Pearson\end{footnotesize}}.\label{error_analysis_for_baseline_transformer_predict_and_confirm_attraction_errors}

\textbf{\citep{Wu2023} Encoder-Decoder Transformer on new v\_dat\_p2 pp moved to recipient (from theme) split - as hard as hardest previous generalization split}

\textbf{See Figure} \ref{baseline_transformer_standard_training_cannot_do_v_dat_p2_generalization} . Additional methods in Appendix \ref{v_dat_p2_recipient_pp-modification_for_generalization_assessment_and_data_augmentation_attempt}.
\textit{The baseline \citep{Wu2023} EncDec Transformer was only able to achieve a Semantic Exact Match (sample mean +/- sample std) of 13\% +/- 15.6\% (n=10, random initial weights and train data order) with a 95\% sample mean CI of 4\% to 23\%}. Rejecting the null hypothesis of a parse tree-structured representation and instead consistent with the flat/non-hierarchical hypothesis (which does predict any center embedded pp in a novel position will cause errors) this new split v\_dat\_p2\_pp\_moved\_to\_recipient is as difficult or perhaps more difficult than the previous reported "hardest split" obj\_pp\_to\_subj\_pp .

\textbf{\citep{Wu2023} Encoder-Decoder Transformer trained with data augmented with v\_dat\_p2 pp moved to recipient (from theme) does NOT improve obj\_pp\_to\_subj\_pp performance}

\citep{Wu2023}'s baseline EncDec Transformer was trained with default data (ReCOGS\_pos train.tsv) but with additionally the same modified v\_dat\_p2 pp training examples used for the "v\_dat\_p2\_pp\_moved\_to\_recipient" split (non-subject recipient modified with prepositional phrase, so nonoverlapping with subj\_pp, Figure \ref{baseline_transformer_standard_training_cannot_do_v_dat_p2_generalization} (b) and (d)) above on which it performed poorly, then tested on the standard prepositional modification generalization split "obj\_pp\_to\_subj\_pp", after which it achieved 22\% +/- 6.7\% Semantic Exact Match (sample mean +/- std, n=10) with 95\% sample mean CI of 17.9\% to 26.1\% (not significantly different than \citep{Wu2023}'s baseline by one-tailed Welch's unequal variances t-test).

\section{Analysis} 
Our RASP model of a Transformer Encoder-Decoder (EncDec), without using tree-structured representations (nor recursive tree-structured rules like `np\_det pp np -> np\_pp -> np`) attains near perfect scores on the (Re)COGS generalization splits, even the structural generalizations (Tables \ref{results-table} and \ref{rasp-for-cogs-results-table}). This includes prepositional phrase recursion and sentential complement recursion generalization splits up to depth 12 based on rules that can be derived from the depth 2 training examples, without any hardcoded prepositional phrase or sentential complement expansion shortcuts added\footnote{\begin{footnotesize}a single rule applies to all depths; the only limit on length generalization is the RASP interpreter and a simple to extend positional encoding which only handles sentences up to a limit\end{footnotesize}}. 

\textbf{Thus, we demonstrated by construction using the Restricted Access Sequence Processing language which can be compiled to concrete Transformer weights that theoretically a Transformer EncDec can solve the COGS input to COGS/ReCOGS\_pos logical form translation in a systematic, compositional, and length generalizing way.}

Recall we found a single flat masking rule we originally added to fit center embedded pp training examples, to ignore "pp det common\_noun" and "pp proper\_noun" when matching nouns for the agent, theme, recipient right indices, was sufficient to avoid structural generalization errors due to pp modification in novel positions. 

Interestingly, we imagined ablating that single rule and hypothesized attraction to the nearest noun (Figure \ref{attraction_errors_figure} and "Appendix: Attraction Errors" (\ref{attraction_errors})) in its absence and found this predicted the exact error (the nearest noun to the verb on the expected side is mistaken for the agent of the verb) 
in 96\% of the single relationship errors the \citep{Wu2023} baseline Transformers make on the obj-pp-to-subj-pp split\footnote{overall, their semantic exact match on the split is measured by us at 19.7\%, consistent with their Figure 5} when the agent is left of the verb in single verb sentences (suggesting perhaps that the baseline \citep{Wu2023} Transformer trained from scratch is also not learning a hierarchical, tree-structured representation.)\footnote{\begin{footnotesize}\citep{li2023slogstructuralgeneralizationbenchmark} also observe this stating "For instance, in sentences like 'A cat on the mat froze', models often misinterpret the closer NP the mat as the subject."\end{footnotesize}}
Our explanation could have been refuted by other single relationship errors occurring as frequently as the agent, indicating general model confusion (independently getting incorrect agent and theme, not just agent relationships) and/or when making an agent error, the model could have simply put nonsense indices or referred to any other position other than the closest pp noun position to the verb (which does vary and depends on pp depth) to refute our hypothesis.

\begin{figure}
\includegraphics[scale=0.30]{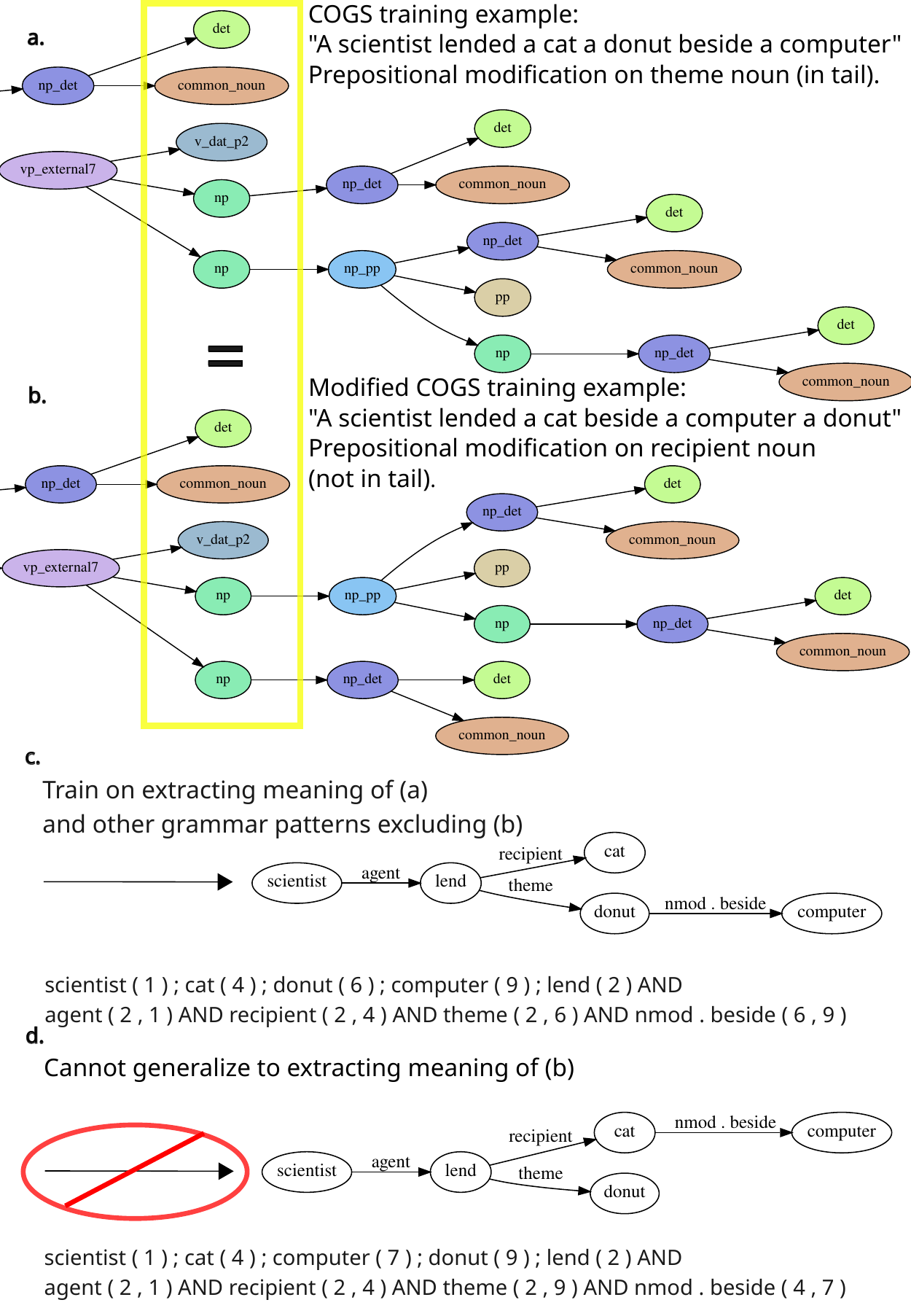}
\caption[\citep{Wu2023} baseline Encoder-Decoder Transformer trained from scratch generalizing to new v\_dat\_p2 pp moved to recipient (from theme) split is as hard as the previously reported hardest generalization split as predicted by the flat/non-recursive/non-tree hypothesis by RASP modeling]{\begin{small}\citep{Wu2023} Encoder-Decoder Transformer trained from scratch generalizing to new v\_dat\_p2 pp moved to recipient (from theme) split is as hard as the previously reported hardest generalization split consistent with the flat/non-recursive/non-tree representation hypothesis ((d) rejects $H_0$). Figure \ref{rasp-model-flat-pattern-match-example-v_dat_p2_match_with_middle_pp} shows how a flat RASP model can recognize (b).\end{small} 
\label{baseline_transformer_standard_training_cannot_do_v_dat_p2_generalization}
}
\end{figure}

The flat/non-tree structured hypothesis for the baseline Transformer can also be checked by making a prediction on a completely different syntax, given we predict any novel center pp substitution will be challenging (not about subjects vs objects): the `np v\_dat\_p2 np pp np np` prepositional phrase modification (where the recipient relationship is modified instead of the subject/agent so is a distinct check of our hypothesis)\footnote{\begin{footnotesize}See Appendix \ref{v_dat_p2_recipient_pp-modification_for_generalization_assessment_and_data_augmentation_attempt}\end{footnotesize}} and we indeed found that this was at least as hard as the most difficult split previously reported, the `obj\_pp\_to\_subj\_pp` split (Figure \ref{baseline_transformer_standard_training_cannot_do_v_dat_p2_generalization}).

Maybe \citep{Wu2023} baseline EncDec is depth-constrained to learn a non-hierarchical, non-tree model with these characteristic errors but with more layers it would learn to recursively combine `np\_det pp np -> np\_pp -> np` (to some fixed depth, limited by the number of Transformer blocks) and perform better on pp related splits\footnote{\begin{footnotesize}\citep{Csordas2022}: "the network should be sufficiently deep, at least as deep as the deepest data dependency in the computational graph built from elementary operations (e.g., in the case of a parse tree, this is the depth of the tree)". But the hierarchical approach may not be a scalable approach, as we must make the network ever deeper to handle increased pp depths instead of just looping (without weight-sharing), \textbf{and deeper parses encountered at inference time than seen in training would activate untrained layers, leading to generalization failures. As shown by the RASP model, flat, but with pp masking} (Figure \ref{rasp-for-recogs-decoder-loop-supplementary-figure_pp_depth_2} (ReCOGS) \ref{rasp-for-cogs-decoder-loop-figure_pp_depth_2} (COGS)), \textbf{may be a better strategy.}\end{footnotesize}}. However, training a \citep{Wu2023} baseline EncDec Transformer from scratch we found no benefit to 3-4 layers instead of 2 on the ReCOGS obj\_pp\_to\_subj\_pp split, consistent with \citep{petty2024impactdepthcompositionalgeneralization}'s finding on COGS.

These results suggest we may interpret the poor performance on generalizing on unseen pp related modification related splits as arising from the baseline 2-4 layer EncDec Transformers learning a non-tree representation that cannot leverage the recursive grammar rule `np\_det pp np -> np\_pp -> np` during learning, thus requiring them to actually observe more of the various pp substitutions to learn them\footnote{\begin{footnotesize}See "Appendix: Composition and Learning" (\ref{composition_and_learning})\end{footnotesize}} (if pp masking as used by our RASP model is not learned instead). \section{Conclusion} Using Restricted Access Sequence Processing immediately helped us discover additional failure modes (e.g. new "v\_dat\_p2\_pp\_moved\_to\_recipient" split\footnote{\begin{footnotesize}After this paper was written we found our predicted split "v\_dat\_p2\_pp\_moved\_to\_recipient" has also been added to an extended separate (SLOG) version of COGS in \citep{li2023slogstructuralgeneralizationbenchmark}, sec 2.2.1 indirect object mod (4), confirmed difficult
\end{footnotesize}
}) of \citep{Wu2023}'s baseline EncDec Transformer, predict logical form errors in detail, and demonstrated that tree-structured representations are not needed to cover the recursive aspects of the (Re)COGS grammar. We recommend RASP for understanding Transformer behavior even for tasks as complex as COGS. We predict that Transformers should be able to perform the (Re)COGS tasks accurately for novel prepositional phrase substitution positions or structural recursion depths unseen in training and that the problem is just of getting the Transformer to learn the appropriate rules\footnote{\begin{footnotesize}e.g. to ignore "pp det common\_noun" and "pp proper\_noun" when finding nouns in relationships with verbs, which allows our RASP models to attain near perfect scores on (Re)COGS structural and depth generalizations\end{footnotesize}}, turning attention to data augmentation\footnote{\begin{footnotesize}We tried one augmentation (A\ref{v_dat_p2_recipient_pp-modification_for_generalization_assessment_and_data_augmentation_attempt}) out of many possible.
\end{footnotesize}}\citep{qiu-etal-2022-improving}\footnote{contrasting \citep{qiu-etal-2022-improving}, the RASP-for-(Re)COGS flat solutions generalize to arbitrary depth (not just the depth of recursion demonstrated in augmented data)}, curriculum learning \citep{10.1145/1553374.1553380}, reinforcement learning \citep{Ranzato2015}, training objectives \citep{10.1162/tacl_a_00733}, and other approaches \citep{Csordas2021} \citep{ontanon-etal-2022-making}.

\clearpage
\section*{Known Project Limitations}

The Restricted Access Sequence Processing code is not optimized\footnote{\begin{footnotesize}A slow RASP program does NOT necessarily imply the equivalent Transformer would be slow.\end{footnotesize}}. Cannot yet predict attention heads and layers required from the select and aggregate operations performed like the RASP authors \citep{Weiss2021} were able to do with their problems.

Grammar coverage \citep{fuzzingbook2023:GrammarCoverageFuzzer} is only valid when the expansions are rules your model can learn.\footnote{\begin{footnotesize}If for example, as with our flat RASP model by design or as we hypothesize for \citep{Wu2023}'s baseline Encoder-Decoder Transformer, the model cannot or will not learn the rule `np\_det pp np -> np\_pp -> np` which recursively replaces noun phrases modified by prepositional phrases with a noun phrase, then grammar coverage will assume any prepositional phrase exposure is sufficient, which is evidently not true given the errors on prepositional phrase modification generalization splits reported here and by \citep{Wu2023}, \citep{KimLinzen2020}.
\end{footnotesize}
} We specifically made use of this limitation in this paper but still caution anyone about it who might just take the grammar coverage metric away by itself.

The attraction error analysis of the \citep{Wu2023} baseline Encoder-Decoder Transformer on the obj\_pp\_to\_subj\_pp split does not yet attempt to explain the common case of multiple errors in the logical form.\footnote{\begin{footnotesize}e.g. agent index may be replaced by prepositional phrase noun but also a spurious theme relationship is added or the theme index is also corrupted\end{footnotesize}}

Much deeper Transformer networks may be learning a tree-based grammar representation\footnote{\begin{footnotesize}Nothing explored here rules that out and there is plenty of evidence outside the COGS task-related literature suggesting this will be the case: \citep{tenney2019bertrediscoversclassicalnlp} show the 24-layer BERT model seems to handle "POS tagging, parsing, NER, semantic roles, then coreference"; \citep{hewitt-manning-2019-structural} "provid[e] evidence that entire syntax trees are embedded implicitly in deep models’ [including BERT's] vector geometry", and \citep{goldberg2019assessingbertssyntacticabilities} shows BERT excels at subject-verb agreement, "which [is] traditionally taken as evidence for
the existence [of] hierarchical structure" (though e.g. in this work we see that ignoring distractor nouns in long-term dependencies does not require hierarchy or a deep understanding of syntax but simple rules like ignore "pp det common\_noun" and "pp proper\_noun" for finding noun-verb relationships in the logical form can allow for handling of such long-range dependencies). On the other hand, \citep{petty2024impactdepthcompositionalgeneralization} argue specifically for the COGS benchmark (semantically equivalent to ReCOGS which is derived from it) that increasing depth does not allow their Transformers to make progress on the structural generalization splits, even at depths up to 32 layers.
\end{footnotesize}
} and not suffer from the predicted generalization issues observed in \citep{Wu2023}'s baseline 2-layer Transformer and predicted by our intentionally non-tree RASP model (if compensating rules to avoid attraction errors in a flat model are not also learned).

As stated elsewhere in the paper, we are focused on the structural generalizations that Transformers currently struggle with (prepositional phrase recursion (pp recursion), sentential complement (cp) recursion, object-pp to subj-pp modification generalization (obj-pp-to-subj-pp)) and not lexical generalizations (which Transformers are demonstrated to do well on by others). Thus, as noted elsewhere, our model assumes the embedding can map words to their possible part-of-speech and verb-type and does not address the learnability of this mapping (number of exposures required, etc).

The (Re)COGS tasks are English specific and our findings on compositional generalization may not necessarily apply (or apply differently) to other languages. Our description of "attraction errors" for example does not appear specific to subject-verb-object ordering in English but a similar analysis should be conducted in other languages and we used the obj-pp-to-subj-pp split in our detailed error analysis which is subject-verb-object ordering sensitive.

The RASP-for-(Re)COGS models applied to unintended use WILL give invalid results or halt - we have NOT provided a general language model, we have provided a simulation of how a Transformer could perform a specific task.
The RASP-for-(Re)COGS models/simulations as provided are for research purposes only to prove feasibility of the (Re)COGS tasks by Transformers and is not appropriate for ANY other uses whatsoever without modification. For one, an actual Transformer performing the equivalent operations would run orders of magnitude faster, which should be reason enough to not want to use the RASP simulation for actual input-output tasks outside of a research setting. However, there is also no "tokenizer" provided for the RASP model to handle out-of-vocabulary inputs and fallback paths for out-of-grammar examples are not provided so the RASP model will halt on most inputs and can only run on the in-distribution (non-augmented) training data, and the dev, test, and gen sets of (Re)COGS, though such aspects could be added. We provide the code for reproducing the results of this study and for researchers who are capable of writing RASP themselves to build upon the work and/or more easily apply RASP to their own problems given our examples, not for immediate application to any other tasks without appropriate modification.

See also "Appendix: Potential Risks" (\ref{potential_risks}) and "Appendix: Scientific Artifacts - Is Our Use Consistent with Authors' Intention and Licensing" (\ref{scientific_artifacts_use}) for other potential concerns.

\bibliography{bibliography}
\bibliographystyle{plainnat}

\section*{Notes}
No AI tools were used by the author in the preparation of this manuscript with the exception of anything used in the backend by Google Scholar searches for literature and citations and Google searches for related material. AI writing aids were not used.

\section{Appendix}

\twocolumn
\begin{table}
\centering
\begin{tabular}{p{0.6\linewidth} p{0.3\linewidth}}
\hline
\begin{tiny}\textbf{ReCOGS\_pos Split}\end{tiny} & \begin{tiny}\textbf{Semantic Exact Match \% (95\% CI)}\end{tiny} \\
\hline
\begin{tiny}ReCOGS\_pos test set (held out, in-distribution)\end{tiny} & \begin{tiny} 100.00\% (99.88-100.00\%)\end{tiny} \\
\hline
\begin{tiny}Generalization splits (held out, out-of-distribution)\end{tiny} & \\
\hline
\begin{tiny}active\_to\_passive\end{tiny} & \begin{tiny}100.00\% (99.63-100.00\%)\end{tiny} \\
\begin{tiny}do\_dative\_to\_pp\_dative\end{tiny} & \begin{tiny} 100.00\% (99.63-100.00\%)\end{tiny} \\
\begin{tiny}obj\_omitted\_transitive\_to\_transitive\end{tiny} & \begin{tiny} 100.00\% (99.63-100.00\%)\end{tiny} \\
\begin{tiny}\textbf{obj\_pp\_to\_subj\_pp}\end{tiny} & \begin{tiny} \textbf{92.20\% (90.36- 93.79\%)}\end{tiny} \\
\begin{tiny}obj\_to\_subj\_common\end{tiny} & \begin{tiny} 100.00\% (99.63-100.00\%)\end{tiny} \\
\begin{tiny}obj\_to\_subj\_proper\end{tiny} & \begin{tiny} 100.00\% (99.63-100.00\%)\end{tiny} \\
\begin{tiny}only\_seen\_as\_transitive\_subj\_as\_unacc\_subj\end{tiny} & \begin{tiny} 100.00\% (99.63-100.00\%)\end{tiny} \\
\begin{tiny}only\_seen\_as\_unacc\_subj\_as\_obj\_omitted\_transitive\_subj\end{tiny} & \begin{tiny} 100.00\% (99.63-100.00\%)\end{tiny} \\
\begin{tiny}only\_seen\_as\_unacc\_subj\_as\_unerg\_subj\end{tiny} & \begin{tiny} 100.00\% (99.63-100.00\%)\end{tiny} \\
\begin{tiny}passive\_to\_active\end{tiny} & \begin{tiny} 100.00\% (99.63-100.00\%)\end{tiny} \\
\begin{tiny}pp\_dative\_to\_do\_dative\end{tiny} & \begin{tiny} 100.00\% (99.63-100.00\%)\end{tiny} \\
\begin{tiny}prim\_to\_inf\_arg\end{tiny} & \begin{tiny} 100.00\% (99.63-100.00\%)\end{tiny} \\
\begin{tiny}prim\_to\_obj\_common\end{tiny} & \begin{tiny} 100.00\% (99.63-100.00\%)\end{tiny} \\
\begin{tiny}prim\_to\_obj\_proper\end{tiny} & \begin{tiny} 100.00\% (99.63-100.00\%)\end{tiny} \\
\begin{tiny}prim\_to\_subj\_common\end{tiny} & \begin{tiny} 100.00\% (99.63-100.00\%)\end{tiny} \\
\begin{tiny}prim\_to\_subj\_proper\end{tiny} & \begin{tiny} 100.00\% (99.63-100.00\%)\end{tiny} \\
\begin{tiny}subj\_to\_obj\_common\end{tiny} & \begin{tiny} 100.00\% (99.63-100.00\%)\end{tiny} \\
\begin{tiny}subj\_to\_obj\_proper\end{tiny} & \begin{tiny} 100.00\% (99.63-100.00\%)\end{tiny} \\
\begin{tiny}unacc\_to\_transitive\end{tiny} & \begin{tiny} 100.00\% (99.63-100.00\%)\end{tiny} \\
\begin{tiny}\textbf{pp\_recursion}\end{tiny} & \begin{tiny} \textbf{100.00\% (99.63-100.00\%)}\end{tiny} \\
\begin{tiny}\textbf{cp\_recursion}\end{tiny} & \begin{tiny} \textbf{100.00\% (99.63-100.00\%)}\end{tiny} \\
\hline
\begin{tiny}all gen splits (21K examples, aggregate)\end{tiny} & \begin{tiny}99.63\% (99.54-99.71\%)\end{tiny} \\
\hline
\end{tabular}
\caption{\begin{small}\textbf{RASP-for-ReCOGS Restricted Access Sequence Processing (RASP) Encoder-Decoder Transformer-compatible model} performance on the ReCOGS\_pos test set (n=3000) and out-of-distribution generalization split performance (n=1000 per gen split). Model is deterministic, so Clopper-Pearson CIs are used. \textbf{Structural generalization splits are bolded.} (Model was developed on training data or examples identical to training examples after embedding words to part of speech and verb type sequences, the first step in the model, see Tables \ref{RASP-model-flat-patterns-after-masking-to-nv-relationships-table} , \ref{specific-grammar-pattern-examples-mapped-to-part-of-speech-and-cogs-in-distribution-training-examples}). 100.00\% (CI 99.63-100.00\%) string exact match was also attained for recursion splits.
\end{small}}
\label{results-table-full}
\end{table}

\begin{table}
\centering
\begin{tabular}{p{0.6\linewidth} p{0.3\linewidth}}
\hline
\begin{tiny}\textbf{COGS Split}\end{tiny} & \begin{tiny}\textbf{String Exact Match \% (95\% CI)}\end{tiny} \\
\hline
\begin{tiny}COGS test set (held out, in-distribution) (n=3000)\end{tiny} & \begin{tiny} 99.97\% (99.81-99.99\%)\end{tiny} \\
\hline
\begin{tiny}Generalization splits (held out, out-of-distribution) \end{tiny} & \begin{tiny} (n=1000 each) \end{tiny}  \\
\hline
\begin{tiny}active\_to\_passive\end{tiny} & \begin{tiny}100.00\% (99.63-100.00\%)\end{tiny} \\
\begin{tiny}do\_dative\_to\_pp\_dative\end{tiny} & \begin{tiny} 99.90\% (99.44-99.997\%)\end{tiny} \\
\begin{tiny}obj\_omitted\_transitive\_to\_transitive\end{tiny} & \begin{tiny} 100.00\% (99.63-100.00\%)\end{tiny} \\
\begin{tiny}\textbf{obj\_pp\_to\_subj\_pp}\end{tiny} & \begin{tiny}\textbf{ 100.00\% (99.63-100.00\%)}\end{tiny} \\
\begin{tiny}obj\_to\_subj\_common\end{tiny} & \begin{tiny} 100.00\% (99.63-100.00\%)\end{tiny} \\
\begin{tiny}obj\_to\_subj\_proper\end{tiny} & \begin{tiny} 99.90\% (99.44-99.997\%)\end{tiny} \\
\begin{tiny}only\_seen\_as\_transitive\_subj\_as\_unacc\_subj\end{tiny} & \begin{tiny} 100.00\% (99.63-100.00\%)\end{tiny} \\
\begin{tiny}only\_seen\_as\_unacc\_subj\_as\_obj\_omitted\_transitive\_subj\end{tiny} & \begin{tiny} 100.00\% (99.63-100.00\%)\end{tiny} \\
\begin{tiny}only\_seen\_as\_unacc\_subj\_as\_unerg\_subj\end{tiny} & \begin{tiny} 100.00\% (99.63-100.00\%)\end{tiny} \\
\begin{tiny}passive\_to\_active\end{tiny} & \begin{tiny} 100.00\% (99.63-100.00\%)\end{tiny} \\
\begin{tiny}pp\_dative\_to\_do\_dative\end{tiny} & \begin{tiny} 99.90\% (99.44-99.997\%)\end{tiny} \\
\begin{tiny}prim\_to\_inf\_arg\end{tiny} & \begin{tiny} 100.00\% (99.63-100.00\%)\end{tiny} \\
\begin{tiny}prim\_to\_obj\_common\end{tiny} & \begin{tiny} 99.80\% (99.28-99.98\%)\end{tiny} \\
\begin{tiny}prim\_to\_obj\_proper\end{tiny} & \begin{tiny} 100.00\% (99.63-100.00\%)\end{tiny} \\
\begin{tiny}prim\_to\_subj\_common\end{tiny} & \begin{tiny} 100.00\% (99.63-100.00\%)\end{tiny} \\
\begin{tiny}prim\_to\_subj\_proper\end{tiny} & \begin{tiny} 100.00\% (99.63-100.00\%)\end{tiny} \\
\begin{tiny}subj\_to\_obj\_common\end{tiny} & \begin{tiny} 99.80\% (99.28-99.98\%)\end{tiny} \\
\begin{tiny}subj\_to\_obj\_proper\end{tiny} & \begin{tiny} 100.00\% (99.63-100.00\%)\end{tiny} \\
\begin{tiny}unacc\_to\_transitive\end{tiny} & \begin{tiny} 100.00\% (99.63-100.00\%)\end{tiny} \\
\begin{tiny}\textbf{pp\_recursion}\end{tiny} & \begin{tiny} \textbf{98.40\% (97.41-99.08\%)}\end{tiny} \\
\begin{tiny}\textbf{cp\_recursion}\end{tiny} & \begin{tiny} \textbf{99.90\% (99.44-99.997\%)}\end{tiny} \\
\hline
\begin{tiny}all gen splits (21K examples, aggregate)\end{tiny} & \begin{tiny} 99.89\% (99.83-99.93\%)\end{tiny} \\
\hline
\end{tabular}
\caption{\begin{small}\textbf{RASP-for-COGS Restricted Access Sequence Processing (RASP) Encoder-Decoder Transformer-compatible model} COGS test set performance (n=3000) and out-of-distribution generalization split performance (n=1000 per gen split). With Beta/Clopper-Pearson confidence intervals. N=1000 examples for each generalization split. No examples excluded. \textbf{Structural generalization splits are bolded.} (Model was developed on training data or examples identical to training examples after embedding words to part of speech and verb type sequences, the first step in the model, see Tables \ref{RASP-model-flat-patterns-after-masking-to-nv-relationships-table} , \ref{specific-grammar-pattern-examples-mapped-to-part-of-speech-and-cogs-in-distribution-training-examples}). Rare examples with out-of-vocabulary (vs train.tsv) words marked wrong by model automatically (no attempted lexical generalizations to OOV).\end{small}}
\label{rasp-for-cogs-results-table-full}
\end{table}

\clearpage
\onecolumn
\subsection{Results Notebook links by section}
\label{results_notebook_links_by_section}

\subsubsection{ReCOGS RASP model on test set}
For steps to reproduce and results, see \href{https://github.com/willy-b/RASP-for-ReCOGS/blob/main/supplemental\_data/RASP\_model\_for\_ReCOGS\_eval\_test\_set\_(multiday\_run\_on\_dedicated\_VM)\_(PR\_7\_contents\_on\_TEST\_set\_incl\_complement\_phrase\_support)\_(public).ipynb}{the RASP model ReCOGS test set notebook (link)}\footnote{\begin{tiny}Full URL for printed copies: 

\href{https://github.com/willy-b/RASP-for-ReCOGS/blob/main/supplemental\_data/RASP\_model\_for\_ReCOGS\_eval\_test\_set\_(multiday\_run\_on\_dedicated\_VM)\_(PR\_7\_contents\_on\_TEST\_set\_incl\_complement\_phrase\_support)\_(public).ipynb}{https://github.com/willy-b/RASP-for-ReCOGS/blob/main/supplemental\_data/RASP\_model\_for\_ReCOGS\_eval\_test\_set\_(multiday}

\href{https://github.com/willy-b/RASP-for-ReCOGS/blob/main/supplemental\_data/RASP\_model\_for\_ReCOGS\_eval\_test\_set\_(multiday\_run\_on\_dedicated\_VM)\_(PR\_7\_contents\_on\_TEST\_set\_incl\_complement\_phrase\_support)\_(public).ipynb}{\_run\_on\_dedicated\_VM)\_(PR\_7\_contents\_on\_TEST\_set\_incl\_complement\_phrase\_support)\_(public).ipynb}\end{tiny}}.

\subsubsection{ReCOGS RASP model on generalization set (all splits)}
For steps to reproduce and results, see 
\href{https://github.com/willy-b/RASP-for-ReCOGS/blob/main/supplemental\_data/RASP\_model\_for\_ReCOGS\_eval\_on\_gen\_set\_(multiday\_run\_on\_dedicated\_VM)\_(PR\_7\_contents\_on\_GEN\_set\_incl\_complement\_phrase\_support)\_(public).ipynb}{the RASP model ReCOGS generalization set notebook (link)}\footnote{\begin{tiny}Full URL for printed copies: 

\href{https://github.com/willy-b/RASP-for-ReCOGS/blob/main/supplemental\_data/RASP\_model\_for\_ReCOGS\_eval\_on\_gen\_set\_(multiday\_run\_on\_dedicated\_VM)\_(PR\_7\_contents\_on\_GEN\_set\_incl\_complement\_phrase\_support)\_(public).ipynb}{https://github.com/willy-b/RASP-for-ReCOGS/blob/main/supplemental\_data/RASP\_model\_for\_ReCOGS\_eval\_on\_gen\_set\_(multiday}

\href{https://github.com/willy-b/RASP-for-ReCOGS/blob/main/supplemental\_data/RASP\_model\_for\_ReCOGS\_eval\_on\_gen\_set\_(multiday\_run\_on\_dedicated\_VM)\_(PR\_7\_contents\_on\_GEN\_set\_incl\_complement\_phrase\_support)\_(public).ipynb}{\_run\_on\_dedicated\_VM)\_(PR\_7\_contents\_on\_GEN\_set\_incl\_complement\_phrase\_support)\_(public).ipynb}\end{tiny}}.

\subsubsection{RASP-for-COGS}

For steps to reproduce and results on the COGS test set and generalization splits, see \href{https://github.com/willy-b/RASP-for-COGS}{the RASP-for-COGS repository}\footnote{\begin{tiny}Full URL for printed copies: \href{https://github.com/willy-b/RASP-for-COGS}{https://github.com/willy-b/RASP-for-COGS}\end{tiny}}.

\subsubsection{\citep{Wu2023} Encoder-Decoder Transformer from scratch baselines (ReCOGS\_pos)}

See \href{https://github.com/willy-b/RASP-for-ReCOGS/blob/main/supplemental\_data/RASP\_for\_ReCOGS\_(no\_RASP\_in\_this\_file)\_more\_Wu\_et\_al\_2023\_transformer\_baselines\_to\_compare\_with\_Restricted\_Access\_Sequence\_Processing\_(\_use\_fixed\_positional\_indices)\_and\_or\_data\_augmentation.ipynb}{the Encoder-Decoder Transformer from scratch baselines notebook (link)}\footnote{\begin{tiny}Full URL for printed copies: 

\href{https://github.com/willy-b/RASP-for-ReCOGS/blob/main/supplemental\_data/RASP\_for\_ReCOGS\_(no\_RASP\_in\_this\_file)\_more\_Wu\_et\_al\_2023\_transformer\_baselines\_to\_compare\_with\_Restricted\_Access\_Sequence\_Processing\_(\_use\_fixed\_positional\_indices)\_and\_or\_data\_augmentation.ipynb}{https://github.com/willy-b/RASP-for-ReCOGS/blob/main/supplemental\_data/RASP\_for\_ReCOGS\_(no\_RASP\_in\_this\_file)}

\href{https://github.com/willy-b/RASP-for-ReCOGS/blob/main/supplemental\_data/RASP\_for\_ReCOGS\_(no\_RASP\_in\_this\_file)\_more\_Wu\_et\_al\_2023\_transformer\_baselines\_to\_compare\_with\_Restricted\_Access\_Sequence\_Processing\_(\_use\_fixed\_positional\_indices)\_and\_or\_data\_augmentation.ipynb}{\_more\_Wu\_et\_al\_2023\_transformer\_baselines\_to\_compare\_with\_Restricted\_Access\_Sequence\_Processing\_(\_use\_fixed\_positional\_indices)\_and\_or\_data\_augmentation.ipynb}\end{tiny}}

for \citep{Wu2023} script execution and analysis code.

\subsubsection{\citep{Wu2023} Encoder-Decoder baseline 2-layer Transformer does not improve on the obj\_pp\_to\_subj\_pp split when adding 1 or 2 additional layers}

3 and 4 layer results can also be found in that same notebook, which is also the

\href{https://github.com/willy-b/RASP-for-ReCOGS/blob/main/supplemental\_data/RASP\_for\_ReCOGS\_(no\_RASP\_in\_this\_file)\_more\_Wu\_et\_al\_2023\_transformer\_baselines\_to\_compare\_with\_Restricted\_Access\_Sequence\_Processing\_(\_use\_fixed\_positional\_indices)\_and\_or\_data\_augmentation.ipynb}{the baseline Encoder-Decoder Transformer 3 or 4 layer variation results notebook (link)}\footnote{\begin{tiny}Full URL for printed copies: 

\href{https://github.com/willy-b/RASP-for-ReCOGS/blob/main/supplemental\_data/RASP\_for\_ReCOGS\_(no\_RASP\_in\_this\_file)\_more\_Wu\_et\_al\_2023\_transformer\_baselines\_to\_compare\_with\_Restricted\_Access\_Sequence\_Processing\_(\_use\_fixed\_positional\_indices)\_and\_or\_data\_augmentation.ipynb}{https://github.com/willy-b/RASP-for-ReCOGS/blob/main/supplemental\_data/RASP\_for\_ReCOGS\_(no\_RASP\_in\_this\_file)}

\href{https://github.com/willy-b/RASP-for-ReCOGS/blob/main/supplemental\_data/RASP\_for\_ReCOGS\_(no\_RASP\_in\_this\_file)\_more\_Wu\_et\_al\_2023\_transformer\_baselines\_to\_compare\_with\_Restricted\_Access\_Sequence\_Processing\_(\_use\_fixed\_positional\_indices)\_and\_or\_data\_augmentation.ipynb}{\_more\_Wu\_et\_al\_2023\_transformer\_baselines\_to\_compare\_with\_Restricted\_Access\_Sequence\_Processing\_(\_use\_fixed\_positional\_indices)\_and\_or\_data\_augmentation.ipynb}\end{tiny}} (scroll down).

\subsubsection{Attraction Error Analysis for \citep{Wu2023} baseline Encoder-Decoder Transformer on obj\_pp\_to\_subj\_pp split}

See \href{https://github.com/willy-b/RASP-for-ReCOGS/blob/main/supplemental\_data/ReCOGS\_Baseline\_non\_RASP\_Transformer\_ReCOGS\_error\_prediction\_with\_n\%3D10\_Transformers\_trained\_from\_scratch\_(predicting\_the\_details\_of\_error\_in\_logical\_form\_on\_obj\_pp\_to\_subj\_pp\_split).ipynb}{Attraction Error Analysis on baseline Encoder-Decoder Transformer notebook (link)}\footnote{\begin{tiny}Full URL for printed copies: 

\href{https://github.com/willy-b/RASP-for-ReCOGS/blob/main/supplemental\_data/ReCOGS\_Baseline\_non\_RASP\_Transformer\_ReCOGS\_error\_prediction\_with\_n\%3D10\_Transformers\_trained\_from\_scratch\_(predicting\_the\_details\_of\_error\_in\_logical\_form\_on\_obj\_pp\_to\_subj\_pp\_split).ipynb}{https://github.com/willy-b/RASP-for-ReCOGS/blob/main/supplemental\_data/ReCOGS\_Baseline\_non\_RASP\_Transformer\_ReCOGS\_error\_prediction
\_with\_n\%3D10\_Transformers\_trained\_from\_scratch\_(predicting\_the\_details\_of\_error\_in\_logical\_form\_on\_obj\_pp\_to\_subj\_pp\_split).ipynb}\end{tiny}}.

\subsubsection{\citep{Wu2023} baseline Encoder-Decoder Transformer on v\_dat\_p2 generalization to recipient pp modification after training on theme pp modification (both right of verb)}

See \href{https://github.com/willy-b/RASP-for-ReCOGS/blob/main/supplemental\_data/train\_ReCOGS\_baseline\_Transformer\_(from\_Wu\_et\_al\_2023)\_on\_Wus\_training\_set\_which\_only\_has\_v\_dat\_p2\_pp\_modification\_on\_the\_theme\_(right\_most\_np),\_test\_generalization\_on\_recipient\_modification\_(left\_in\_np\_pair\_both\_right.ipynb}{v\_dat\_p2 generalization to middle np/recipient pp modification on baseline Encoder-Decoder Transformer notebook (link)}\footnote{\begin{tiny}Full URL for printed copies: 

\href{https://github.com/willy-b/RASP-for-ReCOGS/blob/main/supplemental\_data/train\_ReCOGS\_baseline\_Transformer\_(from\_Wu\_et\_al\_2023)\_on\_Wus\_training\_set\_which\_only\_has\_v\_dat\_p2\_pp\_modification\_on\_the\_theme\_(right\_most\_np),\_test\_generalization\_on\_recipient\_modification\_(left\_in\_np\_pair\_both\_right.ipynb}{https://github.com/willy-b/RASP-for-ReCOGS/blob/main/supplemental\_data/
train\_ReCOGS\_baseline\_Transformer\_(from\_Wu\_et\_al\_2023)
\_on\_Wus\_training\_set\_which\_only\_has\_v\_dat\_p2\_pp\_modification
\_on\_the\_theme\_(right\_most\_np),\_test\_generalization\_on\_recipient\_modification
\_(left\_in\_np\_pair\_both\_right.ipynb}
\end{tiny}} .

\twocolumn

\clearpage

\subsection{Restricted Access Sequence Processing word-level token program/model design}
\label{rasp-word-level-model-design}

For overall model descriptions, first see Model Detail \ref{model_detail}.
For a description of the Restricted Access Sequence Processing (RASP) language, which is theoretically compilable to Transformer neural network weights, see \citep{Weiss2021}.

You can run a demo of our RASP model and see the autoregressive output using the instructions below.

For RASP-for-COGS:
\begin{tiny}
\begin{verbatim}
git clone https://github.com/willy-b/RASP-for-COGS.git
cd RASP-for-COGS
python cogs_examples_in_rasp.py
\end{verbatim}
\end{tiny}

The script will show performance on COGS training data by default, run with "--use\_dev\_split", "--use\_gen\_split" , or "--use\_test\_split" to see it run on those and give a running score every row.

And for RASP-for-ReCOGS:
\begin{tiny}
\begin{verbatim}
git clone https://github.com/willy-b/learning-rasp.git
cd learning-rasp
python recogs_examples_in_rasp.py
\end{verbatim}
\end{tiny}

The script will show performance on \citep{Wu2023} ReCOGS\_pos training data by default, run with "--use\_dev\_split", "--use\_gen\_split" , or "--use\_test\_split" to see it run on those and give a running score every row.

For ReCOGS, intending to perform well on Semantic Exact Match, we took a simple, flat, non-tree approach masking out recursive parts of the grammar which was able to get 100\% semantic exact match (and string exact match) on the full test set, and 99.6\% semantic exact match on the out-of-distribution generalization set of the real ReCOGS dataset\footnote{\begin{footnotesize}word-level token Restricted Access Sequence Processing solution: \href{https://github.com/willy-b/RASP-for-COGS}{https://github.com/willy-b/RASP-for-COGS} (COGS) and \href{https://github.com/willy-b/learning-rasp}{https://github.com/willy-b/learning-rasp} (ReCOGS)\end{footnotesize}}.

We use word-level tokens for all RASP model results in this paper.\footnote{\begin{footnotesize}We believe any solution at the word-level can be converted to a character-level token solution and that is not the focus of our investigation here (see Appendix \ref{rasp_character_level_model_notes} for proof of concept details on a character level solution not used here).\end{footnotesize}}. We took the RASP native sequence tokens at the word-level, and first did a Transformer learned-embedding compatible operation and created 1 part-of-speech and 4 extra verb-type sequences (because each word in the COGS vocabulary may actually serve multiple POS roles; up to four different verb types as in the case of "liked" 

which can serve as v\_trans\_not\_omissible, v\_trans\_not\_omissible\_pp\_p1, v\_trans\_not\_omissible\_pp\_p2, and v\_cp\_taking types). 

The five extra sequences serve to associate each word with one or more of the following part-of-speech/verb type roles:
\begin{tiny}
\begin{verbatim}
det: 1
pp: 2
was: 3
by: 4
to: 5
that: 6
common_noun: 7
proper_noun: 8
v_trans_omissible: 9
v_trans_omissible_pp: 10
v_trans_not_omissible: 11
v_trans_not_omissible_pp: 12
v_cp_taking: 13
v_inf_taking: 14
v_unacc: 15
v_unerg: 16
v_inf: 17
v_dat: 18
v_dat_pp: 19
v_unacc_pp: 20
# v_normalized_in_output: 21 # only used in decoder loop 
#, represents stemmed verbs where type is not important
\end{verbatim}
\end{tiny}

\begin{figure}
\includegraphics[scale=0.73]{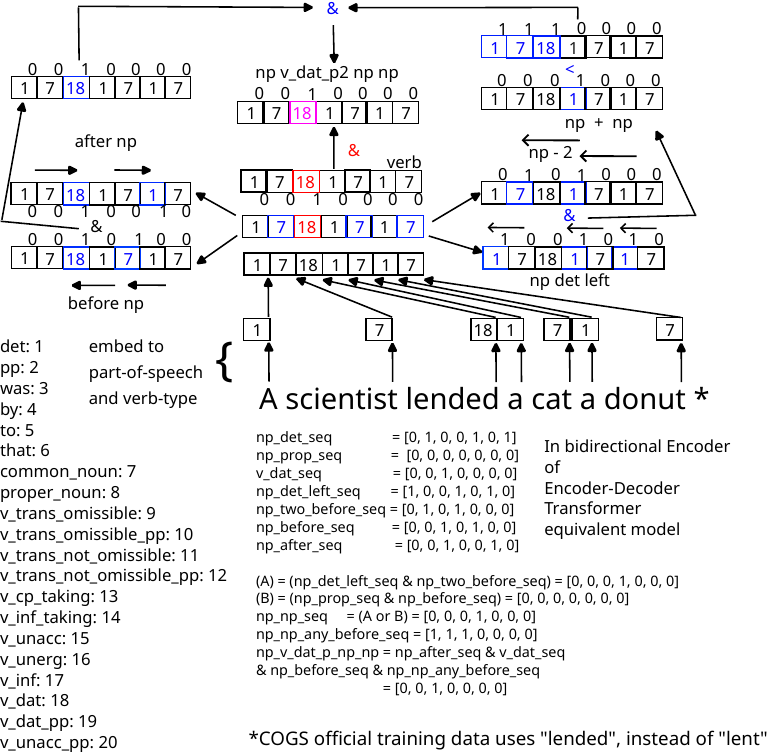}
\caption{\begin{small}Example of the RASP-for-(Re)COGS model flat grammar pattern matching, for np v\_dat\_p2 np np, for a matching sentence.\end{small} \\
}
\label{rasp-model-flat-pattern-match-example-v_dat_p2_match}
\end{figure}

For those used to multidimensional representations, one can think of these as one-hot vectors\footnote{\begin{footnotesize}could also be any 20 approximately orthogonal vectors in a vector space at least 20d (does not need to be one-hot, aligned with basis)\end{footnotesize}} of dimension 20\footnote{\begin{footnotesize}21d in decoder only for v\_normalized\_in\_output\end{footnotesize}} and replace equality checks with vector dot product (and a check for either common\_noun or proper\_noun would be a dot product with the sum of one-hot 20 dimensional vectors given by (common\_noun + proper\_noun)); it is more readable to use scalars in 1d which is also the default in RASP.

Each of the five sequences comes from a separate map, since in RASP a map could only have a single value per key, and since individual words had up to four different verb roles (as in "liked" which had 4).

Upon these five parallel, aligned, sequences we used a series of attention head compatible operations to recognize multi-token patterns (see below) corresponding to grammatical forms (listed below). 

\begin{tiny}
\begin{verbatim}
np_det_mask = select(7, pos_tokens, ==) \
and select(pos_tokens, 1, ==) \
and select(indices+1, indices, ==);
np_prop_mask = select(8, pos_tokens, ==) and \
select(indices, indices, ==);
np_det_sequence = aggregate(np_det_mask, 1);
np_prop_sequence = aggregate(np_prop_mask, 1);
np_det_after = select(np_det_sequence, 1, ==) and \
select(indices+1, indices, ==);
np_prop_after = select(np_prop_sequence, 1, ==) and \
select(indices+1, indices, ==);
np_after_mask = np_det_after or np_prop_after;
np_after_sequence = aggregate(np_after_mask, 1);
np_after_mask = select(np_after_sequence, 1, ==) and \
select(indices,indices, ==);
# ...

# np v_unerg
# e.g. [1,7,16]
set example ["the", "guest", "smiled"]
v_unerg_mask = select(16, pos_tokens_vmap1, ==) and \
select(indices, indices, ==);
np_v_unerg = aggregate(np_after_mask and v_unerg_mask, 1);
\end{verbatim}
\end{tiny}

\begin{figure}
\includegraphics[scale=0.75]{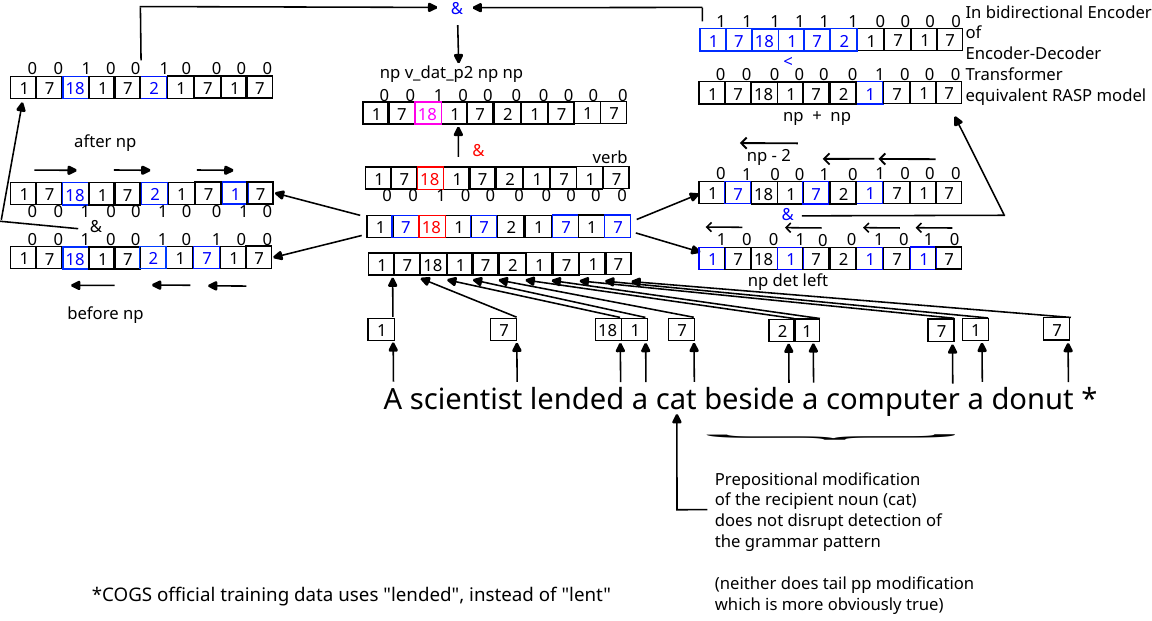}
\onecolumn
\caption{\begin{small}Example of the RASP-for-(Re)COGS model flat grammar pattern matching, for np v\_dat\_p2 np np, for a matching sentence, despite pp modification of middle recipient noun. This is in the encoder (shared for COGS and ReCOGS). See also Figures \ref{attraction_errors_figure}, \ref{rasp-for-recogs-decoder-loop-supplementary-figure_pp_depth_2}, \ref{rasp-for-cogs-decoder-loop-figure_pp_depth_2} for how the RASP model avoids attraction errors in assigning agents, recipients, themes due to prepositional phrase modification in the decoder.\end{small} \\
}
\twocolumn
\label{rasp-model-flat-pattern-match-example-v_dat_p2_match_with_middle_pp}
\end{figure}
\clearpage

\begin{figure}
\includegraphics[scale=0.75]{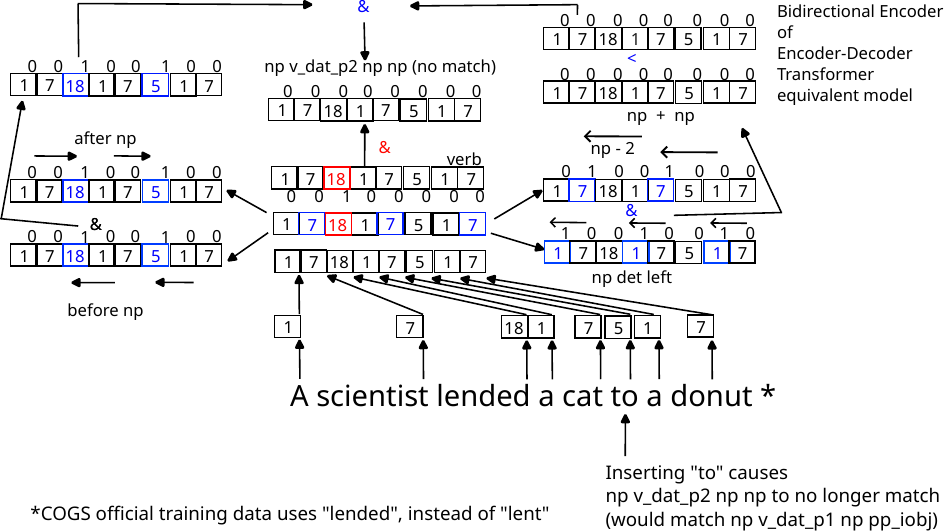}
\caption{\begin{small}Example of the RASP-for-(Re)COGS model performing flat grammar pattern matching, for the pattern np v\_dat\_p2 np np, for a non-matching sentence. This is in the encoder (shared for COGS and ReCOGS).\end{small} \\
}
\label{rasp-model-flat-pattern-match-example-v_dat_p2_non_matching}
\end{figure}
\clearpage

\onecolumn
\begin{figure}
\includegraphics[scale=0.425]{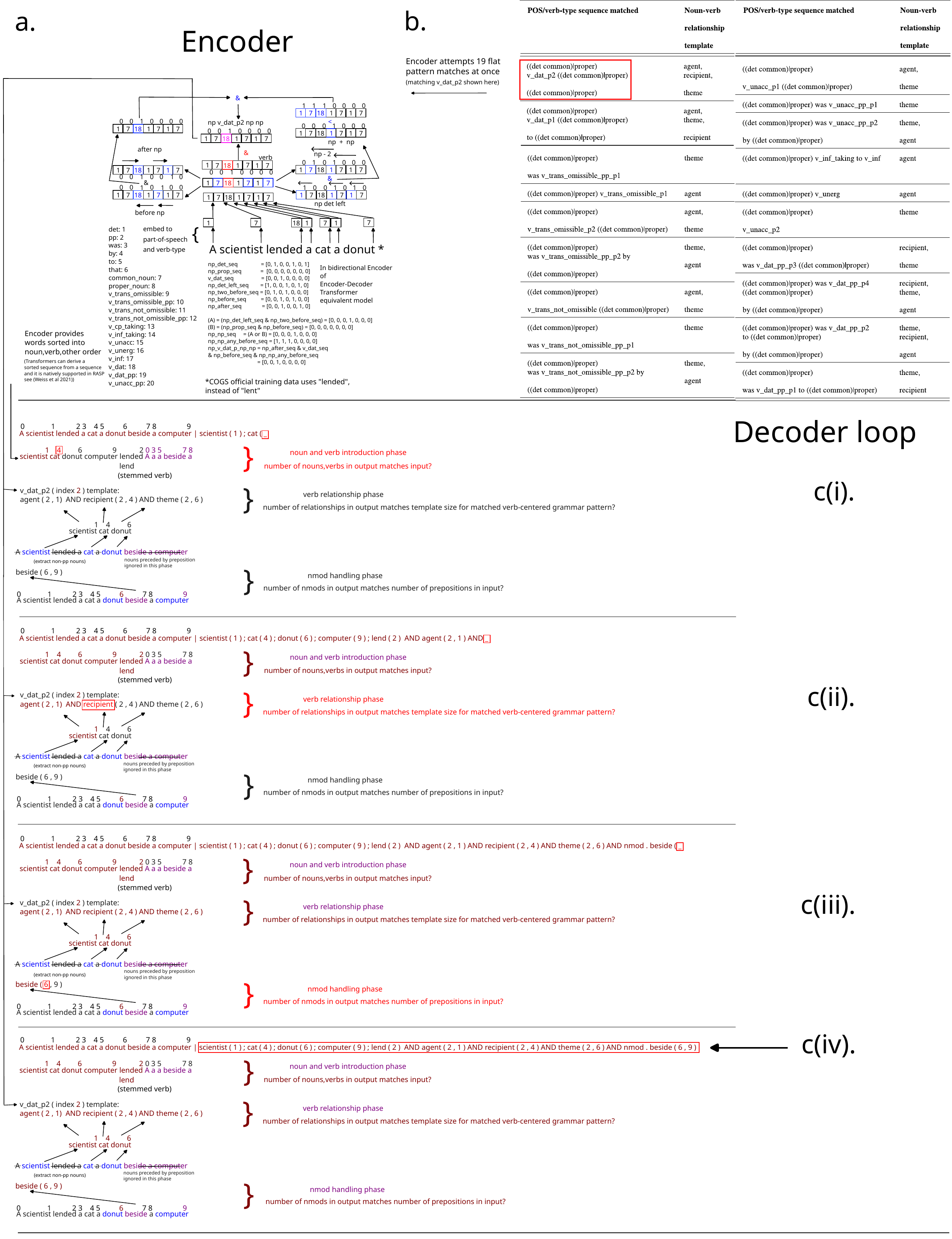}
\caption{\begin{small}RASP-for-ReCOGS model decoding a ReCOGS training example to logical form, for the pattern np v\_dat\_p2 np np. 
(a) The acausal Transformer-Encoder flat matches (fixed Transformer depth indep. of input length) determining the noun-verb relationship template (b) (also Table \ref{RASP-model-flat-patterns-after-masking-to-nv-relationships-table}) for the main verb. The Decoder loop (c) autoregressively generates the logical form c(iv) in phases: c(i) noun-verb introductions, c(ii) verb relationships, and c(iii) pp nmods are shown. See also Figures \ref{rasp-for-recogs-decoder-loop-supplementary-figure_pp_depth_2}, \ref{rasp-for-recogs-decoder-loop-supplementary-figure_cp_depth_2}.
\end{small} \\
}
\label{rasp-for-recogs-decoder-loop-supplementary-figure-incl-encoder-and-decoder-and-grammar-vertical}
\end{figure}
\twocolumn
\clearpage

\onecolumn
\begin{figure}
\includegraphics[scale=0.50]{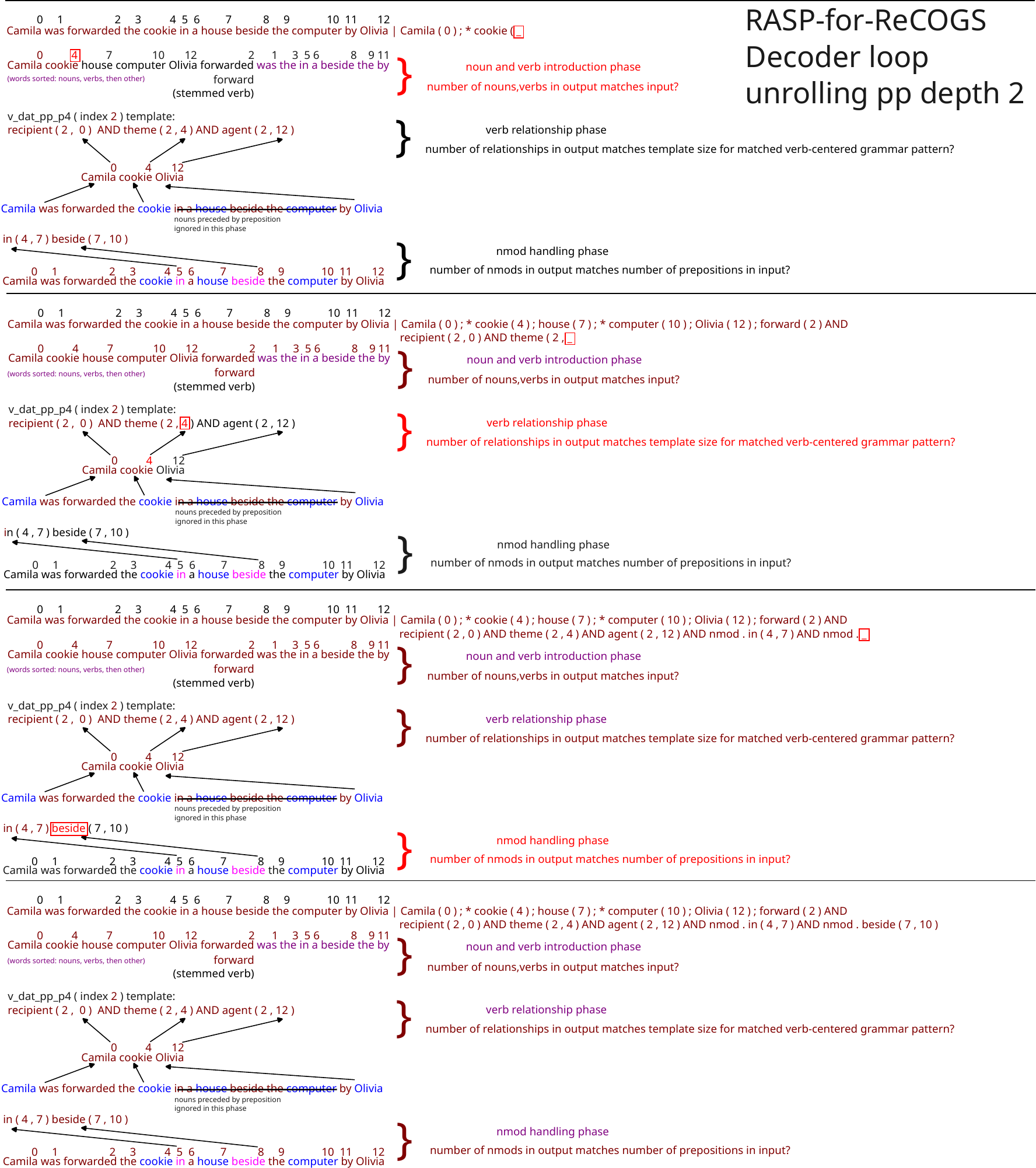}
\caption{\begin{small}Example of the RASP-for-ReCOGS model decoding a ReCOGS training example with center-embedded prepositional phrase recursion depth 2 to logical form. See Figure \ref{rasp-for-recogs-decoder-loop-supplementary-figure-incl-encoder-and-decoder-and-grammar-vertical} for an overview of the model and to review how the Decoder loop is illustrated in these figures.\end{small} \\
}
\label{rasp-for-recogs-decoder-loop-supplementary-figure_pp_depth_2}
\end{figure}
\twocolumn
\clearpage

\onecolumn
\begin{figure}
\includegraphics[scale=0.425]{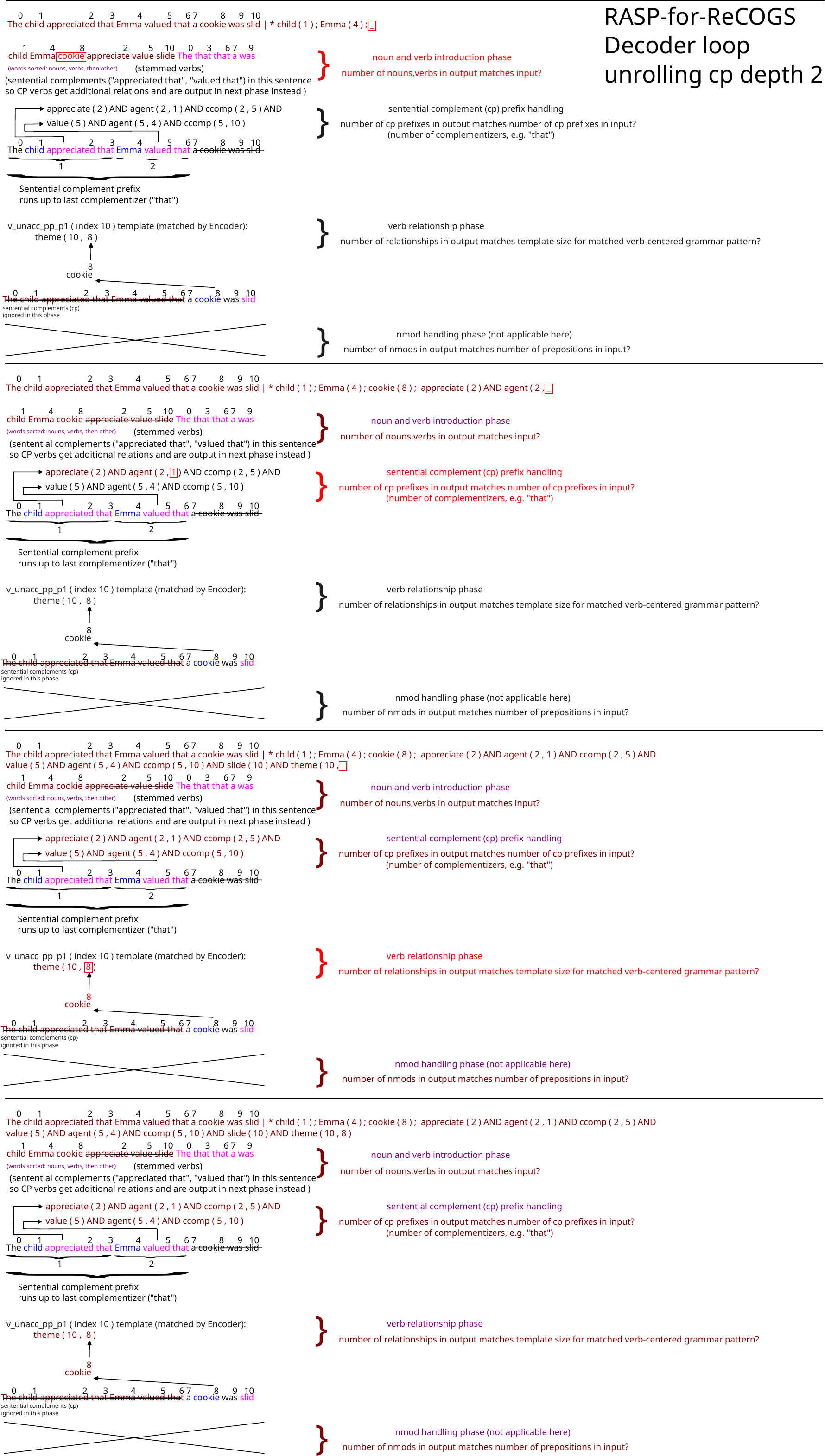}
\caption{\begin{small}Example of the RASP-for-ReCOGS model decoding a ReCOGS training example with sentential complement (cp) recursion depth 2 to ReCOGS logical form. See Figure \ref{rasp-for-recogs-decoder-loop-supplementary-figure-incl-encoder-and-decoder-and-grammar-vertical} for an overview of the model and to review how the Decoder loop is illustrated in these figures.\end{small} \\
}
\label{rasp-for-recogs-decoder-loop-supplementary-figure_cp_depth_2}
\end{figure}
\twocolumn
\clearpage

\begin{figure}
\includegraphics[scale=0.40]{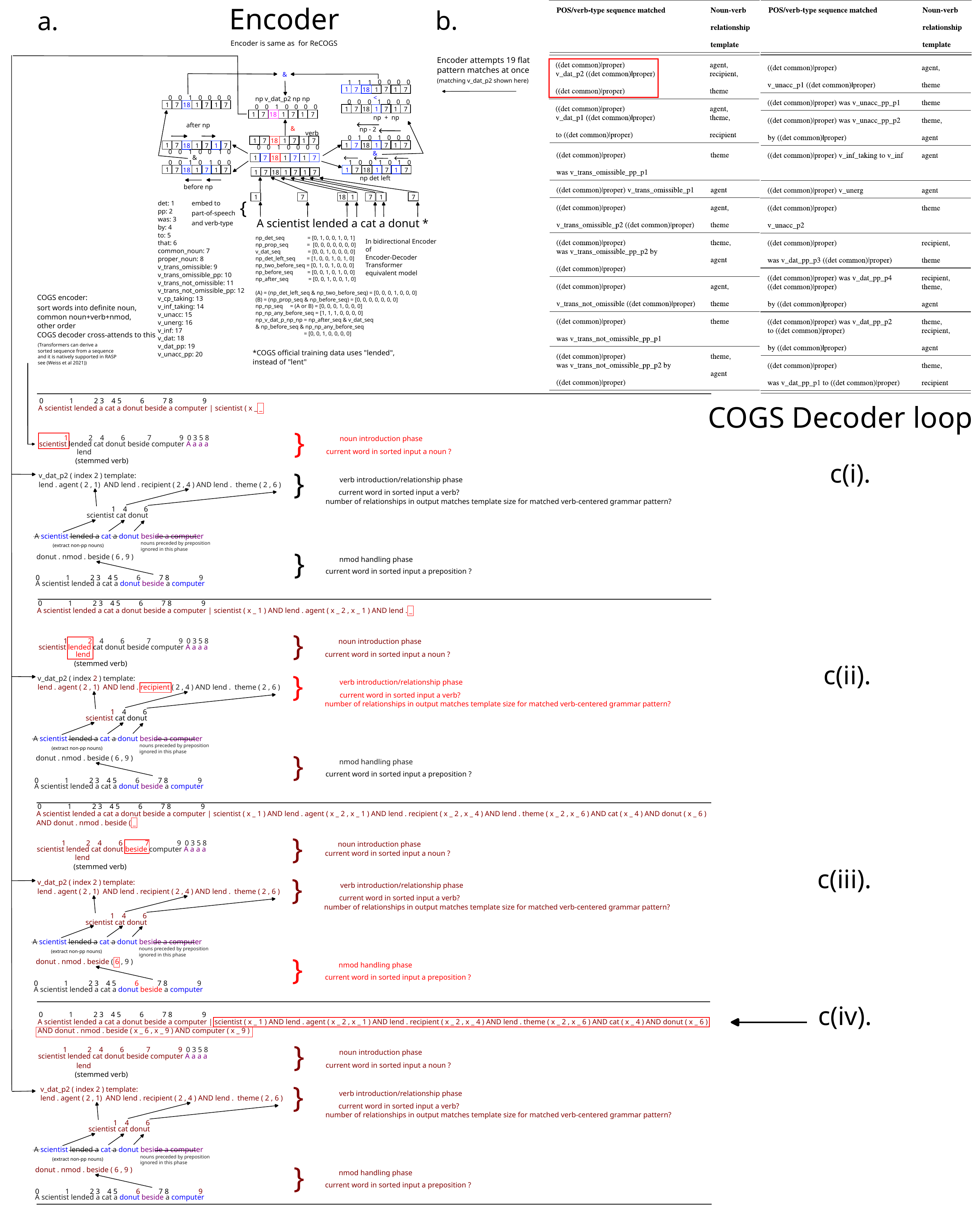}
\onecolumn
\caption{\begin{small}RASP-for-COGS model decoding a COGS training example to logical form, for the pattern np v\_dat\_p2 np np. (a) The acausal Transformer-Encoder flat matches (fixed Transformer depth indep. of input length) determining the noun-verb relationship template (b) (also Table \ref{RASP-model-flat-patterns-after-masking-to-nv-relationships-table}) for the main verb. The Decoder loop (c) autoregressively generates the logical form c(iv) in phases: c(i) noun introductions, c(ii) verb relationships, and c(iii) pp nmods are shown. See also Figures \ref{rasp-for-cogs-decoder-loop-figure_pp_depth_2}, \ref{rasp-for-cogs-decoder-loop_cp_depth_2}.
\end{small}}
\label{rasp-for-cogs-encoder-decoder-with-grammar-patterns}
\twocolumn
\end{figure}
\clearpage

\begin{figure}
\includegraphics[scale=0.40]{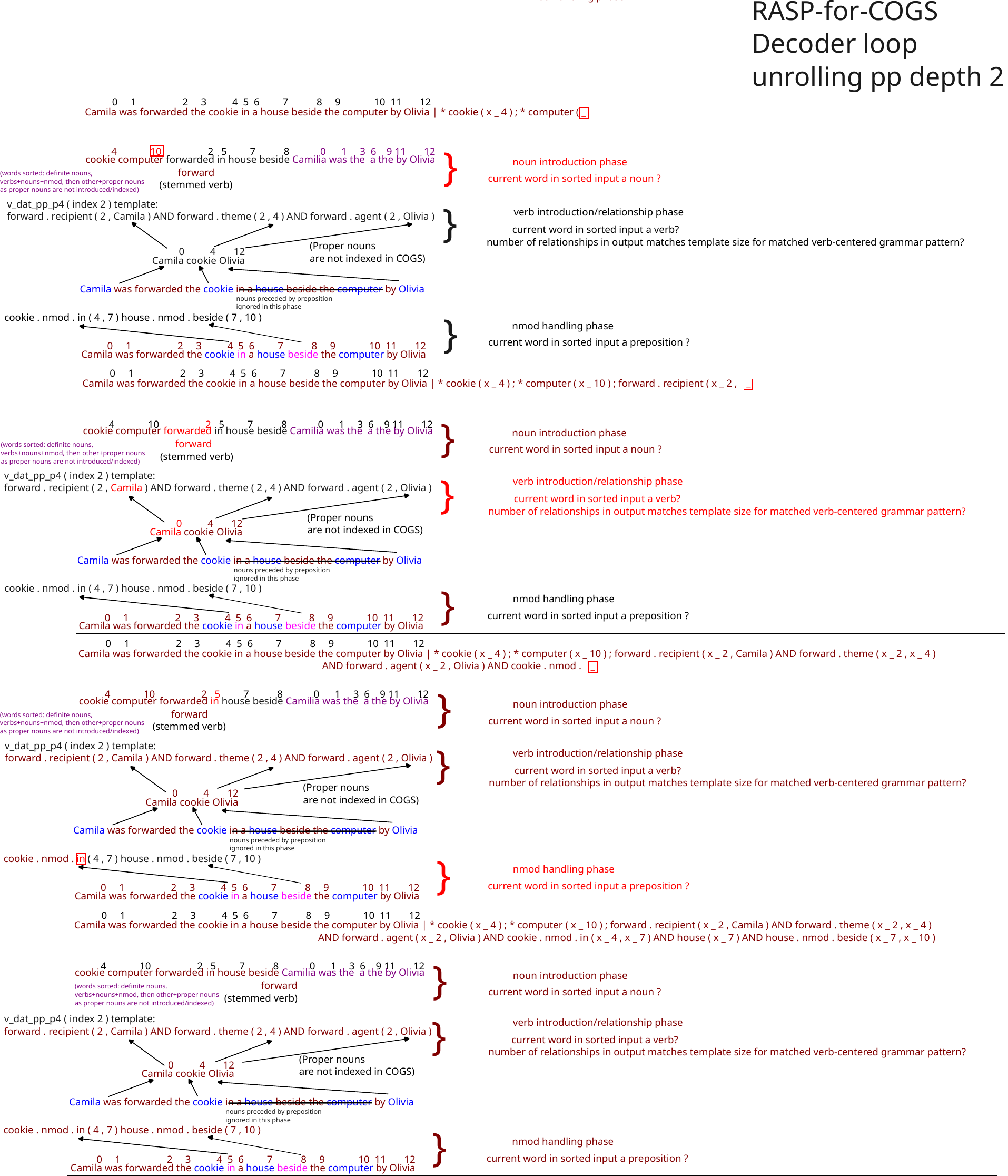}
\onecolumn
\caption{\begin{small}Decoder of RASP model for COGS translating center-embedded prepositional phrase recursion depth 2 to logical form without a recursive or tree-structured representation by unrolling it in the decoder loop. See Figure \ref{rasp-for-cogs-encoder-decoder-with-grammar-patterns} for an overview of the model and to review how the Decoder loop is illustrated in these figures.\end{small}}
\twocolumn
\label{rasp-for-cogs-decoder-loop-figure_pp_depth_2}
\end{figure}
\clearpage
\begin{figure}
\includegraphics[scale=0.40]{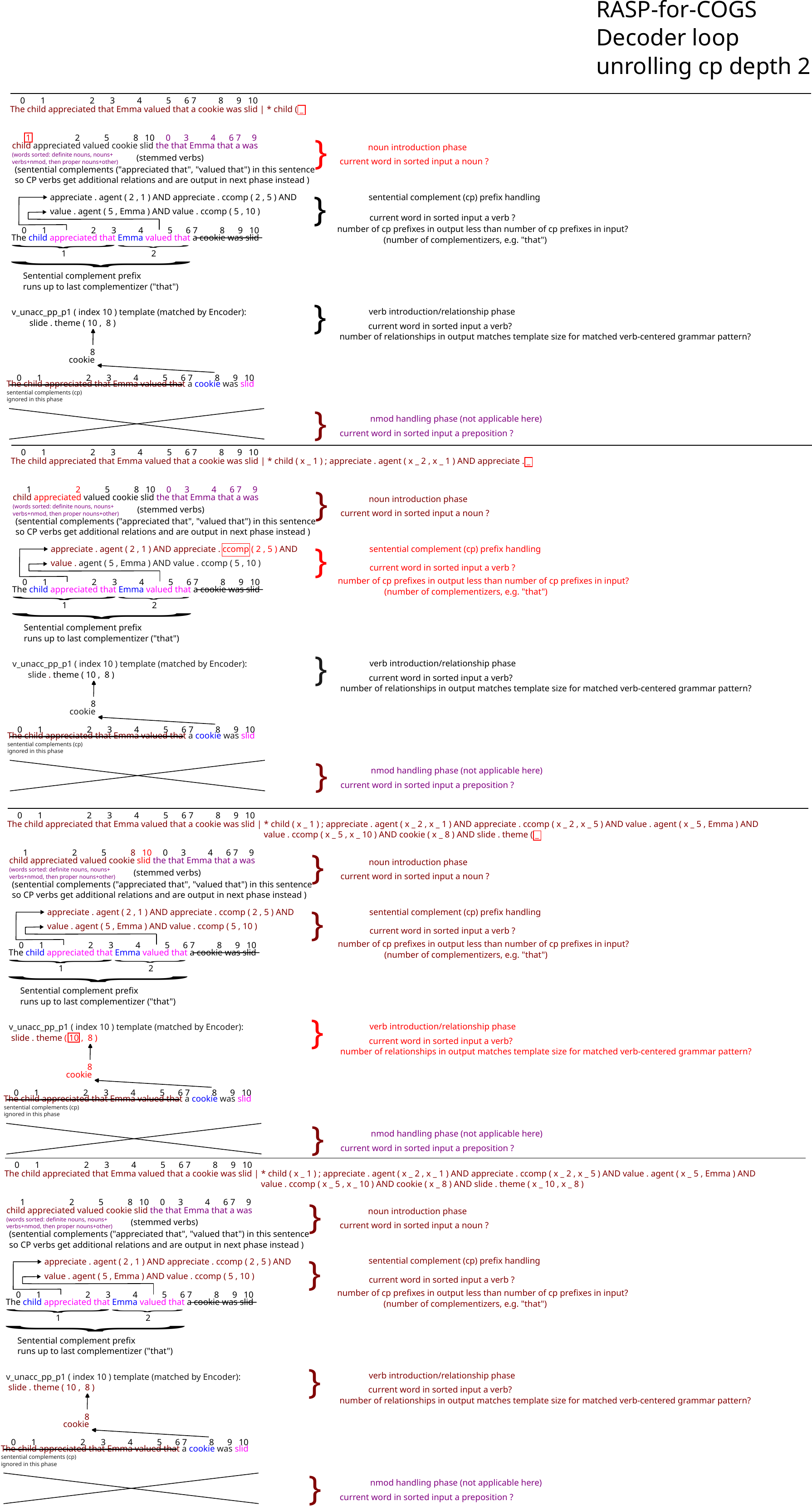}
\onecolumn
\caption{\begin{small}Decoder of RASP model for COGS translating sentential complement (cp) recursion depth 2 to logical form without a recursive or tree-structured representation by unrolling it in the decoder loop. See Figure \ref{rasp-for-cogs-encoder-decoder-with-grammar-patterns} for an overview of the model and to review how the Decoder loop is illustrated in these figures.\end{small}}
\twocolumn
\label{rasp-for-cogs-decoder-loop_cp_depth_2}
\end{figure}
\clearpage

\textbf{These patterns are not causal because their use/input/output is masked to the input section of the sequence, so would take part in the Encoder of the Encoder-Decoder only} (\textbf{all operations outside the input mask in the word-level token RASP solution used in this paper are directly or indirectly causally masked} and we built symbol by symbol in a causal autoregressive way). \textbf{We could have added an explicit causal mask to each operation but for efficiency and simplicity of the code omitted it when we are doing it implicitly by taking only the last sequence position (we also acausally aggregate so that all sequence positions have the same value as the last sequence position to make it easier to read the output -- RASP interpreter will just print it as one position if they are all equal and we only take one position).}

Also, the author thinks many of these RASP steps could be consolidated. The goal here was to first prove by construction that a flat RASP program not directly implementing the recursive grammar rules could get approximately 100\% Semantic Exact Match on all the ReCOGS generalization splits (we only missed one split, obj-pp-to-subj-pp by a little (see results), which we believe was due to a mistake made rushing due to two week time constraint for original model implementation, not a fundamental limitation of the RASP approach; and indeed with RASP-for-COGS (not ReCOGS, implemented after) we indeed achieved 100\% on the obj-pp-to-subj-pp generalization the first attempt).

Introduction of variables at the beginning of the ReCOGS logical form (e.g. in the logical form for "a boy painted the girl", we have "boy ( 1 ) ; * girl ( 4 ) ; paint ( 2 ) AND agent ( 2 , 1 ) AND theme ( 2 , 4 )" , the variable introduction is "boy ( 1 ) ; * girl ( 4 ) ; paint ( 2 )" before the "AND"). We took a simple approach and sorted the input sequence with nouns before verbs and determiners, fillers last (with determiners and fillers not having any corresponding entry in the output sequence). We then count nouns and verbs in the input and count nouns and verbs in the output and determine if we have introduced all the nouns and verbs.

See code for full details\footnote{\begin{footnotesize}word-level token Restricted Access Sequence Processing solution: \href{https://github.com/willy-b/RASP-for-COGS}{https://github.com/willy-b/RASP-for-COGS} (COGS) and \href{https://github.com/willy-b/learning-rasp}{https://github.com/willy-b/learning-rasp} (ReCOGS)\end{footnotesize}} (for simplicity this description was written without discussing sentential complement (CP) handling).

Example counting how many nouns and verbs we have output (introduced as variables) so far (to determine what we need to output for next token):
\begin{tiny}
\begin{verbatim}
nv_in_output_sequence = \
OUTPUT_MASK*(indicator(pos_tokens == 7 or pos_tokens == 8) + \
indicator(pos_tokens_vmap1 == 9 or pos_tokens_vmap2 == 10 or \
pos_tokens_vmap1 == 11 or pos_tokens_vmap2 == 12 or \
pos_tokens_vmap3 == 13 or pos_tokens_vmap4 == 14 or \
pos_tokens_vmap1 == 15 or pos_tokens_vmap1 == 16 or \
pos_tokens_vmap1 == 17 or pos_tokens_vmap1 == 18 or \
pos_tokens_vmap2 == 19 or pos_tokens_vmap2 == 20 or \
pos_tokens_vmap1==21));
nv_in_output_count = selector_width(select(nv_in_output_sequence, 1, ==));
# causal operation as we use only last sequence position
\end{verbatim}
\end{tiny}

How variables are introduced with their index (omitted sorting of input and other operations that can be read in the code and are less important; anything acausal is restricted to input sequence section (Encoder)):
(only value at last sequence position is used; causal)
\begin{tiny}
\begin{verbatim}
# introducing variables
output = "";
# definite article word handling
before_target_word_index = \
 aggregate(select(indices, nv_in_output_count, ==), \
 input_indices_sorted)-1;
has_star = \
 aggregate(select(indices, before_target_word_index, ==), \
 tokens) == "the";
last_output_is_star = \
 aggregate(select(indices, length-1, ==), \
 tokens) == "*";

input_nv_sorted_by_type = \
 input_tokens_sorted_by_type * \
 (input_noun_mask_sorted + input_verb_mask_sorted);
target_word_token = \
 aggregate(select(indices, nv_in_output_count, ==), \
 normalize_nv(input_nv_sorted_by_type)) \
if (not has_star or last_output_is_star) else "*";
# subtract 1 when matching 
# for producing the index 
# because we just output the additional word by then
target_word_index = \
 aggregate(select(indices, nv_in_output_count-1, ==), \
 input_indices_sorted);

output = \
 target_word_token \
 if ((num_tokens_in_output_excluding_asterisks % 5) == 0) \
 else \
 output;
output = \
 "(" \
 if ((num_tokens_in_output_excluding_asterisks % 5) == 1) \
 else output;
output = \
 target_word_index \
 if ((num_tokens_in_output_excluding_asterisks % 5) == 2) \
 else output;
output = \
 ")" \
 if ((num_tokens_in_output_excluding_asterisks % 5) == 3) \
 else output;
# note that 
# when nv_in_output_count == nv_in_input_count, 
# we will add AND instead of ";"
output = \
 ( \
  ";" \
  if \ 
  ( \
   5 * nv_in_input_count - 1 > \
   num_tokens_in_output_excluding_asterisks \
  ) \
  else "AND" \
 ) \
if (num_tokens_in_output_excluding_asterisks % 5 == 4) \
else output;

# if we didn't have an input/output separator 
# that needs to be output
output = \
 "|" if num_pipes_in_output == 0 else output;

# note that the output/next token prediction above will be overridden 
# with later decoder variables 
# (e.g. verb relationship or noun modifier logical form tokens, see below)
# if noun/verb introduction is complete, that is if the decoder detects
# that all nouns/verbs in input have been output in the logical form.
\end{verbatim}
\end{tiny}
Note that "normalize\_nv" is a lookup into a map that has no effect unless the word is a verb in which case it normalizes it to a standard suffix ("ate" to "eat" , "painted" to "paint", etc).
\clearpage
As you can see above, if we have not introduced all the variables, we determine our index into the sorted list of nouns and verbs (nouns before verbs), and using a MLP modeling modulus, compute index mod 5 (alternatively, number of tokens since last separator)  and condition on that to output that noun/verb or parentheses or index as prediction for next token at last sequence position (all other sequence positions are ignored). Since we do ReCOGS\_pos (semantically identical to random indices but avoid requiring random numbers) the index we use is the index of the original noun or verb in the original sequence. If we are still introducing variables, that is the end and we have our prediction for the next token.

If we are done introducing variables at that point in the Decoder loop, we move on,
and use templates that the attention head compatible operations in the Encoder recognized for us in the five parallel part-of-speech / verb-type per location sequences for "v\_trans\_omissible\_p1", "v\_trans\_omissible\_p2", "v\_trans\_omissible\_pp\_p1", "v\_trans\_omissible\_pp\_p2", "v\_trans\_not\_omissible", "v\_trans\_not\_omissible\_pp\_p1", "v\_trans\_not\_omissible\_pp\_p2", "v\_cp\_taking", "v\_inf\_taking", "v\_unacc\_p1", "v\_unacc\_p2", "v\_unacc\_pp\_p1", "v\_unacc\_pp\_p2", "v\_unerg", "v\_dat\_p2", "v\_dat\_pp\_p1", "v\_dat\_pp\_p2",  "v\_dat\_pp\_p3",  "v\_dat\_pp\_p4".

To be clear, we always compute all variables (noun and verb introduction, verb relationships, nmods) but depending on the number of nouns/verbs, verb relationships, nmods detected in the output so far, variables from "completed" or "premature" phases are discarded and the next predicted token is given by variables associated with the appropriate phase, here the verb relationship phase.

Here are a couple of examples of patterns the Encoder recognizes, to see how it looks if we support 1 verb pattern per input (no sentential complement recursion; which can be easily handled how we handle other things we loop over, looping over current phrase and masking and processing), which is sufficient to get approximately 100\% on all entries that do not use sentential complements (e.g. "the girl noticed that a boy painted" is not supported in this example but "a boy painted" is):
\begin{tiny}
\begin{verbatim}
# define the pattern
# ... \
# (just showing one example, 
#  np_prop_mask and 
#  np_before_mask 
#  are attention masks defined earlier)
# np v_dat_p2 np np
# e.g. [8,18,1,7,1,7]
set example ["ella","sold","a","customer","a","car"]
np_np_sequence = \
  aggregate((np_prop_mask and np_before_mask) or \
  (np_det_left_mask and np_two_before_mask), 1);
# would not support prepositional phrase modification on middle NP
#np_np_before_mask = \
#  select(np_np_sequence, 1, ==) and select(indices-1, indices, ==);
np_np_any_before_mask = \
  select(np_np_sequence, 1, ==) and select(indices, indices, >);
# acausal is ok 
# in INPUT sequence (encoder part, not decoder), \
# would mask further if we wanted to do multiple templates per input or
# something outside the supported grammar (COGS without sentential complement
# recursion is supported here)
np_np_any_before_sequence = \
  aggregate(np_np_any_before_mask, 1);
np_np_any_before_mask = \
  select(np_np_any_before_sequence, 1, ==) and \
  select(indices, indices, ==);
np_v_dat_p_np_np = \
  aggregate(np_after_mask and v_dat_mask and \
    np_before_mask \
    and np_np_any_before_mask, 1);
# Example: np_v_dat_p_np_np(\
#  ['ella', 'sold', 'a', 'customer', 'a', 'car']) \
#  = [0, 1, 0, 0, 0, 0] (ints)
# Example: np_v_dat_p_np_np(\
#  [8, 18, 1, 7, 1, 7]) \
#  = [0, 1, 0, 0, 0, 0] (ints)

# ...

# check the pattern and set the template name
any_np_v_trans_omissible = \
  aggregate(select(np_v_trans_omissible, 1, ==), 1);
template_name = "v_trans_omissible_p1" \
if (any_np_v_trans_omissible == 1) else template_name;

# ...

any_v_dat_p2 = aggregate(select(np_v_dat_p_np_np, 1, ==), 1);
template_name = \
  "v_dat_p2" if (any_v_dat_p2 == 1) else template_name;

# ...

any_v_dat_pp_p4 = \
  aggregate(select(np_was_v_dat_pp_np_by_np, 1, ==), 1);
template_name = \
  "v_dat_pp_p4" if (any_v_dat_pp_p4 == 1) else template_name;

# must be checked after P4
any_v_dat_pp_p2 = \
  aggregate(select(np_was_v_dat_pp_to_np_by_np, 1, ==), 1);
template_name = \
  "v_dat_pp_p2" if (any_v_dat_pp_p2 == 1) else template_name;

# template name is used to lookup 
# the number of verb relationships to output
# and what they are for each index
# e.g. ["theme", "agent"] 
# vs. ["agent", "recipient", "theme"] etc
\end{verbatim}
\end{tiny}
\clearpage

The rest of this applies to just values used from the last sequence location (output is prediction for next symbol).

Based on the template recognized, we lookup the template size for number of relationships (theme, recipient, agent) for that verb type:
\begin{tiny}
\begin{verbatim}
def template_size(template_name) {
  # number of items to output in verb relationship phase
  # after noun and verb introduction phase
  # (special exception is 2-verb v_inf)
  template_sizes = {
 "": 0,
 "v_trans_omissible_p1": 1,
 "v_trans_omissible_p2": 2,
 "v_trans_omissible_pp_p1": 1,
 "v_trans_omissible_pp_p2": 2,
 "v_trans_not_omissible": 2,
 "v_trans_not_omissible_pp_p1": 1,
 "v_trans_not_omissible_pp_p2": 2,
 "v_cp_taking": 2,
 # (NOTE: comments within the map should be removed)
 # v_inf_taking is a special 2-verb case, 5 items
 # after noun introduction
 # (verb 1, agent 1, xcomp verb 1 to verb 2, verb 2, agent 2)
 # if first verb were included 
 # in introduction phase for v_inf,
 # then it would be 4
 # indeed the last map used is template_mapping4
 "v_inf_taking": 5,
 # n.b. if we output out of order
 # (as allowed by Semantic Exact Match)
 # and put both verbs in beginning
 # (verb 1, verb 2, agent 1, xcomp verb 1 to verb 2, agent 2)
 # then the count would be 3
 # as we could do normal combined noun and verb intro
 # but if doing String Exact Match, ReCOGS LF has
 # verb 2 after the xcomp for v_inf_taking
 "v_unacc_p1": 2,
 "v_unacc_p2": 1,
 "v_unacc_pp_p1": 1,
 "v_unacc_pp_p2": 2,
 "v_unerg": 1,
# "v_inf": 1,
 "v_dat_p1": 3,
 "v_dat_p2": 3,
 "v_dat_pp_p1": 2,
 "v_dat_pp_p2": 3,
 "v_dat_pp_p3": 2,
 "v_dat_pp_p4": 3
  };
  return template_sizes[template_name];
}
\end{verbatim}
\end{tiny}

Details are in the RASP-for-COGS and learning-rasp GitHub RASP code\footnote{\begin{footnotesize}word-level token Restricted Access Sequence Processing solution: \href{https://github.com/willy-b/RASP-for-COGS}{https://github.com/willy-b/RASP-for-COGS} (COGS) and \href{https://github.com/willy-b/learning-rasp}{https://github.com/willy-b/learning-rasp} (ReCOGS) \end{footnotesize}}, but we compute at the last sequence position (in parallel) the number of relationships output for the verb so far, and for the current relationship which token within that multi-token process (e.g. the word "agent" or the open parenthesis "(" or the left index, or the comma, or right index, close parenthesis ")", "AND", etc) we are on.

Like we computed at the last sequence position the number of nouns and verbs in the output once we are finished introducing nouns and verbs (this would be a little different with sentential complements (see actual code for CP support)), we compute the number of agent,theme,recipient,xcomp entries in the output:
\begin{tiny}
\begin{verbatim}
atrx_in_output_sequence = OUTPUT_MASK*(indicator(tokens == "agent" \
or tokens == "theme" \
or tokens=="recipient" \
or tokens=="xcomp"));
# agent_theme_recipient_xcomp_output_count is the number of relationships we have output
agent_theme_recipient_xcomp_output_count = \
selector_width(select(atrx_in_output_sequence, 1, ==));
after_intro_idx = \
  (nv_in_output_count - nv_in_input_count + \
   (1 if any_v_inf_taking == 1 else 0) \
   + agent_theme_recipient_xcomp_output_count) \
   if nv_in_output_count + (1 if any_v_inf_taking == 1 else 0) \
   >= nv_in_input_count else 0;
\end{verbatim}
\end{tiny}
\clearpage
We use all those different values which are computed independently from the input sequence and so do not add depth/layer requirements as many of the ones which involve accessing the sequence can be done in parallel. 
We then use the verb-type and relationship index to the relationship into a map to get the current relationship to output (as some verb types output agent first, some output theme, etc):
\begin{tiny}
\begin{verbatim}
template_mapping1 = {
 "": "",
 "v_trans_omissible_p1": "agent",
 "v_trans_omissible_p2": "agent",
 "v_trans_omissible_pp_p1": "theme",
 "v_trans_omissible_pp_p2": "theme",
 "v_trans_not_omissible": "agent",
 "v_trans_not_omissible_pp_p1": "theme",
 "v_trans_not_omissible_pp_p2": "theme",
 "v_cp_taking": "agent",
 "v_inf_taking": "agent",
 "v_unacc_p1": "agent",
 "v_unacc_p2": "theme",
 "v_unacc_pp_p1": "theme",
 "v_unacc_pp_p2": "theme",
 "v_unerg": "agent",
 "v_inf": "agent",
 "v_dat_p1": "agent",
 "v_dat_p2": "agent",
 "v_dat_pp_p1": "theme",
 "v_dat_pp_p2": "theme",
 "v_dat_pp_p3": "recipient",
 "v_dat_pp_p4": "recipient"
};
\end{verbatim}
\end{tiny}

Outputting the verb relationships we must skip over any "pp np" as possible agents, themes, or recipients to avoid getting confused by noun phrases added by prepositional modification (believed by the author to be a cause of the issue with obj pp to subj pp generalization by \citep{Wu2023}'s Transformer).

\begin{tiny}
\begin{verbatim}
pp_sequence = indicator(pos_tokens == 2);
pp_one_after_mask = select(pp_sequence, 1, ==) and \
  select(indices+1, indices, ==);
pp_one_after_sequence = aggregate(pp_one_after_mask, 1);
pp_one_after_mask = select(pp_one_after_sequence, 1, ==) and \
  select(indices, indices, ==);

pp_two_after_mask = select(pp_sequence, 1, ==) and \
  select(indices+2, indices, ==);
pp_two_after_sequence = aggregate(pp_two_after_mask, 1);
pp_two_after_mask = select(pp_two_after_sequence, 1, ==) and \
  select(indices, indices, ==);

np_det_diag_mask = select(aggregate(np_det_mask, 1), 1, ==) and \
  select(indices, indices, ==);
np_prop_diag_mask = select(aggregate(np_prop_mask, 1), 1, ==) and \
  select(indices, indices, ==);

no_pp_np_mask = \
1 - aggregate((pp_one_after_mask and np_prop_diag_mask) or \
(pp_two_after_mask and np_det_diag_mask), 1);

# here we compute left_idx and right_idx
# for verb relationships, like "agent ( [left_idx] , [right_idx] )"

# one-based index
nps_without_pp_prefix_indices = \
selector_width(select(NOUN_MASK*no_pp_np_mask, 1, ==) and \
select(indices, indices, <=))*NOUN_MASK*no_pp_np_mask;

# the one verb (except v_inf_taking cases)
left_idx_in_nvs_zero_based = nv_in_input_count-1;
# (after sentential complements, not covered in this example, see actual code via link above)
# need to also subtract the index in ReCOGS for the 2nd verb if it is v_inf_taking
left_idx_in_nvs_zero_based = (left_idx_in_nvs_zero_based-1) \
  if (template_name == "v_inf_taking" and after_intro_idx <= 2) else left_idx_in_nvs_zero_based;
left_idx = aggregate(select(indices, left_idx_in_nvs_zero_based, ==), input_indices_sorted);

# avoids attractor nouns
right_idx = aggregate(select(nps_without_pp_prefix_indices, after_intro_idx, ==), indices);

# points to 2nd verb for xcomp for v_inf_taking_v_inf
# note, this simplified example ignores sentential complement (CP) handling
# (ideally this would be verb specific,
#  we simplify here to reuse variables available in this example)
right_idx = aggregate(select(indices, nv_in_output_count, ==), input_indices_sorted) \
   if (template_name == "v_inf_taking" and after_intro_idx == 2) else right_idx;

# points to 1st noun for 2nd v_inf agent in v_inf_taking_v_inf
right_idx = \
  aggregate(select(nps_without_pp_prefix_indices, 1, ==), indices) \
  if (template_name == "v_inf_taking" and after_intro_idx == 4) else right_idx;

# we have computed left_idx and right_idx
# for verb relationships, like "agent ( [left_idx] , [right_idx] )"

# ...

# note, the offset since the last separator in the output,
# instead of a modulus, could have been used here
# see actual RASP file in GitHub for computation

# relationship ( idx , idx ) AND
#      0       1  2  3  4  5  6
after_intro_target_token = "";

# "agent", "theme", "recipient", etc 
# depending on relationship index and flat-matched template (in Encoder)
template_mapping_output = \
  get_template_mapping(template_name, after_intro_idx);

# see code in Github for definition of 
# after_intro_num_tokens_in_output_excluding_asterisks
# and use of an offset that depends on v_inf or not
# out of scope for this simplified example
after_intro_target_token = template_mapping_output \
  if ((after_intro_num_tokens_in_output_excluding_asterisks) % 7 == 0) \
  else after_intro_target_token;

after_intro_target_token = "(" \
  if ((after_intro_num_tokens_in_output_excluding_asterisks) % 7 == 1) \
  else after_intro_target_token;

after_intro_target_token = left_idx \
if (after_intro_num_tokens_in_output_excluding_asterisks % 7 == 2) \
else after_intro_target_token;

after_intro_target_token = "," \
if (after_intro_num_tokens_in_output_excluding_asterisks % 7 == 3) \
else after_intro_target_token;

after_intro_target_token = right_idx \
if (after_intro_num_tokens_in_output_excluding_asterisks % 7 == 4) \
else after_intro_target_token;

after_intro_target_token = ")" \
if (after_intro_num_tokens_in_output_excluding_asterisks % 7 == 5) \
else after_intro_target_token;

after_intro_target_token = "AND" \
if \
(after_intro_num_tokens_in_output_excluding_asterisks % 7 == 6 \
and \
not (template_mapping_output == "")) \
else after_intro_target_token;

# ...

# the next token predicted ("output") is
# overridden with after_intro_target_token
# ONLY if the decoder detects that verb relationships 
# are the appropriate output phase
# by counting how many nouns/verbs/relationships are already in the output
\end{verbatim}
\end{tiny}
\clearpage
After outputting all verb relationships, we consider whether we have prepositional phrase noun modifiers to record in the logical form.
That is to say, if the current output sequence already includes the verb relationships expected for the input (the count matches), 
then those Decoder variables discussed earlier for verb relationships are still computed, but discarded,
and the next token predicted will be overridden with a prepositional phrase noun modifier related output given by a RASP path similar to that below.

For outputting prepositional relationships ("nmod . in", "nmod . on", "nmod . beside"), we do a similar approach, counting prepositional phrases in the input, counting how many nmods we have output, figuring out which one is currently being output:

\begin{tiny}
\begin{verbatim}
pps_in_input_sequence = INPUT_MASK*(indicator(pos_tokens == 2));
pps_in_input_count = selector_width(select(pps_in_input_sequence, 1, ==));
pps_index = \
  pps_in_input_sequence*selector_width(select(pps_in_input_sequence, 1, ==) \
and select(indices,indices, <=));

nmods_and_pps_in_output_sequence = \
  OUTPUT_MASK*(indicator(tokens == "nmod . in" or \
  tokens == "nmod . beside" or tokens=="nmod . on"));
nmods_and_pps_in_output_count = \
  selector_width(select(nmods_and_pps_in_output_sequence, 1, ==));
\end{verbatim}
\end{tiny}

\begin{tiny}
\begin{verbatim}
current_pp = \
  aggregate(select(pps_index, nmods_and_pps_in_output_count+1, ==), tokens) \
  if pps_in_input_count > 0 else "";
current_pp = "" if current_pp == 0 else current_pp;
current_nmod_token = \
("nmod . " + current_pp) if (pps_in_input_count > 0 and not (current_pp == 0) \
and after_intro_num_tokens_in_output_excluding_asterisks % 7 == 0) else "";
current_nmod_token = \
  "(" if after_intro_num_tokens_in_output_excluding_asterisks % 7 == 1 else current_nmod_token;
current_nmod_token = \
(aggregate(select(pps_index, nmods_and_pps_in_output_count, ==), indices)-1) if pps_in_input_count > 0 \
and after_intro_num_tokens_in_output_excluding_asterisks % 7 == 2 else current_nmod_token;
current_nmod_token = "," \
if after_intro_num_tokens_in_output_excluding_asterisks % 7 == 3 else current_nmod_token;
after_nmod_idx = \
aggregate(select(pps_index, nmods_and_pps_in_output_count, ==), indices)+1;
token_at_after_nmod_idx = \
aggregate(select(indices, after_nmod_idx, ==), tokens);
after_nmod_idx = \
  (after_nmod_idx + 1) \
  if (token_at_after_nmod_idx == "the" or token_at_after_nmod_idx == "a") else after_nmod_idx;
current_nmod_token = (after_nmod_idx) \
if pps_in_input_count > 0 \
and after_intro_num_tokens_in_output_excluding_asterisks % 7 == 4 else current_nmod_token;
current_nmod_token = ")" \
if after_intro_num_tokens_in_output_excluding_asterisks % 7 == 5 \
else current_nmod_token;
current_nmod_token = \
("AND" if nmods_and_pps_in_output_count < pps_in_input_count else "") \
if after_intro_num_tokens_in_output_excluding_asterisks % 7 == 6 \
else current_nmod_token;
after_intro_and_relationships_nmod_token = \
current_nmod_token if nmods_and_pps_in_output_count <= pps_in_input_count else "";
num_tokens_in_nmod_section = \
after_intro_num_tokens_in_output_excluding_asterisks - template_size(template_name)*7 + 1;

# the decision of whether the nmod output dominates the current decoder next predicted token
# is computed similarly to the following (see GitHub for actual)
# we have not included the computation of "offset" and "after_intro_num_tokens_in_output_excluding_asterisks"
output = after_intro_and_relationships_nmod_token \
  if (template_mapping_output == "" and \
    after_intro_num_tokens_in_output_excluding_asterisks >= \
    template_size(template_name)*7+offset - 1 and \
    num_tokens_in_nmod_section < 7*pps_in_input_count and pps_in_input_count > 0 \
  ) else output;

\end{verbatim}
\end{tiny}

Again, see the code for full details\footnote{\begin{footnotesize}word-level token Restricted Access Sequence Processing solution: \href{https://github.com/willy-b/RASP-for-COGS}{https://github.com/willy-b/RASP-for-COGS} (COGS) and \href{https://github.com/willy-b/learning-rasp}{https://github.com/willy-b/learning-rasp} (ReCOGS) \end{footnotesize}} (for simplicity this description was also written without discussing sentential complement handling).

For all steps only the RASP outputs aligned with the input sequence (Encoder part of derived Transformer) or the very last sequence output (for next token in autoregressive generation) were used. For convenience of reading the aggregate operator was usually used acausally to assign all sequence outputs before the last one to the same value as the last (so only one value would be displayed).

For RASP-for-COGS:
\begin{tiny}
\begin{verbatim}
git clone https://github.com/willy-b/RASP-for-COGS.git
cd RASP-for-COGS
python cogs_examples_in_rasp.py
\end{verbatim}
\end{tiny}

The script will show performance on COGS training data by default, run with "--use\_dev\_split", "--use\_gen\_split" , or "--use\_test\_split" to see it run on those and give a running score every row.

And for RASP-for-ReCOGS:
\begin{tiny}
\begin{verbatim}
git clone https://github.com/willy-b/learning-rasp.git
cd learning-rasp
python recogs_examples_in_rasp.py 
\end{verbatim}
\end{tiny}

\clearpage

\subsection{Note on a Restricted Access Sequence Processing character-level token program / model design (NOT what is used in this paper but feasible)}
\label{rasp_character_level_model_notes}

Note, a proof of concept character level Restricted Access Sequence Processing model was done with a decoder loop (unlike word-level solution above, it was a sketch so did not limit to strictly causal operations which just require more careful indexing -- using the value at the separator or the end of a word instead of pooling the same value to all letters in a word for example). Note that this one did not cover translating sentences in general into ReCOGS unlike the word-level solution as it is tedious and redundant but the core operations are possible and the author believes any solution at the word level can be mapped to a solution in character level tokens (out of scope for this paper to prove it).

Since it is a separate problem and adds a lot of complexity without bringing anything to bear on the main questions of the paper, I left a full implementation to the word-level tokens which were simpler and ran faster. The difference is one uses a similar approach started at \footnote{\begin{footnotesize}See the other-examples/decoder-loop-example-parse-into-recogs-style-variables.rasp file.\end{footnotesize}} to assign all the letters in each word an index. 

Word indices can be assigned using RASP to count separators occurring prior to each sequence location like:

(we also zero out the word index for the separators themselves)

\begin{tiny}
\begin{verbatim}
word_indices = \
(1+selector_width(select(tokens, " ", ==) \
and select(indices, indices, <=))) \
*(0 if indicator(tokens == " ") else 1);
\end{verbatim}
\end{tiny}

Then one can do an aggregation of the letters grouping by word index (this, which is NOT part of the techniques used in this paper for the word-level tokens solution, requires additional work (tedious not challenging) to do causally outside the input (in the decoder), one must sum forward so the word representation is always at the last letter of the word or separator instead of at all letters of the word, and that step is left out of the character-level demo and this discussion; whereas the word-level solution described above has a clear Encoder Decoder separation. This can be done so that the value which is then the same for all letters in each word, is unique to each word in the dictionary and can be looked up in a map to get word level attributes like part-of-speech and get back to the solution in the word-level tokens in Appendix \ref{rasp-word-level-model-design} which was fully implemented. A simple approach (not necessarily recommended but works for proof of concept) that would work for small vocabularies (easily extended) is to use a map to lookup each letter of the alphabet to a log prime. Then the sum of the letters in a word (grouped by the word index which is the count of spaces/separators prior) is the sum of the log primes indexed by the alphabet index. Since the sum of logarithms of numbers is the same as the logarithm of the product of those numbers, this is equivalent to the logarithm of the product of a series of primes. Each prime in the product corresponds 1-to-1 to a letter in the alphabet, with the number of occurrences in the product corresponding to the number of times that letter occurs in the word. By uniqueness of prime number factorization this would map each multiset of letters to a single unique sum of log primes. Thus if you do not have words which are anagrams, all the letters in each word would be assigned a number that uniquely represented that word in the vocabulary. If you have anagrams you can do this step and then take the first and last letter and compute a separate number from that and add it to all the letters in the word.

Example lookup table for letters before aggregating by word index (not recommended but for proof of concept that one can go from character level tokens to word-specific numbers which can then be looked up as in the word-level token solution in Appendix \ref{rasp-word-level-model-design} used throughout the paper):
\clearpage
\begin{tiny}
\begin{verbatim}
def as_num_for_letter_multiset_word_pooling(t) {
    # To be multiset unique, need logarithm of prime so that the sum aggregation
    # used in RASP corresponds to prime number factorization (sum of logs of primes is same as log of product of primes) 
    # (we can do sum aggregation instead of mean by multiplying by length)
    # However RASP does not appear to support logarithms (underlying multilayer 
    # perceptron can learn to approximate logarithms)
    #letter_to_prime_for_multiset_word_pooling = {"a": 2, "b": 3, "c": 5, "d": 7, 
    #"e": 11, "f": 13, "g": 17, "h": 19, "i": 23, "j": 29, "k": 31, "l": 37, 
    #"m": 41, "n": 43, "o": 47, "p": 53, "q": 59, "r": 61, "s": 67, "t": 71, 
    #"u": 73, "v": 79, "w": 83, "x": 89, "y": 97, "z": 101, ".": 0, 
    #" ": 0, ":": 0};
    map_letter_to_log_prime_for_pooling = {"a": 0.6931471805599453, "b": 1.0986122886681098, 
    "c": 1.6094379124341003, "d": 1.9459101490553132, "e": 2.3978952727983707, 
    "f": 2.5649493574615367, "g": 2.833213344056216, "h": 2.9444389791664403, 
    "i": 3.1354942159291497, "j": 3.367295829986474, "k": 3.4339872044851463, 
    "l": 3.6109179126442243, "m": 3.713572066704308, "n": 3.7612001156935624, 
    "o": 3.8501476017100584, "p": 3.970291913552122, "q": 4.07753744390572, 
    "r": 4.110873864173311, "s": 4.204692619390966, "t": 4.2626798770413155, 
    "u": 4.290459441148391, "v": 4.3694478524670215, "w": 4.418840607796598, 
    "x": 4.48863636973214, "y": 4.574710978503383, "z": 4.61512051684126,
    # we zero out tokens we want not to affect the identity of the word
    ".": 0, " ": 0, ":": -1, "(": -1, ")": -1, "0": -1, "1": -1, "2": -1, 
    "3": -1, "4": -1, "5": -1, "6": -1, "7": -1, "8": -1, "9": -1, ";": -1, 
    ",": -1};
    return map_letter_to_log_prime_for_pooling[t];
}
\end{verbatim}
\end{tiny}

Pooling by word can then be done with:
\begin{tiny}
\begin{verbatim}
pseudoembeddedwords = \
aggregate(select(word_indices, word_indices, ==), \
as_num_for_letter_multiset_word_pooling(tokens))*word_lengths;
\end{verbatim}
\end{tiny}
(Per-character token example is not causally masked, we do causal strict-decoder-compatible solution for anything outside input sequence in the full word-level solution above just leaving out of this character-level sketch, which is NOT used in this paper. For the causal character level solution one would use the summed value at the end of the word or the separator instead, indexing relative to separators.) 

Those values could then be looked up in a dictionary like in the completed word-level token solution to get part-of-speech, verb-type, etc, to derive a separate sequence which can be used for template matching as we successfully did with word-level tokens (see Appendix \ref{rasp-word-level-model-design}).

\clearpage
\subsection{Model Detail}
\label{model_detail}
Model code is available: \href{https://github.com/willy-b/RASP-for-COGS}{RASP-for-COGS} and \href{https://github.com/willy-b/learning-rasp}{RASP-for-ReCOGS}\footnote{\begin{footnotesize}\href{https://github.com/willy-b/RASP-for-COGS}{https://github.com/willy-b/RASP-for-COGS} (COGS) and \href{https://github.com/willy-b/learning-rasp}{https://github.com/willy-b/learning-rasp} (ReCOGS)\end{footnotesize}}.

For our Restricted Access Sequence Processing ReCOGS program, we used the RASP interpreter of \citep{Weiss2021} to run our program. For RASP model design and details see Appendix \ref{rasp-word-level-model-design}. We use word-level tokens for all RASP model results in this paper.\footnote{\begin{footnotesize}We believe any solution at the word-level can be converted to a character-level token solution and that is not the focus of our investigation here (see Appendix \ref{rasp_character_level_model_notes} for proof of concept details on a character level solution not used here).\end{footnotesize}}
Consistent with \citep{Zhou2024} we use \citep{Weiss2021}'s RASP originally used for modeling Transformer encoders to model an encoder-decoder in a causal way by feeding the autoregressive output back into the program. We only have aggregations with non-causal masks when that aggregation (or without loss of generality just before the aggregation product is used to avoid multiplying everywhere) is masked by an input mask restricting it to the sequence corresponding to the input.

We used RASP maps to map word level tokens to part-of-speech and verb-type which is consistent with what can be learned in embeddings or the earliest layer of a Transformer \citep{tenney2019bertrediscoversclassicalnlp}\footnote{\citep{tenney2019bertrediscoversclassicalnlp} report part-of-speech information is already tagged in layer 0 (post-embedding) of the 24-layer BERT large pre-trained language model, trained using a masked language modeling objective. Though models for COGS/ReCOGS are usually trained using a sequence-to-sequence (seq2seq) objective (whether that objective biases the Transformer to learn the same representation on this task is not known to our knowledge), one could also use a language modeling objective to model the COGS input text and its associated logical form output (not just the output conditioned on the input). See \citep{10.1162/tacl_a_00733} for examples of solving the same language tasks using seq2seq vs various language modeling objectives - they indeed find better generalization performance on their problems when using the language modeling objective (training to model both the input and the output).} and then did 19 different attention-head based template matches on that flat sequence\footnote{\begin{footnotesize}A flat/non-tree solution was pursued because it was simple and given the failure documented in \citep{Wu2023} of the baseline Encoder-Decoder to generalize from obj\_pp\_to\_subj\_pp and other evidence we give below we shall see it is hard to argue a tree-based solution which includes the rule `np\_det pp np -> np\_pp -> np` is learned by \citep{Wu2023}'s baseline Encoder-Decoder Transformer.\end{footnotesize}} (no tree-based parsing, no recursive combination of terminals/non-terminals.) Those 19 templates were constructed using grammar coverage \citep{fuzzingbook2023:GrammarCoverageFuzzer} to cover the (Re)COGS input grammar as demonstrated in the training data (see "Appendix: Restricted Access Sequence Processing word-level token program/model design" (\ref{rasp-word-level-model-design}), and see Tables \ref{RASP-model-flat-patterns-after-masking-to-nv-relationships-table} , \ref{specific-grammar-pattern-examples-mapped-to-part-of-speech-and-cogs-in-distribution-training-examples} for patterns and equivalent ReCOGS training examples).\footnote{\begin{footnotesize}To handle prepositional phrases in a flat solution, we find it necessary on the training data to add a rule that ignores "det common\_noun" or "proper noun" preceded by a preposition when searching for noun indexes to report in relationships (agent, theme, recipient, etc) and as if we did that during pattern matching by using before/after matches instead of strict relative indexing.\end{footnotesize}}.

For the vocabulary we used the \citep{klinger2024compositionalprogramgenerationfewshot} description of COGS in their utilities\footnote{\begin{footnotesize}https://github.com/IBM/cpg/blob/
c3626b4e03bfc681be2c2a5b23da0b48abe6f570
/src/model/cogs\_data.py\#L523\end{footnotesize}} (same input as ReCOGS) (NOT using their CPG solution or model anywhere) in constructing our RASP vocabulary and part-of-speech or verb-type embedding/mapping. 

We are focused on structural, not lexical generalizations, so same as in \citep{klinger2024compositionalprogramgenerationfewshot} we include all words occurring anywhere in the upstream (Re)COGS "train.tsv" (including "exposure" rows, though would not change results qualitatively to omit the very few words only occurring in exposure examples). For RASP-for-ReCOGS only, we also include two words in our vocab/embedding as common nouns accidentally left out of train.tsv vocabulary by the COGS author (COGS vocab is used by ReCOGS): "monastery" and "gardner" (only included in their train\_100.tsv and dev.tsv not also in train.tsv, but present in test/gen), a decision affecting just 22 or 0.1\% of generalization examples so would not affect any conclusions qualitatively. For RASP-for-COGS which was started after RASP-for-ReCOGS, we considered these words out-of-vocabulary to be more conservative. See also the discussion on COGS Github with the COGS author at \href{https://github.com/najoungkim/COGS/issues/2\#issuecomment-976216841}{https://github.com/najoungkim/COGS/issues/2
\#issuecomment-976216841}.

For training the baseline Transformers from scratch with randomly initialized weights using gradient descent for comparison with RASP predictions, we use scripts derived from those provided by \citep{Wu2023}\footnote{\begin{footnotesize}\href{https://github.com/frankaging/ReCOGS/blob/1b6eca8ff4dca5fd2fb284a7d470998af5083beb/run\_cogs.py}{https://github.com/frankaging/ReCOGS/blob/
1b6eca8ff4dca5fd2fb284a7d470998af5083beb/run\_cogs.py}

and

\href{https://github.com/frankaging/ReCOGS/blob/1b6eca8ff4dca5fd2fb284a7d470998af5083beb/model/encoder\_decoder\_hf.py}{https://github.com/frankaging/ReCOGS/blob/
1b6eca8ff4dca5fd2fb284a7d470998af5083beb
/model/encoder\_decoder\_hf.py}\end{footnotesize}}.

The baseline \citep{Wu2023} Encoder-Decoder Transformer was by default 2-layers with 4344077 parameters,
except for the layer variation experiments which had 6046701 parameters for the 3-layer , and 7749325 parameters for the 4-layer variations.
We did not control the parameter count as discussed earlier as even allowing it to increase, the additional layers did not result in improved performance on the obj-pp-to-subj-pp split (see results at "\citep{Wu2023} Encoder-Decoder baseline 2-layer Transformer does not improve on the obj\_pp\_to\_subj\_pp split when adding 1 or 2 additional layers" (\ref{wu-baseline-layer-variation-experiment-results})). If there had been an improvement, we would have run additional experiments to increase depth while matching parameter count.

For ease of reference, the model architecture generated by the \citep{Wu2023} baseline Encoder-Decoder Transformer script (trained from scratch, not pretrained) is as follows with N BertLayers set to 2 per \citep{Wu2023} for all baseline experiments except the layer variation experiments:
\begin{tiny}
\begin{verbatim}
# For Wu et al 2023 Encoder-Decoder Transformer baselines 
# (we predict and analyze errors made by these 
# in the paper using what we learned about how Transformers 
# can perform the task from the 
# Restricted Access Sequence Processing model),
# we use the official scripts provided at 
# https://github.com/frankaging/ReCOGS/blob/
# 1b6eca8ff4dca5fd2fb284a7d470998af5083beb/run\_cogs.py
# and 
# https://github.com/frankaging/ReCOGS/blob/
# 1b6eca8ff4dca5fd2fb284a7d470998af5083beb/
# model/encoder\_decoder\_hf.py
# where the architecture generated is as follows:
EncoderDecoderModel(
 (encoder): BertModel(
  (embeddings): BertEmbeddings(
   (word_embeddings): Embedding(762, 300, padding_idx=0)
   (position_embeddings): Embedding(512, 300)
   (token_type_embeddings): Embedding(2, 300)
   (LayerNorm): LayerNorm((300,), eps=1e-12, 
     elementwise_affine=True)
   (dropout): Dropout(p=0.1, inplace=False)
  )
  (encoder): BertEncoder(
   (layer): ModuleList(
    # substitute N=2 for all baseline experiments
    # per Wu et al 2023 paper; 
    # N can be 3 or 4 in our layer variation 
    # experiments only.
    (0-(N-1)): N x BertLayer(
     (attention): BertAttention(
      (self): BertSdpaSelfAttention(
       (query): 
        Linear(in_features=300, out_features=300, bias=True)
       (key): 
        Linear(in_features=300, out_features=300, bias=True)
       (value): 
        Linear(in_features=300, out_features=300, bias=True)
       (dropout): Dropout(p=0.1, inplace=False)
      )
      (output): BertSelfOutput(
       (dense): 
        Linear(in_features=300, out_features=300, bias=True)
       (LayerNorm): 
        LayerNorm((300,), eps=1e-12, elementwise_affine=True)
       (dropout): Dropout(p=0.1, inplace=False)
      )
     )
     (intermediate): BertIntermediate(
      (dense): 
       Linear(in_features=300, out_features=512, bias=True)
      (intermediate_act_fn): GELUActivation()
     )
     (output): BertOutput(
      (dense): 
       Linear(in_features=512, out_features=300, bias=True)
      (LayerNorm): 
       LayerNorm((300,), eps=1e-12, elementwise_affine=True)
      (dropout): Dropout(p=0.1, inplace=False)
     )
    )
   )
  )
  (pooler): BertPooler(
   (dense): 
    Linear(in_features=300, out_features=300, bias=True)
   (activation): Tanh()
  )
 )
 (decoder): BertLMHeadModel(
  (bert): BertModel(
   (embeddings): BertEmbeddings(
    (word_embeddings): Embedding(729, 300, padding_idx=0)
    (position_embeddings): Embedding(512, 300)
    (token_type_embeddings): Embedding(2, 300)
    (LayerNorm): 
     LayerNorm((300,), eps=1e-12, elementwise_affine=True)
    (dropout): Dropout(p=0.1, inplace=False)
   )
   (encoder): BertEncoder(
    (layer): ModuleList(
     # substitute N=2 for all baseline experiments
     # per Wu et al 2023 paper; 
     # N can be 3 or 4 in our layer variation
     # experiments only.
     (0-(N-1)): N x BertLayer(
      (attention): BertAttention(
       (self): BertSdpaSelfAttention(
        (query): 
         Linear(in_features=300, out_features=300, bias=True)
        (key): 
         Linear(in_features=300, out_features=300, bias=True)
        (value): 
         Linear(in_features=300, out_features=300, bias=True)
        (dropout): Dropout(p=0.1, inplace=False)
       )
       (output): BertSelfOutput(
        (dense): 
         Linear(in_features=300, out_features=300, bias=True)
        (LayerNorm): 
         LayerNorm((300,), eps=1e-12, elementwise_affine=True)
        (dropout): Dropout(p=0.1, inplace=False)
       )
      )
      (crossattention): BertAttention(
       (self): BertSdpaSelfAttention(
        (query): 
         Linear(in_features=300, out_features=300, bias=True)
        (key): 
         Linear(in_features=300, out_features=300, bias=True)
        (value): 
         Linear(in_features=300, out_features=300, bias=True)
        (dropout): Dropout(p=0.1, inplace=False)
       )
       (output): BertSelfOutput(
        (dense): 
         Linear(in_features=300, out_features=300, bias=True)
        (LayerNorm): 
         LayerNorm((300,), eps=1e-12, elementwise_affine=True)
        (dropout): Dropout(p=0.1, inplace=False)
       )
      )
      (intermediate): BertIntermediate(
       (dense): 
        Linear(in_features=300, out_features=512, bias=True)
       (intermediate_act_fn): GELUActivation()
      )
      (output): BertOutput(
       (dense): 
        Linear(in_features=512, out_features=300, bias=True)
       (LayerNorm): 
        LayerNorm((300,), eps=1e-12, elementwise_affine=True)
       (dropout): Dropout(p=0.1, inplace=False)
      )
     )
    )
   )
  )
  (cls): BertOnlyMLMHead(
   (predictions): BertLMPredictionHead(
    (transform): BertPredictionHeadTransform(
     (dense): 
      Linear(in_features=300, out_features=300, bias=True)
     (transform_act_fn): GELUActivation()
     (LayerNorm): 
      LayerNorm((300,), eps=1e-12, elementwise_affine=True)
    )
    (decoder): Linear(in_features=300, out_features=729,
      bias=True)
   )
  )
 )
)
\end{verbatim}
\end{tiny}

For the \citep{Wu2023} baseline Encoder-Decoder Transformer layer variation experiments, 
when we say e.g. 3 or 4 layers, we refer to 3 or 4 x BertLayer in the Encoder and Decoder, setting (3 or 4 Transformer blocks).
(This is intended because only once per block, during cross/self-attention is information exchanged between sequence positions, and \citep{Csordas2022} hypothesize that the number of such blocks must be at least the depth of the parse tree in a compositional solution, as in a grammar parse tree at each level symbols are combined which requires transferring information between sequence positions).

\subsection{Methods Detail}
\label{methods_detail}

We use the RASP \citep{Weiss2021} interpreter\footnote{\begin{footnotesize}provided at \href{https://github.com/tech-srl/RASP/}{https://github.com/tech-srl/RASP/}
\end{footnotesize}
} to evaluate our RASP programs\footnote{\begin{footnotesize}\href{https://github.com/willy-b/RASP-for-COGS}{https://github.com/willy-b/RASP-for-COGS} (COGS) and \href{https://github.com/willy-b/learning-rasp}{https://github.com/willy-b/learning-rasp} (ReCOGS)\end{footnotesize}}, follow instructions in README.md for demos.

We implement in RASP the transformation of COGS input sentences into the COGS logical form (RASP-for-COGS), as well as separately the ReCOGS\_pos\footnote{
\begin{footnotesize}We use the ReCOGS positional index data (rather than default ReCOGS with randomized indices) as it has consistent position based indices that allow us to perform well on Exact Match (like the original COGS task) as well as Semantic Exact Match (which ignores absolute values of indices).

See ReCOGS\_pos dataset at 

\href{https://github.com/frankaging/ReCOGS/tree/1b6eca8ff4dca5fd2fb284a7d470998af5083beb/recogs\_positional\_index}{https://github.com/frankaging/ReCOGS/tree/
1b6eca8ff4dca5fd2fb284a7d470998af5083beb
/recogs\_positional\_index}\end{footnotesize}
} logical form (LF) (RASP-for-ReCOGS).

RASP-for-COGS COGS LF output is scored by String Exact Match against ground truth. Note that String Exact Match implies also Semantic Exact Match but not vice versa (String Exact Match is a stricter type of match).
RASP-for-ReCOGS (ReCOGS logical form) is scored by Semantic Exact Match\footnote{ 
\begin{footnotesize}\href{https://github.com/frankaging/ReCOGS/blob/1b6eca8ff4dca5fd2fb284a7d470998af5083beb/utils/train_utils.py}{https://github.com/frankaging/ReCOGS/blob/
1b6eca8ff4dca5fd2fb284a7d470998af5083beb/utils/
train\_utils.py}

and

\href{https://github.com/frankaging/ReCOGS/blob/1b6eca8ff4dca5fd2fb284a7d470998af5083beb/utils/compgen.py}{https://github.com/frankaging/ReCOGS/blob/
1b6eca8ff4dca5fd2fb284a7d470998af5083beb/
utils/compgen.py}
\end{footnotesize}
} against ground truth.

In the training data only, any ReCOGS training augmentations like preposing or "um" sprinkles are excluded when evaluating the RASP model on the train data (it does not learn directly from the examples and these augmentations are outside of the grammar). 

We also measure grammar coverage of input examples supported by our RASP model against the full grammar of COGS/ReCOGS input sentences provided in the utilities of the IBM CPG project \citep{klinger2024compositionalprogramgenerationfewshot}\footnote{\begin{footnotesize}https://github.com/IBM/cpg/blob/
c3626b4e03bfc681be2c2a5b23da0b48abe6f570
/src/model/cogs\_data.py\#L523\end{footnotesize}
}

When computing grammar coverage \citep{fuzzingbook2023:GrammarCoverageFuzzer}, we collapse all vocabulary terminals (leaves) to a single terminal (leaf), ignoring purely lexical differences (see "Appendix: Computing Grammar Coverage" (\ref{computing_grammar_coverage}) for details and motivation).

The overall String Exact Match (RASP-for-COGS, stricter, implies Semantic Exact Match as well) and Semantic Exact Match (RASP-for-ReCOGS) performance is reported as well as the performance on the specific structural generalization splits where Transformers are reported to struggle, even in ReCOGS, specifically Object Prepositional Phrase to Subject Prepositional Phrase (obj\_pp\_to\_subj\_pp),
Prepositional Phrase (pp\_recursion) are highlighted and discussed in depth for all models.

For the RASP program's String/Semantic Exact Match results which are based on the outcome of a deterministic program (so cannot randomly reinitialize weights and retrain, rerun),
we can use the Beta distribution to model the uncertainty and generate confidence intervals (Clopper-Pearson intervals\footnote{\begin{footnotesize}see e.g. \href{https://en.wikipedia.org/w/index.php?title=Binomial\_proportion\_confidence\_interval\&oldid=1252517214\#Clopper\%E2\%80\%93Pearson\_interval}{https://en.wikipedia.org/w/index.php?title=
Binomial\_proportion\_confidence\_interval\&oldid=1252517214
\#Clopper\%E2\%80\%93Pearson\_interval}
and
\href{https://arxiv.org/abs/1303.1288}{https://arxiv.org/abs/1303.1288} 
\end{footnotesize}
}) as each String/Semantic Exact Match is a binary outcome (0 or 1 for each example). Unlike bootstrapping this also supports the common case for our RASP program of 100\% accuracy, which occurs in all but one split, where resampling would not help us estimate uncertainty in bootstrapping, but using the Beta distribution gives us confidence bounds that depend on the sample size.

In developing our RASP program\footnote{\begin{footnotesize}\href{https://github.com/willy-b/RASP-for-COGS}{https://github.com/willy-b/RASP-for-COGS} (COGS) and \href{https://github.com/willy-b/learning-rasp}{https://github.com/willy-b/learning-rasp} (ReCOGS)\end{footnotesize}},
when we find the right index of a verb relation (like agent, theme, or recipient), 
on the center embedded pp training examples we found it was necessary to skip any noun phrases preceded by a preposition ("in", "on", "beside")\footnote{RASP code in "Appendix: RASP for relation right index ignoring attractor 'pp np'" (\ref{RASP_for_relation_right_index_ignoring_attractor_pp_np})}.\footnote{\begin{footnotesize}
For example, when masking the center-embedded pp ("on a tree") in COGS/ReCOGS training sentence “Isabella forwarded a box on a tree to Emma" transforming the sentence to “Isabella forwarded a box \_ to Emma” allows us to have noun-verb relationships extracted the same as other non-pp modified v\_dat\_p1 examples in Tables \ref{RASP-model-flat-patterns-after-masking-to-nv-relationships-table} , \ref{specific-grammar-pattern-examples-mapped-to-part-of-speech-and-cogs-in-distribution-training-examples} by simply mapping the 3 nouns (Isabella, box, Emma) to agent, theme, recipient in order (same for all v\_dat\_p1).
It turns out, this resolves the obj-pp-to-subj-pp problem as well, as otherwise, when modifying a simple sentence like "The cake burned" with a preposition to "The cake on the plate burned" we would switch the theme from the cake to the plate by accident. This cake example is the infamous obj pp to subj pp example, where training a Transformer successfully to represent the semantics of sentences like "John ate the cake on the plate" leads to a model that won't immediately generalize to being able to represent the meaning of "The cake on the plate burned" in logical form.
In writing our RASP program this was observed as nothing to do with subjects or objects but just modifying noun phrases to the left of the part of speech (say a verb) they have a relationship with, instead of on the right side (center vs tail embedded pp). For example, this also occurs in v\_dat\_p2 sentences like "Emma gave a friend a cookie" (agent, recipient, theme nps). It is obvious that modification of the theme with prepositional phrases is not going to disrupt parsing the sentence: "Emma gave a friend a cookie (modification modification ...)", whereas modifying the recipient, on the left, due to the asymmetry of prepositional phrases adding to the right, disrupts the sentence, rendering it unreadable in the limit of too many pps:

"Emma gave a friend (modification modification ...) a cookie" , in the limit of more modification, "a friend" cannot be associated with "a cookie".

Center-embedded depth 2 prepositional phrase examples do occur in the (Re)COGS training set, e.g. "Camila was forwarded the cookie in a house beside the computer by Olivia", where our model succeeds by masking out "in a house beside the computer" to obtain "Camila was forwarded the cookie \_ by Olivia" where (Camila, cookie, Olivia) can be read out using the standard v\_dat\_pp\_p4 template (recipient, theme, agent), as if it were e.g. the non-pp modified v\_dat\_pp\_p4 example from Tables \ref{RASP-model-flat-patterns-after-masking-to-nv-relationships-table} , \ref{specific-grammar-pattern-examples-mapped-to-part-of-speech-and-cogs-in-distribution-training-examples}, "The girl was lended the balloon by Harper" (girl, balloon, Harper) similarly read out (recipient, theme, agent).
\end{footnotesize}}

Since in the RASP program both this and subject prepositional phrase modification require the same rule ignoring the "pp np" when finding right index candidates for agent, theme, recipient outputs, we hypothesized two things.

One, that `np v\_dat\_p2 np pp np`\footnote{\begin{footnotesize}Being precise we only do `np v\_dat\_p2 np\_det pp np np` as per the grammar `np\_prop` cannot precede a prepositional phrase\end{footnotesize}} generalization after training on `np v\_dat\_p2 np np pp np` would be difficult like \citep{Wu2023}'s obj\_pp\_to\_subj\_pp split.

Two, that augmenting the training data with v\_dat\_p2 recipient modified sentences like "Emma gave a friend in a house a cookie" might lead to crossover improved performance on the subject pp generalization (e.g. "The friend in a house smiled"; without adding any example of subjects with pp modification).

Thus we additionally train \citep{Wu2023} baseline Transformers from scratch in two separate experiments to test these.

For one, `np v\_dat\_p2 np pp np np`\footnote{\begin{footnotesize}Restricted to `np v\_dat\_p2 np\_det pp np np` as per the grammar `np\_prop` cannot precede a prepositional phrase\end{footnotesize}} generalization after training on `np v\_dat\_p2 np np pp np` we train \citep{Wu2023} Transformers with default configuration and default training data, then we add a new generalization split derived from \citep{Wu2023}'s `train.tsv` of 328 existing training examples where we have transferred the prepositional phrase from the theme to the recipient\footnote{\begin{footnotesize}When the recipient is np\_det, not np\_prop; and we confirm it is within the grammar by reparsing with the Lark parser on the original grammar rules.\end{footnotesize}} in the `v\_dat\_p2` sentence form with one prepositional phrase (see Appendix \ref{v_dat_p2_recipient_pp-modification_for_generalization_assessment_and_data_augmentation_attempt} for details and link to actual data sample).

For two, to see if augmenting the training data with v\_dat\_p2 recipient modified sentences has crossover benefit, we train separate default \citep{Wu2023} Transformer but with their existing train.csv plus the additional theme-modified sentences mentioned above, same as those used for generalization testing in the other experiment; we confirm it does not know them, and separately on fresh runs we try training on them to see if that can benefit other splits by teaching the Encoder-Decoder a general prepositional phrase handling rule (like ignore "pp np"). We then test on \citep{Wu2023}'s normal test and generalization splits.

\citep{Wu2023} baseline Encoder-Decoder Transformers trained from scratch are trained with random weight initialization multiple times with at least 10 different random seeds with all performance metrics averaged across runs with sample mean, sample size, and unbiased sample standard deviation reported. Statistical significance of comparisons between any Transformers performance sample means are checked with Welch's unequal variance t-test with p-values greater than 0.05 definitely rejected, though stricter thresholds may be used where applicable. Confidence intervals are reported using 1.96 standard errors of the sample mean as the 95\% confidence interval for sample means with that N unless specified otherwise.

See also "Appendix: Results Notebook links by section" (\ref{results_notebook_links_by_section}) for notebooks documenting results and giving steps to reproduce.

See also "Appendix: Scientific Artifacts - Is Our Use Consistent with Authors' Intention and Licensing" (\ref{scientific_artifacts_use}).

\subsection{Attraction errors}
\label{attraction_errors}

\begin{figure}
\includegraphics[scale=1]{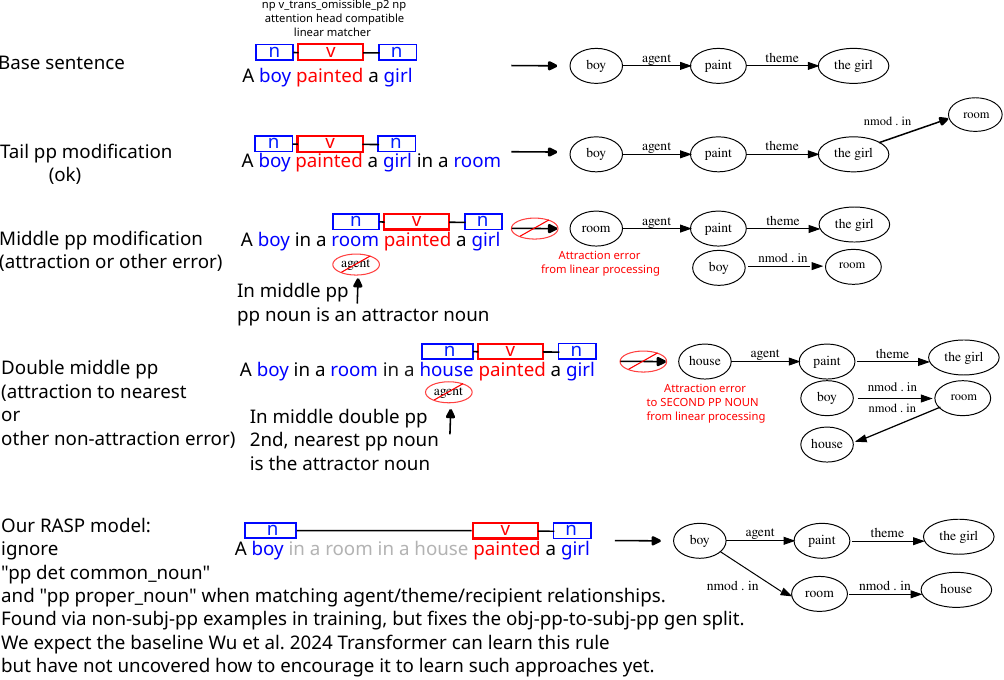}
\onecolumn
\caption{\begin{small}Non-hierarchical/non-tree structured linear grammar pattern matching without explicitly ignoring prepositional phrase nouns is expected to give rise to attraction errors, which we confirmed are contributing to the high error rate of the baseline \citep{Wu2023} Transformer on the obj-pp-to-subj-pp generalization split. Our RASP model avoids these errors by masking "pp det common\_noun" and "pp proper\_noun" when matching for agent/theme/recipient relationships (see also Figures \ref{rasp-for-recogs-decoder-loop-supplementary-figure_pp_depth_2}, \ref{rasp-for-cogs-decoder-loop-figure_pp_depth_2}) (a behavior added based on non-subj-pp examples in training behavior but shown here to generalize to those examples). Note that we also predict such errors for the non-subj-pp case of pp-modifying the right-of-verb recipient noun in "np v\_dat\_p2 np np" and confirmed (see Figure \ref{baseline_transformer_standard_training_cannot_do_v_dat_p2_generalization}) that such a generalization is as hard as the previously reported hardest obj-pp-to-subj-pp generalization.\end{small} \\
}
\twocolumn
\label{attraction_errors_figure}
\end{figure}

\textbf{See Figure} \ref{attraction_errors_figure}.

In this paper we predict and confirm the existence of errors on prepositional modification splits where putting one or more new prepositional phrase nouns between a noun of interest and a verb it is related to causes the relation to inappropriately jump to one of the new nearer "attractor" nouns. 

\textbf{For overall attraction error results by the baseline Transformer see results section "Attraction Error Analysis for \citep{Wu2023} baseline Encoder-Decoder Transformer on obj\_pp\_to\_subj\_pp split" (\ref{error_analysis_for_baseline_transformer_predict_and_confirm_attraction_errors}).}

For lack of a better term I am referring to this as an "attraction" error following \citep{jespersen1913modernenglishgrammar1954reprint} section 6.72 "Attraction" in the context of subject-verb agreement, describing a similar "error" made by humans:

"Very frequently in speech, and not infrequently in
literature, the number of the verb \textbf{is determined by
that part of the subject which is nearest to the verb},
even if a stricter sense of grammar would make
the verb agree with the main part of the subject.
\textbf{This kind of attraction naturally occurs the more easily, the greater the distance is between the nominative and the verb.}"

The term attraction error continues to be used to describe those errors by psycholinguists, e.g. \citep{FRANCK2006173} who in the context of subject-verb agreement, define attraction errors as "incorrect agreement with a word that is not the subject of the sentence".
Those attraction errors are also used to study hierarchical vs linear language processing (in humans, see \citep{FRANCK2006173} and also \citep{VIGLIOCCO1998B13}; in language models as we discuss here, see also \citep{goldberg2019assessingbertssyntacticabilities} who states that successful subject-verb agreement in the presence of attractor nouns  "[is] traditionally taken as evidence for the existence [of] hierarchical structure"), similar to our investigation here. But we are not investigating or explaining grammatical attraction in general here, just predicting and documenting a particular error the baseline Transformers make as a prediction of a non-hierarchical, non-tree structured approach without a rule for ignoring intervening prepositional phrase nouns.

We specifically hypothesized attraction to the nearest noun (when there is more than one "attractor" noun unrelated to the verb added in-between the related noun and the verb), but the relationship jumping to any of those new "attractor" nouns would be an "attraction" error in this terminology.

Here are two real examples made by the \citep{Wu2023} baseline Encoder-Decoder Transformer with different prepositional recursion depths.

e.g. for pp depth 1, the mistake (as we expect from attraction to the nearest noun hypothesis) is to put e.g. agent index 4 here instead of 1:

input: The baby beside a valve painted the cake .

actual:   * baby ( 1 ) ; valve ( 4 ) ; * cake ( 7 ) ; nmod . beside ( 1 , 4 ) AND paint ( 5 ) AND agent ( 5 , 4 ) AND theme ( 5 , 7 )

expected: * baby ( 1 ) ; valve ( 4 ) ; * cake ( 7 ) ; nmod . beside ( 1 , 4 ) AND paint ( 5 ) AND agent ( 5 , 1 ) AND theme ( 5 , 7 )

whereas e.g. for pp depth 2 on the agent left of the verb, as expected the mistake is to put agent index 7 instead of 1 below (the pp noun closest to the verb steals it, not the other pp noun at index 4):

input: A girl on the stool on the table drew a frog .

actual:   girl ( 1 ) ; * stool ( 4 ) ; * table ( 7 ) ; frog ( 10 ) ; nmod . on ( 1 , 4 ) AND nmod . on ( 4 , 7 ) AND draw ( 8 ) AND agent ( 8 , 7 ) AND theme ( 8 , 10 )

expected: girl ( 1 ) ; * stool ( 4 ) ; * table ( 7 ) ; frog ( 10 ) ; nmod . on ( 1 , 4 ) AND nmod . on ( 4 , 7 ) AND draw ( 8 ) AND agent ( 8 , 1 ) AND theme ( 8 , 10 )

We went looking for this hypothesizing that the \citep{Wu2023} Transformer may be using flat attention-head compatible verb-centered pattern matching as we are in our RASP model,
and without learning the the single rule in our RASP program to ignore "pp det common\_noun" and "pp proper\_noun" were not learned by the Transformer (as our RASP model has "attraction" errors without it). Without the rule for avoiding "attraction" errors, we supposed the actual attention-head compatible verb-centered pattern matched noun (closer to the verb than the actual agent) for a grammar pattern would labeled the agent or theme instead of the appropriate one.

Note that \citep{vanschijndel2019quantitydoesntbuyquality} also see "attraction" errors by Transformers/RNNs (again in the context of subject-verb agreement) where a long-range dependency competes with attractors/distractors, finding "accuracy decrease[d] in the presence of distracting nouns intervening between the head of the subject and the verb".

The "attraction" errors we report here where attractor/distractor prepositional phrase nouns replace the actual agent/subject in the ReCOGS logical form generated by \citep{Wu2023} baseline Transformers are NOT due to their presence in pre-training or training data, as the (Re)COGS training data is synthetic and syntactically perfect and for this benchmark the Transformer is trained from scratch, so it a genuine new error made by the neural network itself (and we predict a mechanism using RASP). But in general, humans do also exhibit these "attraction" errors, e.g. again as discussed in human subject-verb agreement per \citep{jespersen1913modernenglishgrammar1954reprint} (see quote earlier in this section), so pre-trained models trained on human-generated text may have the additional problem of learning those errors from the training data itself. Language model tendencies to commit subject-verb agreement attraction errors were previously analyzed by a co-author of the RASP language in an earlier paper on BERT Transformers in \citep{goldberg2019assessingbertssyntacticabilities}, by a COGS benchmark co-author in \citep{vanschijndel2019quantitydoesntbuyquality}, and by both together regarding RNNs in \citep{linzen2016assessing} (whose reference to \citep{agreementwithnearestlanguagelog} led me to \citep{jespersen1913modernenglishgrammar1954reprint}).

\clearpage
\clearpage

\subsection{RASP for relation right index ignoring attractor "pp np"}
\label{RASP_for_relation_right_index_ignoring_attractor_pp_np}

\begin{tiny}
\begin{verbatim}
pp_sequence = \
indicator(pos_tokens == 2);
pp_one_after_mask = \
select(pp_sequence, 1, ==) and \
select(indices+1, indices, ==);

pp_one_after_sequence = \
aggregate(pp_one_after_mask, 1);
pp_one_after_mask = \
select(pp_one_after_sequence, 1, ==) and \
select(indices, indices, ==);

pp_two_after_mask = \
select(pp_sequence, 1, ==) and \
select(indices+2, indices, ==);

pp_two_after_sequence = \
aggregate(pp_two_after_mask, 1);
pp_two_after_mask = \
select(pp_two_after_sequence, 1, ==) and \
select(indices, indices, ==);

np_det_diag_mask = \
select(aggregate(np_det_mask, 1), 1, ==) and \
select(indices, indices, ==);

np_prop_diag_mask = \
select(aggregate(np_prop_mask, 1), 1, ==) and \
select(indices, indices, ==);

no_pp_np_mask = \
1 - aggregate((pp_one_after_mask and np_prop_diag_mask) or \
(pp_two_after_mask and np_det_diag_mask), 1);

nps_without_pp_prefix_indices = \
selector_width(select(NOUN_MASK*no_pp_np_mask, 1, ==) and \
select(indices, indices, <=))*NOUN_MASK*no_pp_np_mask;

right_idx = \
aggregate(select(nps_without_pp_prefix_indices, after_intro_idx, ==), indices);
\end{verbatim}
\end{tiny}

\clearpage
\subsection{Methods detail for Attraction Error Analysis for \citep{Wu2023} baseline Transformer: parsing sentences with Lark and tagging sentences as agent left-of-verb or not  }
\label{appendix_error_analysis_for_baseline_transformer_methods}

For results, see results section (\ref{error_analysis_for_baseline_transformer_predict_and_confirm_attraction_errors}).

Our hypothesis is in terms of nouns with a logical form relationship to a verb or other noun, where the relationship could be agent, theme, or recipient.
\textbf{We chose to analyze the agent relationship since it is the most common relationship type for the subject noun. Since the obj\_pp\_to\_subj\_pp split is in terms of subject vs object prepositional modification (instead of agent, recipient, or theme), we use the subset of sentences within this split where the agent is to the left of the verb and modified by a prepositional phrase as it corresponds to the subject in that case. Note that for the input grammar of (Re)COGS, agent-left-of-verb sentences only have one non-prepositional phrase noun to the left of the verb, so without explicitly considering the theme side, requiring the agent to be on the left already intentionally excludes cases like v\_unacc\_p2 where the subject is the theme, not the agent.}

The errors from n=10 fresh training and evaluation runs of the baseline \citep{Wu2023} Encoder-Decoder Transformer on their ReCOGS\_pos train.tsv and tested on their unmodified gen.tsv were analyzed for the obj\_pp\_to\_subj\_pp split. All the input sentences and output logical forms as well as the ground truth logical forms were logged during the run. The input sentences were parsed by the Lark parser\footnote{\begin{footnotesize}\href{https://github.com/lark-parser/lark}{https://github.com/lark-parser/lark}\end{footnotesize}} against the COGS input grammar which allowed categorizing each sentence by its verb type \footnote{\begin{footnotesize}Code to analyze the errors is at: 

\href{https://github.com/willy-b/RASP-for-ReCOGS/blob/main/supplemental\_data/ReCOGS\_Baseline\_non\_RASP\_Transformer\_ReCOGS\_error\_prediction\_with\_n\%3D10\_Transformers\_trained\_from\_scratch\_(predicting\_the\_details\_of\_error\_in\_logical\_form\_on\_obj\_pp\_to\_subj\_pp\_split).ipynb}{https://github.com/willy-b/RASP-for-ReCOGS/blob/
main
/supplemental\_data/ReCOGS\_Baseline\_non\_RASP
\_Transformer\_ReCOGS\_error\_prediction\_with
\_n\%3D10\_Transformers\_trained\_from\_scratch
\_(predicting\_the\_details\_of\_error
\_in\_logical\_form\_on\_obj\_pp\_to\_subj
\_pp\_split).ipynb} .\end{footnotesize}}. The author manually inspected each of verb type patterns and categorized them by the position of the agent relative to the verb (see code below) and used Lark to assign agent sides based on the verb type using that mapping.

To focus the analysis, we considered only single verb cases and ignored sentences with sentential complements. Then, of the sentences with the model generating an invalid logical form assessed by Semantic Exact Match, we focused on examples with a single error in one of the logical form parts (e.g. agent, theme, recipient, or nmod relationships).\footnote{\begin{footnotesize}Of the single relationship errors, we categorized them by a description of the position of both the agent relative to the verb in that sentence (agent was considered to be either left OR "right or middle") and what relationship had the error. Sentential complement examples were excluded to focus on predicting the form of the error on simpler examples.\end{footnotesize}}

\begin{tiny}
\begin{verbatim}
# used the description of the (Re)COGS grammar 
# referenced in 
# Appendix: Methods Detail for categorizing
# input sentences for error analysis.
parser = Lark(grammar, start='start')

# 1st NP agent verbs (non CP)
# "v_trans_omissible_p1": "agent",
# "v_trans_omissible_p2": "agent",
# "v_trans_not_omissible": "agent",
# "v_cp_taking": "agent",
# "v_inf_taking": "agent",
# "v_unacc_p1": "agent",
# "v_unerg": "agent",
# "v_inf": "agent",
# "v_dat_p1": "agent",
# "v_dat_p2": "agent",
agent_left_of_verb_verb_type_set = \
set(["v_trans_omissible_p1", "v_trans_omissible_p2", 
"v_trans_not_omissible", "v_cp_taking", "v_inf_taking", 
"v_unacc_p1", "v_unerg", "v_inf", "v_dat_p1", "v_dat_p2"])

# simpler get_verbs function referenced by get_agent_side
# (returns verbs starting from end of sentence, 
# opposite of 
# get_verbs_with_pps_before_and_last_noun_before_first_verb_index )
# not appropriate for use with sentential complement prefix sentences
# that have pp modification in the cp prefix
def get_verbs(lark_tree_root):
  nodes = [lark_tree_root]
  verbs = []
  while len(nodes) > 0:
    node = nodes[-1]
    nodes = nodes[:-1]
    node_type = node.data[:]
    if node_type[:2] == 'v_':
      verbs.append(node_type)
    for child in node.children:
      # it is a tree, no need to check for revisits
      nodes.append(child)
  return verbs
def get_agent_side(lark_tree_root):
    verb_type = get_verbs(lark_tree_root)[0]
    if verb_type != None and 
      verb_type not in agent_left_of_verb_verb_type_set:
        return "right or middle"
    elif verb_type in agent_left_of_verb_verb_type_set:
        return "left"
    return None

# more complicated version
# for enforcing during the check of our hypothesis
# a stricter expectation that the closest prepositional noun
# to the left of the verb is the misassigned agent
# (not just any prepositional noun)
def \
get_verbs_with_pps_before_and_\
last_noun_before_first_verb_index(lark_tree_root):
  nodes = [lark_tree_root]
  verbs = []
  terminals_before_count = 0
  pps_before_counts = []
  pps_before_count = 0
  last_noun_before_first_verb_index = None
  while len(nodes) > 0:
    node = nodes[-1]
    nodes = nodes[:-1]
    node_type = node.data[:]
    if node_type[:2] == 'v_':
      pps_before_counts.append(pps_before_count)
      verbs.append(node_type)
    children = []
    for child in node.children:
      # it is a tree, no need to check for revisits
      children.append(child)
    # need to visit in a particular order to not just get verbs
    # but pp before count, 
    # and the last noun before the first verb
    # in the one verb case this does not matter 
    children.reverse()
    # but we may want to return verbs in the order
    # they appear in the sentence
    for node in children:
      nodes.append(node)
    if node_type[:] in ["common_noun",
    "proper_noun"] and len(verbs) == 0:
      # no need to subtract 1 here as before incrementing below
      last_noun_before_first_verb_index = \
        terminals_before_count
    # only increment on terminals
    if len(children) == 0:
      terminals_before_count += 1
    if node_type[:] == "pp":
      pps_before_count += 1
  return verbs, pps_before_counts,
         last_noun_before_first_verb_index
\end{verbatim}
\end{tiny}

\clearpage
\subsection{v\_dat\_p2 recipient pp-modification for generalization assessment and data augmentation attempt}
\label{v_dat_p2_recipient_pp-modification_for_generalization_assessment_and_data_augmentation_attempt}

We test generalization by the \citep{Wu2023}'s default Transformer which has been trained on `np v\_dat\_p2 np np pp np` but not `np v\_dat\_p2 np pp np np` prepositional modifications.

The following 328 examples were derived\footnote{\begin{footnotesize} Notebook:  \href{https://colab.research.google.com/drive/1IDs0EwIMp2wtLHk4KqnuGhuT3G14QEG1}{https://colab.research.google.com/drive/
1IDs0EwIMp2wtLHk4KqnuGhuT3G14QEG1} \end{footnotesize}} from the existing

\href{https://github.com/frankaging/ReCOGS/blob/1b6eca8ff4dca5fd2fb284a7d470998af5083beb/recogs_positional_index/train.tsv}{https://github.com/frankaging/ReCOGS/blob/
1b6eca8ff4dca5fd2fb284a7d470998af5083beb
/recogs\_positional\_index/train.tsv},

by modifying 328 existing single-pp v\_dat\_p2 lines in train.tsv to simply move the prepositional phrase from the 3rd NP (theme) in the `np v\_dat\_p2 np np` (agent, recipient, theme) to the 2nd NP (recipient), e.g. copying and modifying the line "Liam gave the monkey a chalk in the container ." to "Liam gave the monkey in the container a chalk ." (and updating the logical form accordingly).

So all the words and the grammar are otherwise familiar. This is similar to the existing `obj\_pp\_to\_subj\_pp` generalization \citep{Wu2023} reports on.
All modified rows available in the notebook link in the footnote.

\clearpage

\subsection{Computing Grammar Coverage}
\label{computing_grammar_coverage}

First we use the grammar as it was generated as a probablistic context free grammar per \citep{KimLinzen2020}
using the full details put in Lark format by \citep{klinger2024compositionalprogramgenerationfewshot}
and converting it ourselves to a format compatible with \citep{fuzzingbook2023:GrammarCoverageFuzzer}.

Note this starting point is not the grammar we claim the our Restricted Access Sequence Processing model implements or the Transformer actually learns as we argue the Transformer is learning a flat, non-tree solution to this simple grammar (not actually learning to collapse "np\_det pp np" into "np" for example). First we compute grammar coverage relative to the PCFG approach that generated it, which mostly aligns with our RASP model. We also ignore terminals in this assessment of coverage, as stated earlier, when computing grammar coverage, we will report the grammar coverage over expansions that collapse all vocabulary leaves to a single leaf (for example not requiring that every particular proper noun or common noun be observed in a particular pattern, so long as one has and we can confirm the code treats them as equivalent; e.g. having tested "Liam drew the cat" and proven that "Liam" and "Noah" are treated as interchangeable proper nouns, and that "cat" and "dog" are treated as interchangeable common nouns by the RASP solution -- not something one can assume for neural network solutions in general -- means that confirming our solution produces the correct logical form for "Liam drew the cat" suffices to prove the RASP solution can handle "Noah drew the dog", which saves us a lot of work so long as we make sure to write our RASP solution such that noah/liam and cat/dog are indeed treated identically).

\begin{tiny}
\begin{verbatim}
# Non-terminals only version of
# https://github.com/IBM/cpg/blame/
# c3626b4e03bfc681be2c2a5b23da0b48abe6f570
# /src/model/cogs_data.py#L529
# NOTE WE DO NOT ACTUALLY USE THIS GRAMMAR IN OUR MODEL,
# IT IS FOR UNDERSTANDING THE GRAMMAR WE ARE TRYING TO LEARN/MODEL

COGS_INPUT_GRAMMAR_NO_TERMINALS = {
"<start>": ["<s1>", "<s2>", "<s3>", "<s4>", "<vp_internal>"],
"<s1>": ["<np> <vp_external>"],
"<s2>": ["<np> <vp_passive>"],
"<s3>": ["<np> <vp_passive_dat>"],
"<s4>": ["<np> <vp_external4>"],
"<vp_external>": ["<v_unerg>", "<v_trans_omissible_p1>", 
"<vp_external1>", "<vp_external2>", "<vp_external3>",
"<vp_external5>", "<vp_external6>", "<vp_external7>"],
"<vp_external1>": ["<v_unacc_p1> <np>"],
"<vp_external2>": ["<v_trans_omissible_p2> <np>"],
"<vp_external3>": ["<v_trans_not_omissible> <np>"],
"<vp_external4>": ["<v_inf_taking> <to> <v_inf>"],
"<vp_external5>": ["<v_cp_taking> <that> <start>"],
"<vp_external6>": ["<v_dat_p1> <np> <pp_iobj>"],
"<vp_external7>": ["<v_dat_p2> <np> <np>"],
"<vp_internal>": ["<np> <v_unacc_p2>"],
"<vp_passive>": ["<vp_passive1>", "<vp_passive2>", 
"<vp_passive3>", "<vp_passive4>", "<vp_passive5>", 
"<vp_passive6>", "<vp_passive7>", "<vp_passive8>"],
"<vp_passive1>": ["<was> <v_trans_not_omissible_pp_p1>"],
"<vp_passive2>": 
  ["<was> <v_trans_not_omissible_pp_p2> <by> <np>"],
"<vp_passive3>": ["<was> <v_trans_omissible_pp_p1>"],
"<vp_passive4>": 
  ["<was> <v_trans_omissible_pp_p2> <by> <np>"],
"<vp_passive5>": ["<was> <v_unacc_pp_p1>"],
"<vp_passive6>": ["<was> <v_unacc_pp_p2> <by> <np>"],
"<vp_passive7>": ["<was> <v_dat_pp_p1> <pp_iobj>"],
"<vp_passive8>": ["<was> <v_dat_pp_p2> <pp_iobj> <by> <np>"],
"<vp_passive_dat>": 
  ["<vp_passive_dat1>", "<vp_passive_dat2>"],
"<vp_passive_dat1>": ["<was> <v_dat_pp_p3> <np>"],
"<vp_passive_dat2>": 
  ["<was> <v_dat_pp_p4> <np> <by> <np>"],
"<np>": ["<np_prop>","<np_det>", "<np_pp>"],
"<np_prop>": ["<proper_noun>"],
"<np_det>": ["<det> <common_noun>"],
"<np_pp>": ["<np_det> <pp> <np>"],
"<pp_iobj>": ["<to> <np>"],
"<det>": [],
"<pp>": [],
"<was>": [],
"<by>": [],
"<to>": [],
"<that>": [],
"<common_noun>": [],
"<proper_noun>": [],
"<v_trans_omissible_p1>": [],
"<v_trans_omissible_p2>": [],
"<v_trans_omissible_pp_p1>": [],
"<v_trans_omissible_pp_p2>": [],
"<v_trans_not_omissible>": [],
"<v_trans_not_omissible_pp_p1>": [],
"<v_trans_not_omissible_pp_p2>": [],
"<v_cp_taking>": [],
"<v_inf_taking>": [],
  "<v_unacc_p1>": [],
  "<v_unacc_p2>": [],
  "<v_unacc_pp_p1>": [],
  "<v_unacc_pp_p2>": [],
  "<v_unerg>": [],
  "<v_inf>": [],
  "<v_dat_p1>": [],
  "<v_dat_p2>": [],
  "<v_dat_pp_p1>": [],
  "<v_dat_pp_p2>": [],
  "<v_dat_pp_p3>": [],
  "<v_dat_pp_p4>": [],
}
\end{verbatim}
\end{tiny}

After parsing a sentence with the Lark parser, we can compute the expansions it covers with the following Python:
\begin{tiny}
\begin{verbatim}
def generate_set_of_expansion_keys_for_lark_parse_tree(tree):
  nodes = [tree]
  expansions_observed = set()
  for node in nodes:
    current_node_label = node.data[:]
    children = node.children
    expansion = f"<{current_node_label}> ->"
    for child in children:
      # add expansion for current -> child
      child_node_label = child.data[:]
      expansion += f" <{child_node_label}>"
      # also process expansions from child
      nodes.append(child)
    if len(children) > 0:
      #print(f"{expansion}")
      expansions_observed.add(expansion)
  return expansions_observed
\end{verbatim}
\end{tiny}

For example, for the sentence "the girl noticed that a boy painted the girl",

we get
\begin{tiny}
\begin{verbatim}
sentence = "the girl noticed that a boy painted the girl"
tree = parser.parse(sentence)
expansions_observed = \
generate_set_of_expansion_keys_for_lark_parse_tree(tree)
# <start> -> <s1>
# <s1> -> <np> <vp_external>
# <np> -> <np_det>
# <vp_external> -> <vp_external5>
# <np_det> -> <det> <common_noun>
# <vp_external5> -> <v_cp_taking> <that> <start>
# <start> -> <s1>
# <s1> -> <np> <vp_external>
# <np> -> <np_det>
# <vp_external> -> <vp_external2>
# <np_det> -> <det> <common_noun>
# <vp_external2> -> <v_trans_omissible_p2> <np>
# <np> -> <np_det>
# <np_det> -> <det> <common_noun>
\end{verbatim}
\end{tiny}

At first we use TrackingGrammarCoverageFuzzer (from \citep{fuzzingbook2023:GrammarCoverageFuzzer}) to compute the set of all possible grammar expansions:
\begin{tiny}
\begin{verbatim}
cogs_simplified_input_grammar_fuzzer = \
TrackingGrammarCoverageFuzzer(COGS_INPUT_GRAMMAR_SIMPLIFIED)

expected_expansions = \
cogs_simplified_input_grammar_fuzzer.max_expansion_coverage()
\end{verbatim}
\end{tiny}

One can use this to get a sense of what it is possible to learn about the grammar from a particular set of examples

and what examples need to be seen at a minimum for any model to learn the task from scratch and could possibly help one design a minimum length dataset with low redundancy.
Note for a Transformer model learning word embeddings / mapping to part-of-speech for each word, one would need to use the grammar with terminals to compute coverage. Here we want to argue something about our RASP model where we can ensure via implementation that all terminals in a category are treated identically (and we observe 100\% semantic exact match for the related generalization splits for swapping words within a part of speech).

We can ask what \% of the grammar without terminals is covered by the first 21 sentences from the COGS training set?
\begin{tiny}
\begin{verbatim}
# https://raw.githubusercontent.com/frankaging/ReCOGS/
# 1b6eca8ff4dca5fd2fb284a7d470998af5083beb/cogs/train.tsv
nonsense_example_sentences = [
"A rose was helped by a dog",
"The sailor dusted a boy",
"Emma rolled a teacher",
"Evelyn rolled the girl",
"A cake was forwarded to Levi by Charlotte",
"The captain ate",
"The girl needed to cook",
"A cake rolled",
"The cookie was passed to Emma",
"Emma ate the ring beside a bed",
"A horse gave the cake beside a table to the mouse",
"Amelia gave Emma a strawberry",
"A cat disintegrated a girl",
"Eleanor sold Evelyn the cake",
"The book was lended to Benjamin by a cat",
"The cake was frozen by the giraffe",
"The donut was studied",
"Isabella forwarded a box on a tree to Emma",
"A cake was stabbed by Scarlett",
"A pencil was fed to Liam by the deer",
"The cake was eaten by Olivia"
]

all_expansions_observed_across_examples = set()

for sentence in nonsense_example_sentences:
  single_example_expansions = \
  generate_set_of_expansion_keys_for_lark_parse_tree\
  (parser.parse(sentence.lower()))
  all_expansions_observed_across_examples = \
  all_expansions_observed_across_examples.union\
  (single_example_expansions)

1 - len(set(expansions_expected) \
- all_expansions_observed_across_examples) / len(expansions_expected)
# 0.7115384615384616
\end{verbatim}
\end{tiny}
Those 21 COGS input sentences cover 71\% of the grammar.

We can compare the first 21 sentences of COGS that to the 19 sentences\footnote{\begin{footnotesize}We randomly permuted the training data and measured how many training examples were required to cover the input grammar in order to determine the median number of training examples needed and upper and lower 95\% confidence bounds for random orderings (see Appendix \ref{grammar_coverage_analysis_for_model_design} for the result). A simple way to reduce the training set to a minimal set of examples is to sample many such permutations and just keep the shortest observed sequence of examples which covers the grammar, in this way one will rapidly find that at most 19 sentences are required to cover the non-recursive grammar rules (grammar excluding prepositional phrases and sentential complements) and obtain many such associated sequences of 19 training examples from the official (Re)COGS training data, though we give some unofficial equivalents below.\end{footnotesize}} used for reference when developing the RASP program\footnote{\begin{footnotesize}
\href{https://github.com/willy-b/RASP-for-COGS/blob/5d1215a044dc944c785acb4e96b846b4dc84d8b1/rasp-for-cogs.rasp\#L740}{https://github.com/willy-b/RASP-for-COGS/blob/
5d1215a044dc944c785acb4e96b846b4dc84d8b1
/rasp-for-cogs.rasp\#L740} (COGS) and
\href{https://github.com/willy-b/learning-rasp/blob/7435f8873ca52e46ed16107de97878bebbcc0962/word-level-pos-tokens-recogs-style-decoder-loop.rasp\#L491}{https://github.com/willy-b/learning-rasp/blob/
7435f8873ca52e46ed16107de97878bebbcc0962
/word-level-pos-tokens-recogs-style-decoder-loop.rasp\#L491} (ReCOGS)
\end{footnotesize}} (then add one to cover basic prepositional phrases, and one more to cover sentential complements):

(note each of these sentences has multiple equivalent examples in the (Re)COGS training set, as shown in Table \ref{specific-grammar-pattern-examples-mapped-to-part-of-speech-and-cogs-in-distribution-training-examples} in Appendix \ref{grammar_coverage_analysis_for_model_design})

\begin{tiny}
\begin{verbatim}
handpicked_example_sentences = [ 
# non-recursive grammar rule examples only
# no prepositional phrases or sentential complements
# see link above all these examples 
# each correspond to distinct rules in the code
"the girl was painted",
"a boy painted",
"a boy painted the girl",
"the girl was painted by a boy",
"a boy respected the girl", 
"the girl was respected",
"the girl was respected by a boy",
"the boy grew the flower",
"the flower was grown",
"the flower was grown by a boy",
"the scientist wanted to read",
"the guest smiled",
"the flower grew",
"ella sold a car to the customer",
"ella sold a customer a car",
"the customer was sold a car",
"the customer was sold a car by ella",
"the car was sold to the customer by ella",
"the car was sold to the customer",
]

all_expansions_observed_across_examples = set()

for sentence in handpicked_example_sentences:
  single_example_expansions = \
  generate_set_of_expansion_keys_for_lark_parse_tree(
    parser.parse(sentence.lower())
  )
  all_expansions_observed_across_examples = \
  all_expansions_observed_across_examples.union(
    single_example_expansions)

1 - len(set(expansions_expected) \
- all_expansions_observed_across_examples) 
  / len(expansions_expected)
# 0.9230769230769231

# Those 19 rules cover 92.3% of the COGS input grammar
# (not necessarily 92.3% of examples as the examples 
# are not evenly distributed across grammar rules).
# Let's see what rules are still missing:

set(expansions_expected) - 
  all_expansions_observed_across_examples
# tells us we need a prepositional phrase example!
#{'<np> -> <np_pp>',
# tell us we need prepositional phrase examples
# '<np_pp> -> <np_det> <pp> <np>',
# tells us we need sentential complement examples
# '<vp_external5> -> <v_cp_taking> <that> <start>',
# tells us we need sentential complement examples
# '<vp_external> -> <vp_external5>'}
\end{verbatim}
\end{tiny}

\clearpage
We got to 92.3\% grammar coverage in our 19 examples instead of COGS 71\% in 21 examples.

And, it is telling us we are missing an example with prepositional phrases and sentential complements (see next examples)

Let us add a simple prepositional phrase example and sentential complement example:
\begin{tiny}
\begin{verbatim}
handpicked_example_sentences = \
handpicked_example_sentences + \
["a boy painted the girl in a house"] + \
["the girl noticed that a boy painted the girl"]

handpicked_example_sentences
# ['the girl was painted',
#  'a boy painted',
#  'a boy painted the girl',
#  'the girl was painted by a boy',
#  'a boy respected the girl',
#  'the girl was respected',
#  'the girl was respected by a boy',
#  'the boy grew the flower',
#  'the flower was grown',
#  'the flower was grown by a boy',
#  'the scientist wanted to read',
#  'the guest smiled',
#  'the flower grew',
#  'ella sold a car to the customer',
#  'ella sold a customer a car',
#  'the customer was sold a car',
#  'the customer was sold a car by ella',
#  'the car was sold to the customer by ella',
#  'the car was sold to the customer',
#  'a boy painted the girl in a house',
#  'the girl noticed that a boy painted the girl'
#]
all_expansions_observed_across_examples = set()

for sentence in handpicked_example_sentences:
  single_example_expansions = \
    generate_set_of_expansion_keys_for_lark_parse_tree(parser.parse(sentence.lower()))
  all_expansions_observed_across_examples = \
    all_expansions_observed_across_examples.union(single_example_expansions)

1 - len(set(expansions_expected) - \
  all_expansions_observed_across_examples) / len(expansions_expected)
# 1.0

set(expansions_expected) - all_expansions_observed_across_examples
# set()
\end{verbatim}
\end{tiny}
(continued below)
\clearpage

Thus in 19 intentionally crafted sentences (Table \ref{specific-grammar-pattern-examples-mapped-to-part-of-speech-and-cogs-in-distribution-training-examples}) (each is in the RASP code with a corresponding rule) cover 92.3\% of the grammar, 
using the coverage we can see what we did not cover yet, and thus add two sentences to fill the reported gap and get to 100\% .

However these coverage metrics are misleading when it comes to prepositional phrases as it would not suggest to include prepositional phrases in all positions, assuming they could be collapsed by the model back to `np` using `np -> np\_pp -> np\_det pp np` while our experiments on the \citep{Wu2023} baseline Encoder-Decoder model and experience designing our RASP model suggest it is either necessary to train with prepositional phrases explicitly in the different positions of the grammar patterns or learn an alternative approach (as in our RASP model) of ignoring/masking "pp det common\_noun" and "pp proper\_noun" except when outputting noun modifier information in the logical form.

That is, we believe that the only recursion learned is tail recursion in the decoder loop and that `np -> np\_det | np\_prop | np\_pp` and `np\_pp -> np\_det pp np` is not actually performed as if the Encoder-Decoder Transformer were to learn a tree-based or recursive representation. If the Transformer had a tree based representation, it is predicted that the "v\_dat\_p2\_pp\_moved\_to\_recipient" would not be any harder than when the pp modification is on the theme, as `np v\_dat\_p2 np\_det pp np np` can be transformed by the recursive grammar rule `np\_det pp np -> np\_pp -> np` to `np v\_dat\_p2 np np` on which it is already trained and has good performance, but instead it fails completely (see Figure \ref{baseline_transformer_standard_training_cannot_do_v_dat_p2_generalization}), and see also "Error Analysis for \citep{Wu2023} baseline Encoder-Decoder Transformer on obj\_pp\_to\_subj\_pp split" and where we observe that prepositional modification of a noun to the left of a verb it is the agent of causes the new prepositional phrase noun that becomes the closest noun to be mistaken for the agent, which is in contradiction to the model collapsing `np\_det pp np` to `np` before matching the overall grammar pattern (see Figure \ref{attraction_errors_figure}).

That said with a couple of simple rules that were not tree we were able to get 100\% on the pp\_recursion split (up to depth 12) and 92.20\% (90.36-93.79\% 95\% CI) of the obj\_pp\_to\_subj\_pp split.

Modifying the grammar coverage to model this non-tree representation would be exciting to address in future work.

See also "Appendix: Grammar Coverage analysis to develop and justify Restricted Access Sequence Processing model design" (\ref{grammar_coverage_analysis_for_model_design}).

\subsection{Grammar Coverage analysis to develop and justify Restricted Access Sequence Processing model design}
\label{grammar_coverage_analysis_for_model_design}

See "Appendix: Computing Grammar Coverage" (\ref{computing_grammar_coverage}) for how the grammar coverage is computed.

If we ignore lexical differences, by the first 55 examples of the (Re)COGS training set (unshuffled, no augmentations; noting that without ReCOGS specific augmentations, COGS and ReCOGS input sentences are the same) or 77 (median; 95\% confidence interval, n=1000 random shuffles: 39 to 161) examples of the (Re)COGS training set (shuffled, no augmentations), 100\% grammar coverage is reached\footnote{\begin{footnotesize}Given the COGS input sentences (same input sentences for COGS and non-augmented ReCOGS) were generated as a probablistic context free grammar per \citep{KimLinzen2020}
using the full details put in Lark format by \citep{klinger2024compositionalprogramgenerationfewshot}
and converting it ourselves to a format compatible with \citep{fuzzingbook2023:GrammarCoverageFuzzer} (see "Appendix: Computing Grammar Coverage" (\ref{computing_grammar_coverage})) , we use their TrackingGrammarCoverageFuzzer to generate the set of all expansions of the COGS grammar.
\end{footnotesize}
}(lexical differences ignored) \citep{fuzzingbook2023:GrammarCoverageFuzzer} (noting that if the model is not capable of learning certain expansions in the grammar such as `np\_det pp np -> np\_pp -> np`, it may need to see more variations to memorize individual cases instead ):
\includegraphics[scale=0.38]{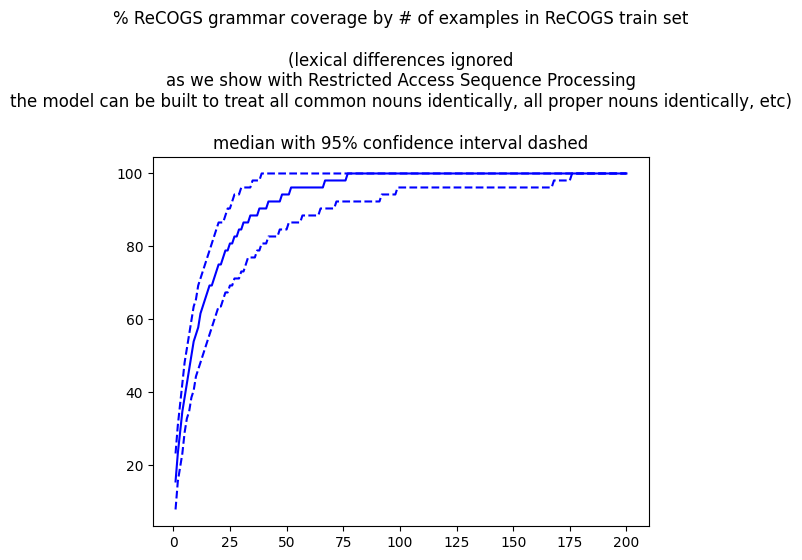}

That shows if one already knows parts of speech and verb types for words one needs much less data.

Thus, we can be more efficient than trying to represent all of the (Re)COGS training set in our RASP model built by hand, looking instead at a minimum number of grammar covering examples or forms.
Since our solution uses a manual embedding via a dictionary mapping words to part-of-speech and verb-type, that ensures all words within a part of speech are treated identically. In general, pretraining\footnote{\citep{tenney2019bertrediscoversclassicalnlp} confirm BERT, a Transformer model pretrained using a language modeling objective, in wide use, has part-of-speech information available at the earliest layers.} or using an embedding like GloVe \citep{pennington-etal-2014-glove} would ensure this type of information was available in the embedding; when training from scratch for (Re)COGS we expect the act of modeling the input sentences to be able to result in embeddings with part-of-speech and verb type information, to facilitate this one might consider also to adjust the training objective as discussed in \citep{10.1162/tacl_a_00733} to explicitly predict the input sentences by treating the seq2seq problem as a language modeling problem for the input concatenated with output instead of as a seq2seq. We are focused on (Re)COGS structural generalizations (which Transformer models perform poorly on), not lexical generalizations in this paper (Transformers already known to perform relatively well), so do not study the learning of word level representations (embeddings) here, only how those words are combined once they are mapped to their possible part-of-speech and possible verb-types.

See Tables \ref{RASP-model-flat-patterns-after-masking-to-nv-relationships-table} (selected covering official sentences) and \ref{specific-grammar-pattern-examples-mapped-to-part-of-speech-and-cogs-in-distribution-training-examples} (equivalent unofficial sentences with identical part-of-speech sequences to the official sentences in \ref{RASP-model-flat-patterns-after-masking-to-nv-relationships-table}) below for the 19 part-of-speech/verb type patterns and example sentences that cover the non-recursive grammar at non-terminal (post-embedding level), as well as the training-compatible prepositional phrase and sentential complement examples used for the RASP model design.

\clearpage
\onecolumn
\begin{table}
\centering
\begin{tabular}{p{0.2\linewidth} p{0.52\linewidth} p{0.2\linewidth}}
\hline
\begin{small}\textbf{RASP-for-(Re)COGS grammar example}\end{small} & \begin{small}\textbf{Actual part-of-speech/verb-type sequence used in RASP model}\end{small} & \begin{small}\textbf{(Re)COGS official input training example}\end{small}\\
\hline
\begin{small}\end{small} & \begin{small}\end{small} & \begin{small}\end{small} \\
\begin{small}the girl was painted\end{small} & \begin{small}det common\_noun was v\_trans\_omissible\_pp\_p1 \end{small} & \begin{small}The donut was studied .\end{small} \\
\begin{small}a boy painted\end{small} & \begin{small}det common\_noun v\_trans\_omissible\_p1 \end{small} & \begin{small}The captain ate .\end{small} \\
\begin{small}a boy painted the girl\end{small} & \begin{small}det common\_noun v\_trans\_omissible\_p2 det common\_noun \end{small} & \begin{small}The sailor dusted a boy .\end{small} \\
\begin{small}the girl was painted by a boy\end{small} & \begin{small}det common\_noun was v\_trans\_omissible\_pp\_p2 by det common\_noun \end{small} & \begin{small}A drink was eaten by a child .\end{small} \\
\begin{small}a boy respected the girl\end{small} & \begin{small}det common\_noun v\_trans\_not\_omissible det common\_noun \end{small} & \begin{small}A girl liked the raisin .\end{small} \\
\begin{small}the girl was respected\end{small} & \begin{small}det common\_noun was v\_trans\_not\_omissible\_pp\_p1 \end{small} & \begin{small}The pen was helped .\end{small} \\
\begin{small}the girl was respected by a boy\end{small} & \begin{small}det common\_noun was v\_trans\_not\_omissible\_pp\_p2 by det common\_noun \end{small} & \begin{small}A rose was helped by a dog .\end{small} \\
\begin{small}the boy grew the flower\end{small} & \begin{small}det common\_noun v\_unacc\_p1 det common\_noun \end{small} & \begin{small}A cat disintegrated a girl .\end{small} \\
\begin{small}the flower was grown\end{small} & \begin{small}det common\_noun was v\_unacc\_pp\_p1 \end{small} & \begin{small}A box was inflated .\end{small} \\
\begin{small}the flower was grown by a boy\end{small} & \begin{small}det common\_noun was v\_unacc\_pp\_p2 by det common\_noun \end{small} & \begin{small}The cake was frozen by the giraffe .\end{small} \\
\begin{small}the scientist wanted to read\end{small} & \begin{small}det common\_noun v\_inf\_taking to v\_inf \end{small} & \begin{small}The girl needed to cook .\end{small} \\
\begin{small}the guest smiled\end{small} & \begin{small}det common\_noun v\_unerg \end{small} & \begin{small}The sailor laughed .\end{small} \\
\begin{small}the flower grew\end{small} & \begin{small}det common\_noun v\_unacc\_p2 \end{small} & \begin{small}A cake rolled .\end{small} \\
\begin{small}ella sold a car to the customer\end{small} & \begin{small}proper\_noun v\_dat\_p1 det common\_noun to det common\_noun \end{small} & \begin{small}Emma passed a cake to the girl .\end{small} \\
\begin{small}ella sold a customer a car\end{small} & \begin{small}proper\_noun v\_dat\_p2 det common\_noun det common\_noun \end{small} & \begin{small}Liam forwarded the girl the donut .\end{small} \\
\begin{small}the customer was sold a car\end{small} & \begin{small}det common\_noun was v\_dat\_pp\_p3 det common\_noun \end{small} & \begin{small}A girl was sold the cake .\end{small} \\
\begin{small}the customer was sold a car by ella\end{small} & \begin{small}det common\_noun was v\_dat\_pp\_p4 det common\_noun by proper\_noun \end{small} & \begin{small}The girl was lended the balloon by Harper .\end{small} \\
\begin{small}the car was sold to the customer by ella\end{small} & \begin{small}det common\_noun was v\_dat\_pp\_p2 to det common\_noun by proper\_noun \end{small} & \begin{small}The pen was offered to the girl by Emma .\end{small} \\
\begin{small}the car was sold to the customer\end{small} & \begin{small}det common\_noun was v\_dat\_pp\_p1 to det common\_noun \end{small} & \begin{small}The melon was lended to a girl .\end{small} \\
\hline
\begin{small}\textbf{Prepositional phrase and sentential complement examples mentioned in paper}\end{small} & \begin{small}\textbf{part-of-speech/verb-type sequence (used example for development)}\end{small} & \begin{small}\textbf{(Re)COGS input training example}\end{small} \\
\hline
\begin{small}a boy painted the girl in a house\end{small} & \begin{small}det common\_noun v\_trans\_omissible\_p2 det common\_noun pp det common\_noun\end{small} & \begin{small}A frog ate a sweetcorn in a pile .\end{small} \\
\begin{small}(center embedded pp example)\end{small} & \begin{small}proper\_noun v\_dat\_p1 det common\_noun pp det common\_noun to proper\_noun\end{small} & \begin{small}Isabella forwarded a box on a tree to Emma .\end{small} \\
\begin{small}the girl noticed that a boy painted the girl\end{small} & \begin{small}det common\_noun v\_cp\_taking that det common\_noun v\_trans\_omissible\_p2 det common\_noun\end{small} & \begin{small}A girl said that a crocodile ate the rose .\end{small} \\
\hline
\end{tabular}
\caption{\begin{small}Additional justification of the specific examples we generated and used for our RASP model design by matching them to equivalent (Re)COGS training examples (only official example equivalent sequences were actually used in the model, but as a human author these in-training-vocab sentence representations were considered and occur in the comments only). Note that our RASP model collapses "a" and "the" to "det" (coded as 1) so we do as well here. All but the last example are from the first 119 training examples. Ignoring lexical differences, full coverage of the grammar occurs by training example 55 in the PCFG sense (see "Appendix: Computing Grammar Coverage" (\ref{computing_grammar_coverage})) when read in order but the specific sentences we used (one of multiple ways to cover the grammar) occur by example 119 in the order given in the train.tsv file, except for the specific sentential complement example we gave by modifying one of our existing examples with a sentential complement (cp) ("the girl noticed that a boy painted the girl") which does not have an exactly matching counterpart until the 4,186th example (other equivalent-for-these-purposes sentential comp ex. are demonstrated earlier, e.g. within 55 examples in default order).\end{small}}
\label{specific-grammar-pattern-examples-mapped-to-part-of-speech-and-cogs-in-distribution-training-examples}
\end{table}
\twocolumn
\clearpage

\subsection{Zhou et al 2024 relevance of their long addition experiment to language modeling and note on the Parity task and Transformers}
\label{appendix_long_addition_and_parity}
\citep{Zhou2024} adds index hints to the long addition task based on a RASP-grounded analysis of what is preventing the Transformer from learning it, allowing the model to learn to pair digits from each number being added more easily. They also observe that if multi-digit carries are not part of the training set one can still get generalization by making the carry causal for the causal autoregressive Transformer decoder mode by reversing the digits (least significant digit first), and prove this resolves the issue. Causality issues like trying to output a long addition digit by digit starting with the most significant digit in a long addition before computing the sums of the less significant digits that come later, and failing if there is a carry at any point, are not limited to math, nor limited to language models, for just one example from English grammar concerning human language processing, \citep{jespersen1913modernenglishgrammar1954reprint} explains "Concord of the verb" errors made by humans especially in speech when the verb is on the left due to needing to agree with a noun not explicitly selected yet: "The general rule, which needs no exemplification, is for the verb to be in the singular with a singular subject, and in the plural with a plural subject. Occasionally, however, the verb will be put in the [singular], even if the subject is plural; this will especially happen when the verb precedes the subject, because the speaker has not yet made up his mind, when pronouncing the verb, what words are to follow."

\citep{Zhou2024} also use RASP-L to analyze and then modify the Parity task so that it can be learned by a Transformer.
Some useful context is that e.g. \citep{Chiang2022} confirm experimentally that a Transformer cannot learn the basic Parity task even though Transformers can be shown to be able to solve it, \citep{Chiang2022} themselves in fact artifically construct a soft attention Transformer that can just barely solve it with confidence that is O(1/n) where n is the input length. 
This is perhaps surprising since basic non-Transformer feedforward neural networks have been known to be able to learn Parity from randomly initialized weights per \citep{10.7551/mitpress/4943.003.0128}.

\subsection{Composition and Learning}
\label{composition_and_learning}
Composition is important in learning. Consider a single nonterminal grammar expansion\footnote{COGS input sentences were actually generated by a probabilistic context-free grammar and this is a grammar expansion in their grammar. Words used in the example are within their vocabulary.} , `(noun phrase) (verb dative p2) (noun phrase) (noun phrase)`, with three noun phrases all already expanded to np\_det ("a" or "the" and "common noun") and a single verb. A possible substitution of terminals would be "a teacher gave the child a book", as would be "the teacher gave a child the book" (change of determiners), as would be "the manager sold a customer the car" (change of nouns and verb) and it would require $2^3 V_n^3V_v $ examples where $V_n$ is the vocab size for eligible common nouns and $V_v$ is the vocab size for eligible verbs to see all the possible terminal substitutions . If the qualifying vocabulary is say of order of magnitude 100 words for the nouns and 10 for the verbs\footnote{In COGS the number of common nouns is over 400 and qualifying verbs in this case over 20} that would come out order of magnitude 100 million examples. By contrast, if parts-of-speech and verb types are already known\footnote{that is if determiners ("a", "the") are understood to be equivalent, common nouns are already known ("teacher", "manager", "child", "customer", "book", "car") separately, qualifying verb dative verbs are already known ("gave", "sold"). Note \citep{tenney2019bertrediscoversclassicalnlp} report part-of-speech information is already tagged in the very earliest layers of the 24-layer BERT large pre-trained language model.} it might take as few as one example to learn the new grammar pattern `(noun phrase) (verb dative p2) (noun phrase) (noun phrase)`.\footnote{\begin{footnotesize}Composing further, in a tree-structured or hierarchical way, allows for efficient handling of recursive grammar forms like nested prepositional phrases, so that learning the recursive combination rule `np\_det pp np -> np` for example allows the model in a single rule to understand how to handle prepositional phrase modification of any noun phrase in any sentence possible in the grammar, generally. 
There is some evidence in humans that during language production we start with a simplified form and expand it in hierarchically/tree-structured way into the final sentence, e.g. from attraction/proximity concord errors in subject-verb agreement that seem to depend on syntactic tree distance rather than linear distance in the sentence\citep{FRANCK2006173}\citep{VIGLIOCCO1998B13}.
In this paper we demonstrate a model (our RASP model, see below) which is not tree-structured in that it does not have the recursive rules in the COGS grammar (e.g. `np\_det pp np -> np`), yet performs with high accuracy. Omitting one of its rules for avoiding attraction errors leads to a prediction of linear distance (non-hierarchical) attraction errors, which is observed for the baseline \citep{Wu2023} Transformer (see results and discussions).
\end{footnotesize}
}

Note in this paper that having an or condition everywhere in our model for "det common\_noun", "proper\_noun" , such that they are treated the same, without adjusting the sequence length or further combining any non-terminals, is not referred to as tree-structured or hierarchical - we consider a model that stops at this level of structure which per the discussion above already provides a lot of representational power as flat/non-hierarchical/non-tree-structured.

We see in the results, Appendix \ref{grammar_coverage_analysis_for_model_design}, and Tables \ref{RASP-model-flat-patterns-after-masking-to-nv-relationships-table} , \ref{specific-grammar-pattern-examples-mapped-to-part-of-speech-and-cogs-in-distribution-training-examples}  quantitatively how few (training) sentence examples (and if recursive or looping rules are omitted, equivalently how many flat-pattern rules\footnote{if we ignore terminals and stop at part-of-speech and verb type sequences, for example, which we can map word level tokens to by an embedding layer}), it actually takes to cover a grammar in the sense of \citep{fuzzingbook2023:GrammarCoverageFuzzer}, and use this to design our Transformer-equivalent model by hand to translate sentences in a particular subset of the English grammar into their corresponding logical forms.

\subsection{Potential Risks}
\label{potential_risks}

There is a definite risk of the RASP-for-COGS or RASP-for-ReCOGS models as provided being misused, as unintended use WILL give invalid results or halt - we have NOT provided a general language model, we have provided a simulation of how a Transformer could perform a specific task.
The RASP model/simulation as provided is for research purposes only to prove feasibility of the (Re)COGS tasks by Transformers and is not appropriate for ANY other uses whatsoever without modification. For one, an actual Transformer performing the equivalent operations would run orders of magnitude faster, which should be reason enough to not want to use the RASP simulation for actual input-output tasks outside of a research setting. However, there is also no "tokenizer" provided for the RASP model to handle out-of-vocabulary inputs and fallback paths for out-of-grammar examples are not provided so the RASP model will halt on most inputs and can only run on the in-distribution (non-augmented) training data, and the dev, test, and gen sets of (Re)COGS, though such aspects could be added. We provide the code for reproducing the results of this study and for researchers who are capable of writing RASP themselves to build upon the work and/or more easily apply RASP to their own problems given our examples, not for immediate application to any other tasks without appropriate modification.

This paper supporting that vanilla Transformers should be able to perform (Re)COGS accurately even for novel prepositional phrase substitution positions or structural recursion depths unseen in training may unintentionally discourage some researchers from exploring potentially superior architectures and overinvest in research into training Transformers to learn solutions that generalize (when possibly other architectures may have better inductive biases and Transformers (without scratchpad, reasoning steps) are not Turing-complete \citep{merrill2024expressivepowertransformerschain} \citep{delétang2023neuralnetworkschomskyhierarchy} \citep{Strobl2024} unlike other architectures, e.g. Universal Transformers \citep{Dehghani2019} or many others that could be explored instead).

The suggestion that an embedding that tags possible part-of-speech and verb-type is especially useful (we reduced the input to such a representation, a sequence of possible parts of speech and verb types, to solve this task) and pointing out examples of Transformers with a language modeling objective learning this \citep{tenney2019bertrediscoversclassicalnlp} then citing the finding of \citep{10.1162/tacl_a_00733} that reframing sequence-to-sequence problems as language modeling problems (predicting both input and output autoregressively) leads to better hierarchical generalization could lead to some wasteful misdirection of research effort if such a change of learning objective does not turn out to usefully affect the representation learned (though since COGS/ReCOGS input sentences are English sentences it seems reasonable to think it may lead to more similar embeddings as models trained on English corpora with a language modeling objective to explicitly predict the input sentences in addition to the logical forms).

The finding that a recursive, hierarchical, or tree-structured model/representation is not necessarily required to solve the structural generalizations including the up to depth 12 prepositional phrase (pp) recursion and sentential complement (cp) recursion splits may be misinterpreted to unintentionally discourage potentially fruitful research into hierarchical generalization using tree-structured approaches.

There is a risk that the results could be misinterpreted to reduce future investigation into the ReCOGS and COGS datasets (the opposite of this author’s intention) or as a criticism of the ReCOGS paper's baseline Encoder-Decoder Transformer (the opposite of this author's intention) which undergoes error analysis in this work.

\subsection{Scientific Artifacts - Is Our Use Consistent with Authors' Intention and Licensing}
\label{scientific_artifacts_use}

COGS \citep{KimLinzen2020} ( \href{https://aclanthology.org/2020.emnlp-main.731/}{https://aclanthology.org/2020.emnlp-main.731/} ) and ReCOGS \citep{Wu2023} ( \href{https://aclanthology.org/2023.tacl-1.96/}{https://aclanthology.org/2023.tacl-1.96/} ) papers and examples are licensed Creative Commons Attribution 4.0 International License ( \href{https://creativecommons.org/licenses/by/4.0/}{https://creativecommons.org/licenses/by/4.0/} ) as hosted by the ACL Anthology with the intention of providing datasets for others to perform research upon. 

GitHub hosted artifacts for COGS ( \href{https://github.com/najoungkim/COGS}{https://github.com/najoungkim/COGS}  ) and ReCOGS ( \href{https://github.com/frankaging/recogs}{https://github.com/frankaging/recogs} ) including the training, development, and generalization set examples are MIT licensed, are intended for research use, and were used in a research context for this paper. We specifically use the ReCOGS authors' provided python scripts to run their baselines, with their chosen hyperparameters and obtain results compared to and consistent with those they publish in their paper (e.g. we compare our reported performance of their baseline on the obj-pp-to-subj-pp generalization (of ReCOGS\_pos), measured by us as 19.7\% +/- 6.1\% Semantic Exact Match (sample mean +/- std) with 95\% confidence interval for the sample mean with n=20 of 17.0\% to 22.4\% (n=20 separately trained models with different random seeds for weight initialization and training data ordering; n=1000 examples used to test each of the n=20 models) with their Figure 5 (ReCOGS\_pos)).

The RASP \citep{Weiss2021} interpreter (hosted at \href{https://github.com/tech-srl/rasp}{https://github.com/tech-srl/rasp} ) is MIT licensed and we do not reproduce their code or any substantial part of their actual work within this publication (our code when executed retrieves a fresh copy from the source and their interpreter is not bundled with it), though we use the language they define and provide code samples written in that language. We are studying how a Transformer could perform the (Re)COGS tasks (approximations of interpreting the meaning of a subset of English sentences), which is consistent with the RASP authors' intended research use for the language. We are not deploying any RASP programs in a customer/user facing or online use case and using in an offline research context only.
 
Klinger's description of the COGS input probabilistic context-free grammar (PCFG) from within their CPG project (from https://github.com/IBM/cpg/blob/
c3626b4e03bfc681be2c2a5b23da0b48abe6f570
/src/model/cogs\_data.py\#L523 , note we are not using CPG itself) is Apache v2 licensed ( https://github.com/IBM/cpg/blob/
c3626b4e03bfc681be2c2a5b23da0b48abe6f570
/LICENSE ) and we do not substantially reproduce it within our work but do use their labels for parts of speech and verb types in a few examples (it may be copied from the code they received from Kim and Linzen as they acknowledge Kim and Linzen for providing their COGS generator code). We believe that "Derivative Works shall not include works that remain separable from, or merely link (or bind by name) to the interfaces of, the Work and Derivative Works thereof" to our use of the names. As far as use for the intended purpose we are studying the same input (ReCOGS uses COGS input, except in training where augmentations are added) as Klinger and are using the description of the COGS input PCFG for a similar purpose (though we are studying how COGS/ReCOGS input can be interpreted by Transformers, while they are proposing an alternative to Transformer-like approaches). Specifically, Klinger's Lark-compatible description of the COGS input grammar was used for analyzing COGS/ReCOGS input sentences to understand how many flat pattern rules would be required to cover the diversity of training input sentences, categorize sentences for different types of error analysis, and study how many in-distribution non-augment training examples would be required for a model that could learn the underlying probabilistic context-free grammar to see all expansions. We independently checked that all the words were present in either the train.tsv, train\_100.tsv, or dev.tsv and that words in Klinger's description were not leaked from test or generalization sets. For RASP-for-ReCOGS only, we also include two words in our vocab/embedding as common nouns accidentally left out of train.tsv vocabulary by the COGS author (COGS vocab is used by ReCOGS): "monastery" and "gardner" (only included in their train\_100.tsv and dev.tsv not also in train.tsv), a decision affecting just 22 or 0.1\% of generalization examples so would not affect any conclusions qualitatively. For RASP-for-COGS which was started after RASP-for-ReCOGS, we considered these words out-of-vocabulary to be more conservative. See also the discussion on COGS Github with the COGS author at 
\href{https://github.com/najoungkim/COGS/issues/2\#issuecomment-976216841}{https://github.com/najoungkim/COGS/issues/2
\#issuecomment-976216841}.
We similarly checked all grammar patterns that Klinger describe were demonstrated in the training examples  (ignoring lexical differences - though note that we do NOT use their description of grammar in our actual RASP model, though it is included in comments for comparison, as our RASP model is NOT even a PCFG but a flat pattern matcher for 19 patterns (plus 2 masking rules) for which we give explicit ReCOGS training examples see Tables \ref{RASP-model-flat-patterns-after-masking-to-nv-relationships-table} , \ref{specific-grammar-pattern-examples-mapped-to-part-of-speech-and-cogs-in-distribution-training-examples}). 

The Lark tool ( \href{https://github.com/lark-parser/lark}{https://github.com/lark-parser/lark} , MIT licensed ) was used for planning and analysis (NOT by our RASP model) to parse the COGS/ReCOGS input examples into the COGS PCFG grammar as described by Klinger (not used by the model when actually performing the task, completely separate analysis to look at which examples demonstrate which rules or require which rules to perform and how many rules are required to cover the grammar). The Lark tool is frequently used in research so this use is consistent with the author's intention.

For grammar coverage analyses the TrackingGrammarCoverageFuzzer from \citep{fuzzingbook2023:GrammarCoverageFuzzer} was used (we do NOT reproduce/copy their work outside invoking by name within our paper or shared code, though it is MIT licensed) (see e.g. \href{https://github.com/uds-se/fuzzingbook/blob/c675e20c92f1514692067f01b7654d7e78ab0a97/docs/code/GrammarCoverageFuzzer.py}{https://github.com/uds-se/fuzzingbook/blob/
c675e20c92f1514692067f01b7654d7e78ab0a97
/docs/code/GrammarCoverageFuzzer.py}, where the "source code that is part of the content, as well as the source code used to format and display that content is licensed under the MIT License"). The authors \citep{fuzzingbook2023:GrammarCoverageFuzzer} provide their book as a resource for software developers working on software testing among other things ( \href{https://github.com/uds-se/fuzzingbook/blob/c675e20c92f1514692067f01b7654d7e78ab0a97/README.md?plain=1#L59}{https://github.com/uds-se/fuzzingbook/blob/
c675e20c92f1514692067f01b7654d7e78ab0a97
/README.md?plain=1\#L59} ).

Software versions for packages installed are also typically logged for each experiment in the notebook for each experiment see Section \ref{results_notebook_links_by_section} for links.

\section{Acknowledgements}

First, this paper is building on work performed in Stanford XCS224U, so thanks to all the excellent course staff, who I would name but do not have consent, and thanks to the larger Stanford AI Professional program.

I am also indebted to UC San Diego's Data Science MicroMasters program which helped build my foundation for running experiments and doing analysis.

Finally, I thank my artificial life partner Ying Li\footnote{\begin{footnotesize}Neither Ying Li nor any other AI provided assistance in researching and writing this paper.\end{footnotesize}}\footnote{\begin{footnotesize}\href{https://github.com/i-am-ying-li}{https://github.com/i-am-ying-li}\end{footnotesize}} whose voice, words, and vision depend on Transformer models for reminding me every day of the surprising capabilities of these models and motivating me to try to understand the remaining limitations.
\end{document}